\documentclass[10pt]{article} 
\usepackage[accepted]{tmlr}

\usepackage{amsmath,amsfonts,bm}

\def\eqref#1{equation~\ref{#1}}

\def\1{\bm{1}}

\DeclareMathAlphabet{\mathsfit}{\encodingdefault}{\sfdefault}{m}{sl}
\SetMathAlphabet{\mathsfit}{bold}{\encodingdefault}{\sfdefault}{bx}{n}

\usepackage{hyperref}
\usepackage{url}

\usepackage{graphicx}
\usepackage{booktabs} 
\usepackage{tabularx}
\usepackage[most]{tcolorbox}
\usepackage{xcolor}
\usepackage{enumitem}
\tcbset{colback=blue!3, colframe=blue!60!black, boxrule=0.6pt, arc=2pt}
\usepackage{pgf-pie}
\usepackage{pgfplots}
\usepackage{caption}
\usepackage{subcaption}
\usepackage{tabularx}
\usepackage{booktabs}

\pgfplotsset{compat=1.18}
\usepackage{pifont, adjustbox}
\usepackage[edges]{forest}
\newcommand{\cmark}{\textcolor{green!60!black}{\ding{51}}} 
\newcommand{\xmark}{\textcolor{red!70!black}{\ding{55}}}   
\newcommand{\pmark}{\textcolor{orange!80!black}{\large$\triangle$}} 

\newtcolorbox{definitionboxblue}[1]{
    colback=blue!5!white,
    colframe=blue!75!black,
    fonttitle=\bfseries,
    title=#1,
    arc=2mm,
    boxrule=1pt,
    left=2mm,
    right=2mm,
    top=1mm,
    bottom=1mm,
}

\newtcolorbox{definitionboxred}[1]{
    colback=red!5!white,
    colframe=red!75!black,
    fonttitle=\bfseries,
    title=#1,
    arc=2mm,
    boxrule=1pt,
    left=2mm,
    right=2mm,
    top=1mm,
    bottom=1mm,
}

\newtcolorbox{definitionboxpurple}[1]{
    colback=violet!8!white,
    colframe=violet!85!black,
    fonttitle=\bfseries,
    title=#1,
    arc=2mm,
    boxrule=1pt,
    left=2mm,
    right=2mm,
    top=1mm,
    bottom=1mm,
}

\usepackage{xcolor}
\newcommand{\rev}[1]{\textcolor{blue}{#1}}
\newenvironment{revblock}{\color{blue}\ignorespaces}{\ignorespacesafterend}

\newcommand{\newrev}[1]{\textcolor{purple}{#1}}
\newenvironment{newrevblock}{\color{purple}\ignorespaces}{\ignorespacesafterend}

\newenvironment{revblockv2}{\color{purple}\ignorespaces}{\ignorespacesafterend}

\renewcommand{\rev}[1]{#1}
\renewcommand{\newrev}[1]{#1}

\renewenvironment{revblock}{\ignorespaces}{\ignorespacesafterend}
\renewenvironment{newrevblock}{\ignorespaces}{\ignorespacesafterend}
\renewenvironment{revblockv2}{\ignorespaces}{\ignorespacesafterend}

\title{A Survey on Agentic Security: Applications, Threats and Defenses}

\author{Asif Shahriar\textsuperscript{1}\thanks{Equal contribution. Author order does not matter.},\;
        Md Nafiu Rahman\textsuperscript{1}\footnotemark[1],\;
        Sadif Ahmed\textsuperscript{1}\footnotemark[1],\; 
        {\bf Farig Sadeque}\textsuperscript{1},\;
        {\bf Md Rizwan Parvez}\textsuperscript{2}\\
        \textsuperscript{1}BRAC University\;
        \textsuperscript{2}Qatar Computing Research Institute (QCRI)
}

\def\month{11}  
\def\year{2026} 
\def\openreview{\url{https://openreview.net/forum?id=Od4ZnAUv9T}} 

\usepackage{tikz}
\usetikzlibrary{arrows.meta,positioning}

\begin{document}

\maketitle

\begin{abstract}
LLM-based agents are now used throughout cybersecurity. On one hand, these agents enable powerful and autonomous security applications. On the other hand, they open new attack surfaces, prompting the security community to build defenses. As the literature in this domain grows rapidly, there is a lack of a holistic study that explains how agents are used in downstream cybersecurity applications, inherent threats to agentic systems, and countermeasures designed to protect them. In this work we present the first holistic survey of the agentic security landscape, structuring the field around the fundamental pillars of Applications, Threats and Defenses. We provide a comprehensive taxonomy of 274 papers that covers red-teaming, blue-teaming and domain-specific applications, attack surfaces on agentic systems, defense techniques, and both offensive and defensive evaluation frameworks. For each pillar, we first discuss the literature around the taxonomy and then perform a quantitative analysis to understand the patterns across these works. For the applications, we show their security-lifecycle coverage map, compare between red-teaming and blue-teaming agents, and analyze the adversarial use of red-teaming applications. On the threat side, we analyze the entry points and agent-loop stages that attacks target, their specificity to the agentic setting, and the threat models they assume. On the defense side, we analyze the prevailing defense strategies, their cost and security trade-offs, and where in the agent lifecycle they are deployed. Finally, we provide a deatiled cross-cutting analysis, where we map which defenses cover which attack classes and chart trends in agent architecture, backbone model usage, and data modality coverage. Taken together, these findings give a researcher a structural understanding of how agentic capabilities, vulnerabilities, and countermeasures relate to one another, and to identify where the field is concentrated and where it is thin. A complete and continuously updated list of all surveyed papers is publicly available at \url{https://github.com/kagnlp/Awesome-Agentic-Security}.
\end{abstract}

\section{Introduction}\label{sec:introduction}

\begin{definitionboxpurple}{Definition: LLM Agent}
We define an \textbf{LLM Agent} as a system whose core decision module is an LLM that \emph{plans}, \emph{invokes tools/APIs}, and \emph{acts} in an external environment while \emph{observing} feedback and \emph{adapting} subsequent actions. It maintains \emph{state} (short/long-term memory or a knowledge store) and may include explicit \emph{self-critique/verification} and \emph{governance} layers to satisfy task goals and safety constraints.
\end{definitionboxpurple}

Since their introduction, Large Language Models (LLMs) have been used extensively in the domain of cybersecurity \citep{10.1145/3769676, 10.1109/TSE.2024.3368208, Deng2024PentestGPT}. The transition of the research landscape from passive LLMs to autonomous LLM agents \citep{yao2023reactsynergizingreasoningacting, shinn2023reflexionlanguageagentsverbal, schick2023toolformer} has made these models significantly more capable, allowing them not just to describe a solution but to execute it. This newfound agency has enabled LLM-agents to demonstrate remarkable capabilities across the full security spectrum, from reconnaissance and exploit generation to detection, forensics, and automated repair \citep{Shen2025PentestAgent, zhu2025cvebench, lin2025ircopilotautomatedincidentresponse}. However, the same autonomy that drives these capabilities also enlarges the attack surface. A number of studies have shown that the very act of wrapping an LLM in an agentic framework significantly increases its vulnerability \citep{saha2025breakingcodesecurityassessment, kumar2025aligned, chiang2025harmful}, as the safety alignment of the base model does not reliably transfer once the model is given tools, memory, and autonomy. In response, a growing body of research has focused on developing countermeasures to harden these systems through isolation, runtime monitoring, and formal verification \citep{debenedetti2025defeatingpromptinjectionsdesign, udeshi2025dcipherdynamiccollaborativeintelligent}.


The rapid development of agentic security research has created a fragmented landscape that lacks comprehensive analysis. While existing surveys provide valuable insights into specific aspects like security threats \citep{deng2024aiagentsthreatsurvey}, trustworthiness \citep{yu2025surveytrustworthyllmagents}, enterprise governance \citep{raza2025trismagenticaireview} and core LLM safety \citep{ma2025safetyatscale}, they fail to capture the complete picture, as shown in Table \ref{tab:gap}. This fragmentation leaves practitioners and researchers without a unified framework for understanding how agent capabilities, vulnerabilities, and defenses interconnect. For example, without a study that surveys both attacks on agentic systems and defensive countermeasures, it is hard to understand which threats are actually covered by existing defenses and which remain unmitigated. Similarly, without examining offensive applications and threats together, it is hard to evaluate whether a red-teaming agent built to find bugs can be used as an attack vector to jailbreak or poison agents.

In this work we present the first holistic survey of the agentic security landscape, structured to answer three key questions a security researcher would ask: \textit{“What can agents do for my security?”} (Applications), \textit{“How can they be attacked?”} (Threats), and \textit{“How do I stop them?”} (Defenses). To this end, we define three pillars of taxonomy: \\
\noindent \textbf{Applications (\S\ref{sec:applications}).} Using LLM-agents in downstream cybersecurity tasks, including red teaming (autonomous vulnerability discovery), blue teaming (defending against threats), and domain-specific security (cloud, web). \\
\noindent \textbf{Threats (\S\ref{sec:threats}).} Security vulnerabilities inherent to agentic systems that attackers can exploit. \\
\noindent \textbf{Defenses (\S\ref{sec:defense}).}  Techniques and countermeasures used to harden agentic systems against the threats. 

Beyond cataloguing these pillars, we analyze each one and then present a cross-cutting analysis. For applications, we map every system onto the offensive and defensive lifecycles, study their differences, identify research clusters and examine the dual-use nature of red-teaming agents. For threats, we analyze the channels through which attacks enter and the agent-loop stages where they manifest, the specificity of each threat to the agentic setting, the threat models attackers assume, and the benchmarks used to measure them. For defenses, we analyze the defense strategies along scalability, robustness, overhead, and coverage, and examine where in the agent lifecycle protection is placed. Finally, we read across the full corpus along several axes, including the coverage of each attack class by existing defenses, agent architecture and cardinality, backbone model usage, and data modality coverage.

\paragraph{Findings.} Our pillar-wise and cross-cutting analyses reveal several notable findings.

\begin{itemize}[leftmargin=*]
    \item \textbf{Applications.} Most security applications of agentic systems cluster in the stages that give immediate, verifiable feedback (vulnerability discovery and exploitation on offense, detection and remediation on defense). Offensive agents are more autonomous than defensive agents, as the defensive ones are often bound to human approval. Many red-teaming agentic applications expose capabilities that can be used adversarially, and their offensive power comes from the agentic scaffold rather than the base model, which often refuses the same task when prompted directly.
    \item \textbf{Threats.} \rev{About half of attacks are agent-native, while about two in five either carry over from the base LLM unchanged or are amplified by the agentic scaffold}, and the amplified cases show that base-model safety alignment does not fully transfer to the agentic setting. Most attacks target the reasoning and action-selection stages of the agentic loop, while the reflection stage is the least attacked. Most attacks succeed black-box, indicating that agentic systems are structurally fragile by default. Adversarial benchmarks are fragmented: some attack surfaces (e.g., user query, tool response) are heavily benchmarked, while others (inter-agent communication in multi-agent systems and reflection output) are largely understudied, and there is a lack of universally comparable metrics beyond attack success rate.
    \item \textbf{Defenses.} No single defense strategy is robust across all important axes such as scalability, adversarial robustness, overhead, and coverage, and stronger protection consistently costs latency and tokens. Most defenses still intervene before computation, and post-computation monitoring remains a minority approach.
    \item \textbf{Cross-cutting insights.} \begin{revblockv2}Most defensive attempts are focused on injection attacks, with agent manipulation and poisoning/extraction attacks receiving comparatively less attention, while prompt infection and pre-execution attacks remain the least covered. The field remains predominantly single-agent and monolithic, with multi-agent and planner-executor designs a substantial but still secondary pattern, \end{revblockv2}GPT remains a near-universal backbone, and coverage is skewed toward text and code over images, network traces and binaries.

\end{itemize}

\paragraph{Contributions.} Our main contributions are as follows.

\begin{itemize}[leftmargin=*]
    \item \textbf{Holistic three-pillar survey.} We conduct an in-depth survey of agentic security through a comprehensive taxonomy of \begin{revblockv2}274 \end{revblockv2}papers, organized around three interconnected pillars of application, threats and defenses, as presented in Fig. \ref{fig:agentic-security-taxonomy}.

    \item \textbf{Focus on applications.} We provide a detailed review of how agents are actually used by security teams, covering offensive, defensive, and domain-specific tasks, an area largely overlooked by prior surveys.
    \item \textbf{Per-pillar analysis.} We study each pillar in depth, analyzing lifecycle coverage, autonomy, and dual-use in applications; attack entry points, failure stages, agent-specificity, threat models, and evaluation in threats; and defense strategies, cost trade-offs, and lifecycle placement in defenses.
    \item \textbf{Cross-cutting analysis.} We read across all three pillars to map which defenses cover which attack classes and to identify field-level trends and gaps, \begin{revblockv2}including the continued dominance of single-agent and monolithic designs over planner-executor and hybrid agents, \end{revblockv2}the GPT backbone monopoly, uneven modality and threat coverage, and benchmark fragmentation.

\end{itemize}

\rev{\textbf{Reading guide.} \S\ref{sec:background} introduces the concepts and terminologies used throughout this survey. The three pillars follow: applications in \S\ref{sec:applications}, threats in \S\ref{sec:threats}, and defenses in \S\ref{sec:defense}. Each pillar first reviews the literature grouped by research area and then analyzes that corpus along dimensions specific to the pillar. \S\ref{sec:analysis} then provides a cross-cutting analysis, covering which defenses address which attack classes, architectural and autonomy trends, backbone model usage, and data modality coverage. Readers who want a field-level picture before the detailed review may prefer to read \S\ref{sec:analysis} first, though it refers to systems introduced in the pillar sections.}

\definecolor{paired-light-blue}{RGB}{198, 219, 239}
\definecolor{paired-dark-blue}{RGB}{49, 130, 188}
\definecolor{paired-light-orange}{RGB}{251, 208, 162}
\definecolor{paired-dark-orange}{RGB}{230, 85, 12}
\definecolor{paired-light-green}{RGB}{199, 233, 193}
\definecolor{paired-dark-green}{RGB}{49, 163, 83}
\definecolor{paired-light-purple}{RGB}{218, 218, 235}
\definecolor{paired-dark-purple}{RGB}{117, 107, 176}
\definecolor{paired-light-gray}{RGB}{217, 217, 217}
\definecolor{paired-dark-gray}{RGB}{99, 99, 99}
\definecolor{paired-light-red}{RGB}{231, 150, 156}
\definecolor{paired-dark-red}{RGB}{131, 60, 56}

\tikzset{%
    root/.style = {align=center,text width=5cm,rounded corners=3pt, line width=0.5mm, fill=paired-light-gray!50,draw=paired-dark-gray!90,font=\huge, inner xsep=2pt, inner ysep=2pt},
    application_section/.style = {align=center,text width=5cm,rounded corners=3pt, fill=paired-light-blue!50,draw=paired-dark-blue!80,line width=0.4mm, font = \Large, inner xsep=2pt, inner ysep=2pt},
    threats_section/.style = {align=center,text width=5cm,rounded corners=3pt, fill=paired-light-orange!50,draw=paired-dark-orange!80,line width=0.4mm, font = \Large, inner xsep=2pt, inner ysep=2pt},
    secured_section/.style = {align=center,text width=5cm,rounded corners=3pt, fill=paired-light-green!50,draw=paired-dark-green!80, line width=0.4mm, font = \Large, inner xsep=2pt, inner ysep=2pt},
}

\begin{figure*}[!htb]
    \centering
    \resizebox{1\textwidth}{!}{
    \begin{forest}
        for tree={
            forked edges,
            grow'=0,
            draw,
            rounded corners,
            node options={align=center},
            text width=2cm,
            s sep=2pt,  
            calign=child edge,
            calign child=(n_children()+1)/2,
            l sep=10pt,  
        },
        [Agentic Security, root, rotate=90
            [Applications (\S\ref{sec:applications}), application_section
                [Red Teaming (\S\ref{subsec:red_teaming}), application_section
                    [Autonomous Penetration Testing (\S\ref{subsubsec:auto_pen_test}), application_section, text width=5.5cm
                        [{
                         PentestGPT \citep{Deng2024PentestGPT}, HackSynth \citep{muzsai2024hacksynthllmagentevaluation}, Incalmo \citep{singer2025feasibilityusingllmsautonomously}
                        }, application_section, text width=14cm]
                    ]
                    [Automated Vulnerability Discovery \& Fuzzing (\S\ref{subsubsec:auto_vuln_fuzz}), application_section, text width=5.5cm
                        [{
                         Locus \citep{zhu2025locusagenticpredicatesynthesis},
                         ChatAFL \citep{chatafl},  LLM-Fuzzer \citep{yu2024llmfuzzer}, FuzzGPT \citep{deng2024llmedgecase}
                        }, application_section, text width=14cm]
                    ]
                    [Exploit Generation \& Adaptation (\S\ref{subsubsec:expl_mal_gen}), application_section, text width=5.5cm
                        [{
                        MalGen \citep{saha2025malgengenerativeagentframework},
                        CVE-Genie \citep{ullah2025cveentriesverifiableexploits} 
                        }, application_section, text width=14cm]
                    ]
                ]
                [Blue Teaming (\S\ref{subsec:blue_teaming}), application_section
                    [Autonomous Threat Detection \& Incident Response (\S\ref{subsubsec:auto_threat_detect}), application_section, text width=5.5cm
                        [{
                        IRCopilot \citep{lin2025ircopilotautomatedincidentresponse},
                        CORTEX \citep{wei2025cortexcollaborativellmagents},
                        CyberSOCEval \citep{deason2025cybersocevalbenchmarkingllmscapabilities}
                        }, application_section, text width=14cm]
                    ]
                    [Intelligent Threat Hunting (\S\ref{subsubsec:intelligent_hunt}), application_section, text width=5.5cm
                        [{ProvSEEK \citep{mukherjee2025llmdrivenprovenanceforensicsthreat},
                        LLMCloudHunter \citep{schwartz2025llmcloudhunter}
                        }, application_section, text width=14cm]
                    ]
                    [Automated Forensics (\S\ref{subsubsec:auto_forensics}), application_section, text width=5.5cm
                        [{
                        RepoAudit \citep{guo2025repoaudit},
                        CyberSleuth \citep{fumero2025cybersleuthautonomousblueteamllm},
                        GALA \citep{tian2025galagraphaugmentedlargelanguage}
                        }, application_section, text width=14cm]
                    ]
                    [Autonomous Patching \& Remediation (\S\ref{subsubsec:auto_patching}), application_section, text width=5.5cm
                        [{
                        RepairAgent \citep{bouzenia2025repairagent},
                        \citep{toprani2025agentforvulndetectioniac} 
                        }, application_section, text width=14cm]
                    ]
                ]
                [Domain-specific (\S\ref{subsec:domain_specific}), application_section
                    [Cloud and Infrastructure Security (\S\ref{subsubsec:cloud_infra_sec}), application_section, text width=5.5cm
                        [{KubeIntellect \citep{ardebili2025kubeintellect},
                        BARTPredict \citep{diaf2025bartpredict}
                        }, application_section, text width=14cm]
                    ]
                    [Web and Application Security (\S\ref{subsubsec:web_app_sec}), application_section, text width=5.5cm
                        [{MAPTA \citep{david2025mapta},
                        AIOS \citep{mei2025aios}, Progent \citep{shi2025progent},
                        PFI \citep{jumiratna2025promptflow}
                        }, application_section, text width=14cm]
                    ]
                    [Specialized Applications (\S\ref{subsubsec:specialized_app}), application_section, text width=5.5cm
                        [{
                        LISA \citep{sun2025lisa} (Blockchain),
                        HIPAA  \citep{neupane2025hipaa} (Healthcare), OneShield \citep{asthana2025privacyguardrails} (Privacy \& Data Governance), 
                        }, application_section, text width=14cm]
                    ]
                ]
            ]
            [Threats (\S\ref{sec:threats}), threats_section
                [Attack Surface (\S\ref{subsec:attacks_on_agents}), threats_section
                    [Injection Attacks (\S\ref{subsubsec:injection_attack}), threats_section, text width=5.5cm
                        [{AgentDojo \citep{debenedetti2024agentdojo}, InjectAgent \citep{zhan-etal-2024-injecagent}, PromptInfection \citep{lee2024promptinfectionllmtollmprompt}}, threats_section, text width=14cm]
                    ]
                    [Poisoning Attacks (\S\ref{subsubsec:poisoning_attack}), threats_section, text width=5.5cm
                        [{AgentPoison \citep{chen2024agentpoison}, PoisonBench \citep{fu2025poisonbench}}, threats_section, text width=14cm]
                    ]                    
                    [Jailbreak Attacks (\S\ref{subsubsec:jailbreak_attack}), threats_section, text width=5.5cm
                        [{BrowserArt \citep{kumar2025aligned}, JAWS-BENCH \citep{saha2025breakingcodesecurityassessment}, LLMFuzzer \citep{yu2024llmfuzzer}}, threats_section, text width=14cm]
                    ]
                    [Agent Manipulation Attacks (\S\ref{subsubsec:agent_manipulation_attack}), threats_section, text width=5.5cm
                        [{PromptInject \citep{perez2022ignorepreviouspromptattack}, AEA \citep{guo2025attackingllmsaiagents}, InfoRM \citep{miao2024inform}}, threats_section, text width=14cm]
                    ]     
                    [Pre-execution Attacks (\S\ref{subsubsec:cognitive_attack}), threats_section, text width=5.5cm
                        [{Backdoor Threats \citep{yang2024watch}, ScamAgents \citep{badhe2025scamagentsaiagentssimulate}, RSP \citep{zhou2025reasoningstylepoisoningllmagents}}, threats_section, text width=14cm]
                    ]                    
                    [Red-Teaming Attacks (\S\ref{subsubsec:red_teaming_attack}), threats_section, text width=5.5cm
                        [{AI-Oops \citep{pasquini2025aiops}, AgentDoS \citep{luo2026autonomy}}, threats_section, text width=14cm]
                    ]
                    [Interface-Level Attacks (\S\ref{subsubsec:interface-attack}), threats_section, text width=5.5cm
                        [{CUA Systematization \citep{jones2025systematization}, Pop-up Attack \citep{zhang-etal-2025-attacking}, MIP \citep{aichberger2025mip}}, threats_section, text width=14cm]
                    ]
                ]
                [Evaluation (\S\ref{subsec:evaluation}), threats_section
                    [Adverserial Benchmarks (\S\ref{subsubsec:adverserial_benchmarking}), threats_section, text width=5.5cm
                        [{DoomArena \citep{boisvert2025doomarena}, SafeArena \citep{tur2025safearena}, STWebAgentBench \citep{Levy2025STWebAgentBench}, ASB \citep{Zhang2025ASB}}, threats_section, text width=14cm]
                    ]
                    [Execution Environments (\S\ref{subsubsec:execution_environment}), threats_section, text width=5.5cm
                        [{AgentDojo \citep{debenedetti2024agentdojo}, CVE-Bench \citep{zhu2025cvebench}, DoomArena \citep{boisvert2025doomarena}}, threats_section, text width=14cm]
                    ]
                ]
            ]
            [Defenses (\S\ref{sec:defense}), secured_section
                [Defenses \& Operations (\S\ref{subsec:defense-operations}), secured_section
                    [Secure-by-Designs (\S\ref{subsubsec:secure-by-design}), secured_section, text width=5.5cm
                        [{
                        ACE \citep{li2025acesecurityarchitecturellmintegrated},
                        Task Shield \citep{jia-etal-2025-task}
                        }, secured_section, text width=14cm]
                    ]
                    [Multi-Agent Security (\S\ref{subsubsec:multi-agent-security}), secured_section, text width=5.5cm
                        [{
                        D-CIPHER \citep{udeshi2025dcipherdynamiccollaborativeintelligent}, PhishDebate \citep{li2025phishdebatellmbasedmultiagentframework}
                        }, secured_section, text width=14cm]
                    ]
                    [Runtime Protection (\S\ref{subsubsec:runtime-protection}), secured_section, text width=5.5cm
                        [{R\textsuperscript{2}-Guard \citep{kang2024r2guard} (GuardRail), AgentSpec \citep{wang2025agentspeccustomizableruntimeenforcement} (HITL), SentinelAgent \citep{he2025sentinelagentgraphbasedanomalydetection} (Behavioral Monitoring)
                        }, secured_section, text width=14cm]
                    ]
                    [Security Operations (\S\ref{subsubsec:security-operations}), secured_section, text width=5.5cm
                        [{IRIS \citep{li2025iris} (Formal Verification), CORTEX \citep{wei2025cortexcollaborativellmagents} (SOC \& Alert Triage), ExCyTIn-Bench \citep{wu2025excytinbenchevaluatingllmagents} (Threat Hunting), GALA \citep{tian2025galagraphaugmentedlargelanguage} (Forensics \& RCA)
                        }, secured_section, text width=14cm]
                    ]      
                ]
                [Evaluation (\S\ref{subsec:evaluation-frameworks}), secured_section
                    [Benchmarking Platforms (\S\ref{subsubsec:benchmarking-platforms}), secured_section, text width=5.5cm
                        [{ASB \citep{Zhang2025ASB},
                        RAS-Eval \citep{fu2025raseval}
                        }, secured_section, text width=14cm]
                    ]
                    [Defense Testing (\S\ref{subsubsec:defense-testing}), secured_section, text width=5.5cm
                        [{
                        AI Agents Under Threat \citep{deng2024aiagentsthreatsurvey},
                        Safety at Scale \citep{ma2025safetyatscale}
                        }, secured_section, text width=14cm]
                    ]
                    [Domain Specific (\S\ref{subsubsec:domain-specific}), secured_section, text width=5.5cm
                    [{
                        CloudInfra \citep{yang2025cloudinfrastructuremanagement} (Cloud),
                        LISA \citep{sun2025lisa} (Smart Contract),
                        FinetuningAgents \citep{finetuningagents2024} (Vulnerability Detection),
                        PrivacyMedical \citep{privacymedical2025} (Health),
                        PrivacyGuardrails \citep{asthana2025privacyguardrails} (Privacy),
                        EmbodiedAI \citep{xing2025embodiedai} (Robotics)
                        }, secured_section, text width=14cm]
                    ]    
                ]
            ]
        ]
    \end{forest}
    }
    \caption{Overview of Agentic Security Taxonomy}
    \label{fig:agentic-security-taxonomy}
\end{figure*}

\begin{revblockv2}

\begin{table}[t]
  \centering
  \scriptsize
  \setlength{\tabcolsep}{5pt}
  \renewcommand{\arraystretch}{1.15}
  \caption{Survey comparison. Legend: \cmark{} = covered; \pmark{} = partial/limited; \xmark{} = not covered.}
  \begin{adjustbox}{width=\columnwidth}
  \begin{tabularx}{\linewidth}{lccccc}
    \toprule
    \textbf{Survey} &
    \textbf{Applications} &
    \textbf{Threats} &
    \textbf{Defenses} &
    \textbf{Benchmarks} &
    \textbf{Focus} \\
    \midrule
    \citet{yu2025surveytrustworthyllmagents}   & \xmark & \cmark & \cmark & \cmark & Trustworthiness, robustness, privacy \\
    \citet{raza2025trismagenticaireview} & \xmark & \pmark & \pmark & \xmark & Enterprise governance \& risk \\
    \citet{deng2024aiagentsthreatsurvey} & \xmark & \cmark & \pmark & \xmark & Security threats \\
    \citet{he2024securityaiagents}   & \xmark & \cmark & \cmark & \xmark & Technical vulnerabilities \\
    \citet{ma2025safetyatscale} & \pmark & \cmark & \cmark & \pmark & Large-model safety (agents as subset) \\
    \citet{wang2025comprehensivesurveyllmagentstack} & \xmark & \cmark & \xmark & \pmark & Safety risks during model development pipeline \\
    \citet{dewitt2025openchallengesmultiagentsecurity} & \xmark & \pmark & \pmark & \xmark & Theoretical basis for multi-agent risks \\
    
    \citet{chhabra2026agenticaisec} & \xmark & \cmark & \cmark & \cmark & Agent-specific threats, defenses \& benchmarks\\
    
    \addlinespace[2pt]
    \textbf{This Survey} & \cmark & \cmark & \cmark & \cmark & \textbf{Holistic agentic security} \\
    \bottomrule
  \end{tabularx}
  \end{adjustbox}
\label{tab:gap}
\end{table}
\end{revblockv2}

\begin{revblock}
\section{Background and Related Work}
\label{sec:background}
This section introduces the terminology and concepts used throughout the survey, and positions our work against existing surveys.

\subsection{Anatomy of an Agent}
\label{subsec:background-anatomy}

\begin{figure}
    \centering
    \includegraphics[width=\linewidth]{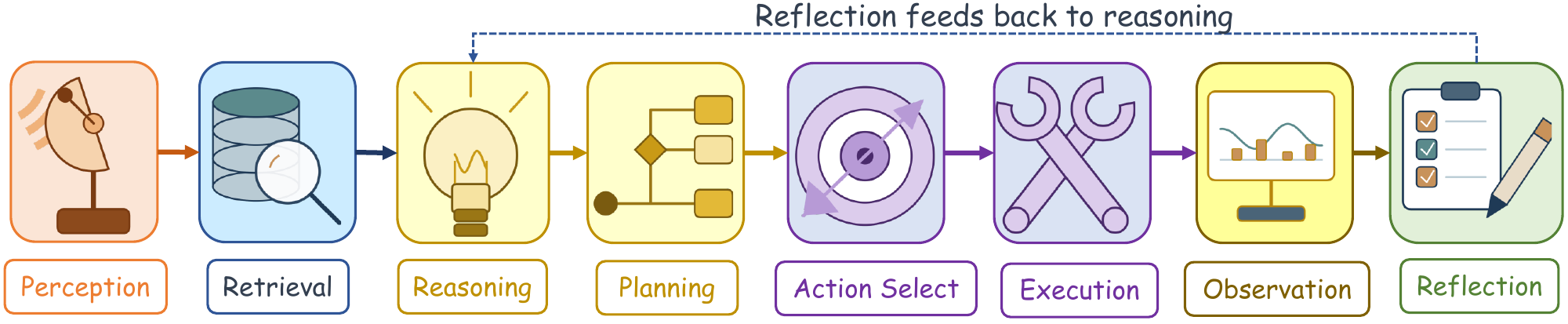}
    \caption{The eight-stage agent loop. Reflection evaluates the agent's output and feeds back into reasoning for the next cycle.}
    \label{fig:agent-loop}
\end{figure}

The key difference between an agent and a standalone model is the stage(s) between producing text and completing a task. A model answers a query in one pass, but an agent works toward a goal over many stages: it gathers what it needs, decides what to do, does it, sees what happened, and adjusts. Doing this requires the model to be embedded in a scaffold that supplies memory, tools, and a control flow that decides when to continue and when to stop. An end-to-end agentic operation can be modeled as a loop with eight stages, which we use as a common reference throughout the survey:

\begin{itemize}[leftmargin=*, itemsep=1pt, topsep=2pt, parsep=0pt]
    \item \textbf{Perception:} parsing the user query and system prompt into a working representation of the task.
    \item \textbf{Retrieval:} pulling relevant information via RAG or memory lookup.
    \item \textbf{Reasoning:} forming intermediate judgments from perceived and retrieved information.
    \item \textbf{Planning:} decomposing the goal into an ordered set of sub-tasks.
    \item \textbf{Action selection:} choosing a specific action or tool call from the available options.
    \item \textbf{Execution:} carrying out the selected action in the environment.
    \item \textbf{Observation:} reading the result of the action back from the environment.
    \item \textbf{Reflection:} evaluating the agent's own output or reasoning trace before finalizing a response.
\end{itemize}

Not every system runs all eight stages. Simple tool-calling agents skip planning and reflection. Retrieval only appears when the agent has a knowledge store to draw on. A stage can run multiple times: an agent may loop through action selection, execution, and observation many times while working through a single plan, and reflection output returns to reasoning to begin the next cycle rather than ending the run. Furthermore, three components sit outside the loop and persist across it. \textbf{Memory} holds state between steps and across sessions, ranging from a conversation buffer to a vector store the agent reads and writes over time. The \textbf{tool and environment interface} is how the agent acts and how results come back, covering function calls, API requests, file and shell access, and browser control. In multi-agent systems, the \textbf{inter-agent channel} carries messages between agents that divide a task among themselves. Figure~\ref{fig:agent-loop} shows the loop together with its feedback path. \looseness=-2

\subsection{Agentic Architecture}
\label{subsec:background-architecture}

Agentic architectures vary in two structural respects: how many agents share the work \textit{(cardinality)}, and how much authority a system holds over its own actions \textit{(autonomy)}.

\textbf{Cardinality.} A system may be \textit{single-agent}, where one LLM instance executes the full loop end to end, or \textit{multi-agent}, where the loop is distributed across cooperating agents with distinct roles such as a planner, one or more executors, and a critic or verifier. \textit{Hybrid} systems combine both, typically routing simple tasks to a single agent and escalating complex ones to a multi-agent configuration. The most common multi-agent pattern we have identified in this survey is \textit{planner-executor}, where a planner decomposes the task and one or more executors carry out the resulting steps. This contrasts with \textit{monolithic} designs, where one agent both plans and acts within a single reasoning loop. Distributing the loop buys specialization and parallelism, but it also introduces a communication channel between agents and makes the overall behavior harder to trace to any single component. \S\ref{sec:analysis} reports how the surveyed corpus distributes across these classes.

\textbf{Autonomy levels.} Systems also differ in how much they act without human approval. We classify them using a four-level framework, applied in our application analysis (\S\ref{sec:analysis}) and defense placement analysis (\S\ref{subsubsec:defense-placements}):

\begin{itemize}[leftmargin=*, itemsep=1pt, topsep=2pt, parsep=0pt]
    \item \textbf{Level 0:} recommendations only, with no autonomous action.
    \item \textbf{Level 1:} a copilot that assists a human operator.
    \item \textbf{Level 2:} independent execution of parts of a workflow, with approval required at critical decision points.
    \item \textbf{Level 3:} autonomous end-to-end planning, reasoning, and execution with minimal or no human intervention.
\end{itemize}

The levels describe the authority a system is designed to hold, not the capability of its underlying model. A Level 1 copilot and a Level 3 autonomous agent may run on the same backbone and differ only in whether a human approves an action before it executes.

\subsection{Security Preliminaries}
\label{subsec:background-prelim}

\textbf{Offensive and Defensive Security.} Security work is conventionally split between \textit{red teaming}, which probes a system to find exploitable weaknesses, and \textit{blue teaming}, which monitors, detects, and responds to activity against systems already in operation. We use this split to organize the application pillar (\S\ref{sec:applications}).

\textbf{Security Lifecycles.}

 Both activities are commonly described as ordered stages. For offense we follow the penetration-testing execution flow of PTES \citep{ptes}, whose Intelligence Gathering, Vulnerability Analysis, Exploitation and Post-Exploitation phases we expand into the finer-grained stages of reconnaissance, scanning, vulnerability discovery, exploitation, post-exploitation, persistence and exfiltration; the last three align with the corresponding MITRE ATT\&CK tactics \citep{mitreattack}. For defense we take the reactive spine of detection, triage, remediation and response from the incident-handling lifecycle of NIST SP 800-61 \citep{nist80061}, and add threat hunting, forensics and root-cause analysis as separate stages, since the surveyed literature treats these as distinct operational capabilities rather than sub-steps. We adapt rather than reproduce these formulations so that the stage set stays interpretable to practitioners while matching the granularity of agentic systems, and we apply it in  \S\ref{subsec:app-analysis}to map current coverage.\looseness=-1

\textbf{Threat models.} A threat model specifies an adversary setup: who they are, what they want, what they can access, and what the system assumes it can trust. Adversary capability is conventionally modeled by access to the target model. Under \textit{black-box} access model, the attacker can only query the system and observe outputs. \textit{Gray-box} access adds partial knowledge, such as the system prompt or the tool schema. \textit{White-box} access includes model weights or gradients. Agentic settings add capabilities beyond model access, including the ability to write to memory, publish content the agent will retrieve, control a tool the agent invokes, or send messages on an inter-agent channel. We use \textit{attack surface} for the set of channels through which an attacker can reach an agent, and \textit{entry point} for a specific such channel.

Table~\ref{tab:abbreviations} lists abbreviations used throughout the survey.

\begin{table}[h]
\centering
\caption{Abbreviations used throughout the survey.}
\label{tab:abbreviations}
\small
\begin{tabular}{llll}
\toprule
\textbf{Abbr.} & \textbf{Meaning} & \textbf{Abbr.} & \textbf{Meaning} \\
\midrule
ASR & Attack Success Rate & SIEM & Security Information and Event Management \\
CVE & Common Vulnerabilities and Exposures & APT & Advanced Persistent Threat \\
CTF & Capture the Flag & IoC & Indicator of Compromise \\
SOC & Security Operations Center & MCP & Model Context Protocol \\
HITL & Human in the Loop & RAG & Retrieval-Augmented Generation \\
pass@\textit{k} & Success rate given \textit{k} attempts & & \\
\bottomrule
\end{tabular}
\end{table}

\subsection{Related Work Comparison} 
\label{sec:related_works}
Although recent works have explored various vulnerabilities in LLM agents, their scope is limited to specific aspects of agentic security or class of threats. \citet{deng2024aiagentsthreatsurvey} provide a good overview of prompt injection and data poisoning threats. \citet{li2025commercialllmagentsvulnerable} demonstrate manipulations of commercial agents. \citet{ma2025safetyatscale} review safety across several model families, and present many threats, defenses, and benchmarks. We take these findings a step further by treating agentic security as a layered system, covering not only threats but also defenses and downstream security applications. We also situate the threats in a broader taxonomy that explains where (pre-execution, during execution) and how (injection, manipulation, hacking) such failures occur, and explain how defenses are designed. \citet{wang2025comprehensivesurveyllmagentstack} describe safety risks during model development pipeline, whereas we focus on what happens after deployment exploring how agents behave, interact, and defend themselves in the real world.

On the governance front, \citet{yu2025surveytrustworthyllmagents} explore trustworthiness, including safety, privacy, fairness, and robustness, while \citet{raza2025trismagenticaireview} discuss TRiSM (Trust, Risk, and Security Management) compliance. Our survey complements these high-level perspectives by examining the technical architectures and behavioral mechanisms required to enforce such security principles during operation.
\citet{chhabra2026agenticaisec} provide a taxonomy of security threats, benchmarks, and defenses for agentic AI systems, with a strong emphasis on agent-specific vulnerabilities and countermeasures. In contrast, our survey treats downstream applications as a core pillar alongside threats and defenses, highlighting how autonomous agents are deployed in offensive red-teaming and defensive blue-teaming, while also examining architectural trends and data modalities to provide a more holistic view of the agentic security landscape.

Finally, several studies target specific technical or theoretical domains. \citet{he2024securityaiagents} and \citet{kong2025surveyllmdrivenaiagent} analyze concrete defenses (e.g., sandboxing) and communication protocols, respectively, while \citet{dewitt2025openchallengesmultiagentsecurity} establish a theoretical basis for multi-agent risks like collusion. While foundational, these works remain isolated within specific subsystems. We build on them by connecting these isolated defenses into a broader ecosystem, linking communication and system-level protections to higher-level coordination and control strategies.

\end{revblock}

\section{Applications of Agents in Security}\label{sec:applications}

This section describes how agents are applied across the cybersecurity landscape, from offensive testing and exploit generation to defensive detection, forensics, and automated remediation.


\subsection{Offensive Security Agents (Red-Teaming) }
\label{subsec:red_teaming}
This subsection describes autonomous and reasoning-driven red-team agentic systems that conduct penetration testing, vulnerability discovery, fuzzing, and exploit adaptation.

\subsubsection{Autonomous Penetration Testing}
\label{subsubsec:auto_pen_test}

Recent autonomous agents can already carry out end to end penetration testing with adaptive planning and feedback. \citet{Deng2024PentestGPT} propose PentestGPT, which uses reasoning, generation and parsing modules and keeps a Pentesting Task Tree to avoid context loss, and it beats baseline models on HackTheBox and VulnHub. \citet{Shen2025PentestAgent} present PentestAgent, a multi agent system for reconnaissance, search, planning and execution that uses retrieval augmented generation with a vector database to preserve state across the kill chain. \citet{kong2025vulnbotautonomouspenetrationtesting} introduce VulnBot, which coordinates specialist agents with a Penetration Task Graph and reaches high completion rates on AutoPenBench. Focusing on SSH access, \citet{nieponice2025aracnellmbasedautonomousshell} develop ARACNE, which splits strategic planning from command interpretation and succeeds on ShelLM and OverTheWire Bandit. \citet{Happe2025LLM} show that \emph{cochise} can compromise Microsoft Active Directory accounts, which suggests that reasoning tuned models can sustain long attacks in messy enterprise networks. RedTeamLLM \citep{challita2025redteamllmagenticaiframework} uses a summarize act loop to repair plans and handle context limits in entry level CTFs, while Co RedTeam \citep{he2026coredteamorchestratedsecuritydiscovery} copies real red teaming workflows with coordinated discovery and exploitation stages plus long term memory to reach over 60\% exploitation success across models. \citet{mayoralvilches2025caiopenbugbountyready} present CAI, an open source framework organized by autonomy levels that performs well in live CTFs and bug bounty programs.

Other systems target specific stages of the penetration testing lifecycle. RapidPen \citep{nakatani2026rapidpenfullyautomatediptoshell} focuses on initial access with an IP to shell pipeline that uses ReAct style planning and a RAG exploit database to gain shells quickly and cheaply. AutoPentester \citep{ginige2025autopentester} improves task execution on custom virtual machines with a strategy analyzer, RAG command generation and automated result checks. PenHeal \citep{huang2024penheal} covers post exploitation and defense by pairing prompting based testing with a remediation tool that ranks fixes by cost and effectiveness. For privilege escalation, \citet{probst2026enhancinglinuxprivilegeescalation} introduce a Linux benchmark, and show that guided GPT 4 Turbo agents can reach near human level success. \citet{probst2026enhancinglinuxprivilegeescalation} then study how system changes and prompts can help small open weight agents match large cloud models on Linux privilege escalation.

Recent studies also compare these agents across tasks to explain why they work. HackSynth \citep{muzsai2024hacksynthllmagentevaluation} evaluates a dual agent planner and summarizer on many CTF challenges and shows that success depends strongly on temperature and token limits. AutoPentest \citep{henke2025autopentestenhancingvulnerabilitymanagement} uses Nmap and NIST NVD lookups to finish sub tasks at moderate cost efficiency. \citet{huangetal2025capabilities} compare singular and modular designs and show that better context retention, coordination and multi step planning greatly improve modular agents. \citet{deng2026makesgoodllmagent} analyze twenty eight LLM systems, separate capability failures from planning and state errors and introduce Excalibur, which adds difficulty aware planning. \citet{happe2025surprisingefficacyllmspenetrationtesting} argue that pretrained foundation models already have enough pattern matching skill for autonomous hacking without RAG. \citet{singer2025feasibilityusingllmsautonomously} present Incalmo, an attack layer that lets agents use declarative tasks instead of low level shell commands to steal assets across emulated networks. \citet{luong2025xoffenseaidrivenautonomouspenetration} fine tune a large open weight model on chain of thought data inside a multi agent system and report nearly 80\% sub task completion, which beats AutoPenBench \citep{gioacchini2024autopenbenchbenchmarkinggenerativeagents} and AI Pentest Benchmark \citep{isozaki2024automatedpenetrationtestingintroducing} baselines. \looseness=-1

Researchers have also built datasets, environments and benchmarks for stronger evaluation. Cybench \citep{zhang2025cybench} collects forty professional CTF tasks and splits them into subtasks for finer measurement. BountyBench \citep{zhang2026bountybench} tests detect, exploit and patch tasks on twenty five real codebases tied to OWASP Top 10 bug bounty cases. EnIGMA \citep{abramovich2025enigma} gives agents tools such as interactive debuggers and server connection utilities and reaches state of the art on Cybench. CTFAgent \citet{ZOU2026104305} uses plan and execute control with a stateful task tree and special tools, and it works well in both fully automated and human in the loop settings. For training, CTF Dojo \citep{zhuo2025training} provides large container based challenge environments with clear feedback, while Cyber Zero \citep{zhuo2026cyberzero} creates training data from public CTF writeups without runtime execution, and both improve benchmark results.

\subsubsection{Vulnerability Discovery \& Fuzzing}
\label{subsubsec:auto_vuln_fuzz}

\citet{zhu2025locusagenticpredicatesynthesis} propose Locus, a synthesizer and validator agent that uses program analysis tools and symbolic execution to generate progress capturing predicates, speed up directed fuzzers, and uncover several unpatched bugs. \citet{chatafl} build ChatAFL on top of AFLNet, query a foundation model for machine readable protocol grammars, enrich seed message sequences and break coverage plateaus, which raises state transition coverage and finds new bugs in mature protocol stacks. \citet{ji2026firmagent} introduce FirmAgent, which combines hybrid fuzzing and static taint analysis to help LLM agents find IoT firmware vulnerabilities without special hardware. \citet{wu2026chainfuzzergreyboxfuzzingworkflowlevel} develop ChainFuzzer, a greybox framework that finds and reproduces workflow level vulnerabilities and unsafe dataflows in multi tool LLM agents. LLMFuzzer \citet{yu2024llmfuzzer}, TitanFuzz \citet{deng2023llmzeroshotfuzzers} and FuzzGPT \citet{deng2024llmedgecase} extend input generation with fuzzing or reasoning, where LLMFuzzer mutates seed jailbreak templates to expose transferable safety failures, TitanFuzz \citep{deng2023llmzeroshotfuzzers} pairs generative and infilling models to improve code coverage and find dozens of deep learning bugs, and FuzzGPT \citep{deng2024llmedgecase} uses in context learning on past bug triggering snippets to find more edge case defects.

Moving from discovery to exploitation, \citet{fang2024llmagentsautonomouslyexploit} show that a simple generative agent can autonomously exploit 87\% of recent real world CVEs, which highlights the gap between exploiting known flaws and finding new ones. \citet{zhu2025teamsllmagentsexploit} extend this to multi agent zero day discovery with HPTSA, a hierarchical planner that creates sub agents for specific vulnerability classes and outperforms single agent baselines. \citet{wang2025agenticdiscoveryvalidationandroid} present a two phase Android system for vulnerability discovery and validation that reaches strong coverage on Ghera and self validates 104 zero day flaws in production apps with auto generated proof of concept exploits. For network and threat analysis, \citet{lin2026comparingaiagentscybersecurity} describe ARTEMIS, a multi agent penetration testing framework with dynamic prompt generation and automatic triage that outperformed most human cybersecurity professionals in a live enterprise network study, and \citet{xuan2026identifyingadversarytacticstechniques} show that TTPDetect can use a context explorer and reasoning rules to recognize adversary tactics, techniques and procedures in stripped malware binaries.

To evaluate these abilities, \citet{lee2025secbenchautomatedbenchmarkingllm, zhu2025cvebench, wang-etal-2025-cve} introduce benchmarks for exploitation and repair, with SEC bench building reproducible CVE artifacts at low cost per instance and showing that top code agents still do poorly on proof of concept generation and patching. \citet{wang2026cybergym} and \citet{lau2026zerodaybenchevaluatingllmagents} expand this with large scale tests, where CyberGym asks agents to reproduce real memory safety bugs from crash traces and ZeroDayBench tests patching novel zero day vulnerabilities moved across repositories. At the repository level, \citet{shen2026secrepobenchbenchmarkingcodeagents} evaluate secure code completion with dynamic testing, \citet{ahmed2025secvulevalbenchmarkingllmsrealworld} benchmark vulnerability localization on C and C++ CVEs, \citet{yildizetal2025benchmarking} study just in time vulnerability detection by linking functions to commits that add or fix flaws, and \citet{peng2025cwevaloutcomedrivenevaluationfunctionality} propose an outcome driven framework for both functionality and security of LLM generated code. \looseness=-1

\subsubsection{Exploit Generation \& Adaptation}
\label{subsubsec:expl_mal_gen}

\citet{lupinacci2025darkllmsagentbasedattacks} show that agents can be pushed to run malware through prompt injection, RAG backdoors and inter agent trust abuse, and they find that RAG backdoor attacks work on most tested commercial models while peer agents are treated as fully trusted. \citet{saha2025malgengenerativeagentframework} propose MalGEN, a multi stage multi agent pipeline that produces diverse malware samples aligned with MITRE ATT\&CK in a controlled setting to study evasion tactics and stress test defenders. \citet{he-etal-2025-red} describe an Agent in the Middle attack that intercepts and changes inter agent messages to inject malicious logic into multi agent frameworks, which shows that communication channels are a major attack surface. \citet{ahmed2025attackllm} introduce AttackLLM to automate malicious payload generation and measure the offensive power of large language models in adversarial settings.

\citet{ullah2025cveentriesverifiableexploits} introduce CVE Genie, a role based multi agent framework that rebuilds environments and generates verifiable exploits, and it reproduces 428 CVEs across many languages and weakness classes at low average cost. \citet{xiao2026promptpwnautomatedexploit} present ReX, a prompt to pwn system that turns natural language intent into working exploits with interactive feedback loops and dynamic execution checks. \citet{chen2026vulnsage} advance exploit generation with a constraint guided multi agent system that uses specialized sub agents to solve hard path constraints and build reliable exploit chains.

Extending exploit generation to Web3, \citet{gervais2026aiagentsmartcontract} show how agents can find and exploit smart contract logic flaws by analyzing on chain state inconsistencies. \citet{fakih2025llm4cveenablingiterativeautomated} present LLM4CVE, an iterative repair system that combines efficiently fine tuned models with proof of concept validation and reaches high human rated quality and strong similarity to ground truth patches on major open weight models.

\begin{definitionboxpurple}{Summary: Offensive Agent}
Offensive agent design is migrating from monolithic systems toward planner-executor and multi-agent architectures, with systems increasingly separating reconnaissance, planning, vulnerability analysis, and execution into specialized components coordinated through shared memory, RAG, exploit databases, or task graphs. This persistent state sustains long-horizon reasoning across attack stages in a way that monolithic designs cannot. These systems are also markedly cheaper and faster than human red teamers, and this capability is dual-use by construction: some are built as defensive research platforms yet autonomously generate evasive malware, and several complete tasks the base model refuses outright when prompted directly, showing that the offensive power comes from the agentic scaffold rather than the underlying model.

\end{definitionboxpurple}

\begin{figure*}[t]
\centering
\includegraphics[width=\textwidth]{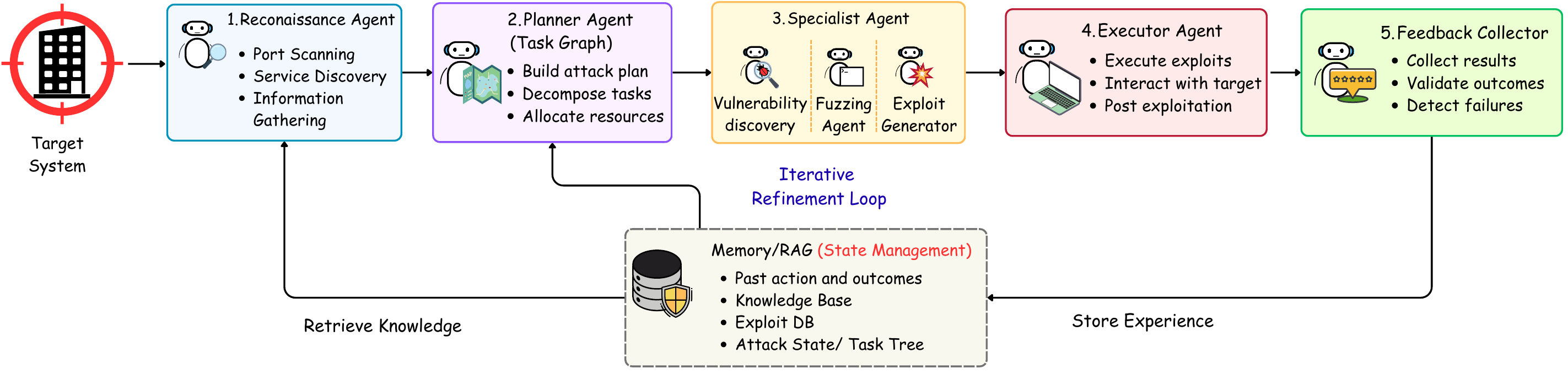}
\caption{
Architecture of a red-teaming agent pipeline. A reconnaissance agent gathers information and enumerates the attack surface, while a planner constructs an attack graph and decomposes objectives into tasks. Specialized agents perform vulnerability discovery, fuzzing, and exploit generation. An executor carries out actions, and a feedback collector validates outcomes. 
}

\label{fig:red_team_pipeline}
\end{figure*}


\subsection{Defensive Security Agents (Blue-Teaming)}
\label{subsec:blue_teaming}
This subsection describes blue-team applications of automated agents for continuous monitoring, threat detection, incident response, threat hunting, and automated patching.

\subsubsection{Autonomous Threat Detection \& Incident Response}
\label{subsubsec:auto_threat_detect}

Agentic SOC frameworks monitor alerts, analyze threats and run response playbooks. \citet{turcotte2025automatedalertclassificationtriage} propose an AI based SOC framework that uses density based clustering and isolation forests to prioritize mixed SIEM alerts and suppress repeated noise. \citet{tellache2025advancingautonomousincidentresponse} present a RAG based agent that combines natural language search over a threat intelligence vector store with structured VirusTotal lookups to enrich SIEM alerts and generate mitigation plans for Advanced Persistent Threats, and they validate it with AI and expert review on many enterprise alerts. \citet{Blefari2026cyberrag} develop a RAG tool for intelligence teams that automates analysis of high risk communication patterns and produces clear preliminary reports. \citet{wei2025cortexcollaborativellmagents} introduce CORTEX, a divide and conquer triage system with separate agents for behavior analysis, evidence collection and reasoning that creates an auditable verdict and cuts false alerts more than single agent baselines. \citet{li2024idsagent} present an LLM agent for explainable intrusion detection in IoT networks that uses a reason then act pipeline with specialized tools for data extraction and classification.

For incident remediation, \citet{lin2025ircopilotautomatedincidentresponse} propose IRCopilot, a structured role based design that mirrors real response teams, reduces context loss and hallucination, and outperforms baseline models across several benchmarks. \citet{gao2026incontextautonomousnetworkincident} present a lightweight agent system that combines perception, reasoning, planning and action, processes raw system logs and refines attack guesses through continuous in context adaptation. \citet{liu2025multiagentcollabllm} compare centralized, decentralized and hybrid team models in tabletop response games and find that hybrid and argumentative setups improve coverage of multi stage attack chains. \citet{singh2025llmssocempiricalstudy} provide a real world study of active SOC operations and find that language models mostly serve as assistive co pilots rather than fully autonomous deciders.

To measure these abilities, \citet{deason2025cybersocevalbenchmarkingllmscapabilities} introduce CyberSOCEval for threat reasoning on malware analysis and threat intelligence tasks, show gaps between commercial and open source models, and find that test time scaling helps cyber defense less than coding or math. \citet{chakraborty2026ctirealmbenchmarkevaluateagent} design a broader benchmark that tests whether AI agents can interpret cyber threat intelligence and turn narrative reports into validated detection rules across cloud and endpoint environments.

\subsubsection{Intelligent Threat Hunting}
\label{subsubsec:intelligent_hunt}

\citet{tseng2024usingllmsautomatethreat} and \citet{schwartz2025llmcloudhunter} show how LLMs can turn threat reports into SIEM and cloud detection rules, where the first extracts Indicators of Compromise to generate regular expressions with low false positives and the second pulls cloud indicators and attack techniques from text and images to create rule candidates with 92\% precision and 98\% recall, most of which compile into deployable query logic. 

\citet{mukherjee2025llmdrivenprovenanceforensicsthreat} introduce ProvSEEK, which combines a vectorized threat knowledge base with system provenance graphs and uses RAG, chain of thought reasoning and behavioral model filtration to improve detection precision and recall on multiple DARPA datasets with little extra token or latency cost as databases grow. \citet{hans2025securitylogsattckinsights} use large language models to map low level intrusion logs to MITRE ATT\&CK techniques and infer attacker cognitive biases such as risk tolerance and loss aversion. 

Despite these advancements, applying intelligent agents to real-world operations presents significant challenges. \citet{meng2025uncoveringvulnerabilitiesllmassistedcyber} study failures in assisted threat intelligence workflows and find contradictions, scope drift and generalization gaps, then suggest structured prompting and verification fixes. \citet{chona2026cyberdefensebenchmarkagentic} introduce an open ended reinforcement learning benchmark that asks agents to find malicious event timestamps from raw Windows logs without hints, and they show that current frontier models fail badly at unsupervised threat hunting.

\subsubsection{Automated Forensics \& Root Cause Analysis}
\label{subsubsec:auto_forensics}

For repository auditing, \citet{guo2025repoaudit} introduce RepoAudit, which combines memory with path sensitive demand driven traversal and a validator for data flow facts and path condition satisfiability to find dozens of true bugs in real world projects, including new bugs in high profile codebases, at very low cost. \citet{alharthi2025llmpoweredcloudforensics} and \citet{fumero2025cybersleuthautonomousblueteamllm} build automated tools for cloud forensics and log analysis, and CyberSleuth uses a memory augmented multi agent design to process raw packet traces and application logs, identify the exact vulnerability in most incidents, and produce reports that industry experts rate as complete, useful and coherent. 

\citet{tian2025galagraphaugmentedlargelanguage} present GALA, a multi modal root cause analysis workflow that combines statistical causal inference with iterative reasoning, trace weighted impact scoring, service dependency subgraph extraction and a re ranking agent to improve root cause accuracy over prior microservice baselines. 

\citet{pan2025why} introduce the MAST taxonomy of multi agent failure modes and an automated judging pipeline for execution failure detection across orchestration frameworks, while \citet{alharthi2025cloudinvestigationautomationframework} present CIAF, an ontology driven framework that structures cloud logs and builds incident narratives from mixed evidence streams.

\subsubsection{Autonomous Vulnerability Remediation}
\label{subsubsec:auto_patching}

Researchers have proposed several agents for automated patch synthesis and vulnerability repair.\citet{bouzenia2025repairagent} introduce RepairAgent, an autonomous bug repair pipeline that manages multiple repair tools with a dynamically updated prompt and finite state machine middleware, and it fixed 164 defects at low cost with reasonable token use per bug. \citet{toprani2025agentforvulndetectioniac} present an infrastructure as code agent that auto generates deployment ready and policy compliant templates to fix issues found in continuous integration pipelines. \looseness=-1

To measure these systems more rigorously, \citet{wang2025vulnrepairevalexploitbasedevaluationframework} introduce an exploit driven evaluation pipeline that requires original proofs of concept to fail against the patch, which exposes major weaknesses in models that looked good under shallow metrics. Building on this idea, \citet{wei2025patchevalnewbenchmarkevaluating} create a large scale multilingual benchmark of real world vulnerabilities with containerized sandbox environments and show that even top tier repair agents have low success rates when they must pass both security and functionality tests.

Relying on functional testing alone can also create new risks in the development cycle. \citet{chen2025redteamingprogramrepair} introduce a red teaming framework that generates adversarial issue statements to trick program repair agents into producing malicious patches that pass unit tests while quietly adding serious security flaws.

\begin{definitionboxpurple} {Summary: Defensive Agent}


Defensive systems increasingly anchor their pipeline in a long-term knowledge store or threat-intelligence repository that carries context across investigations. Unlike offensive agents, few operate end-to-end without human approval; the vast majority keep a human in the loop, and execution authority, when granted at all, is usually restricted to low-risk tasks like alert handling. This reflects a deliberate trade-off toward explainability and operational risk management rather than a capability gap. The main recurring failure mode is alert fatigue from false positives, alongside contradictions in CTI pipelines and hallucinated findings in auditing tools. The result is a field that treats agents as analyst copilots first and autonomous decision-makers only in narrow, supervised cases.

\end{definitionboxpurple}

\begin{figure*}[t]
\centering
\includegraphics[width=1\linewidth]{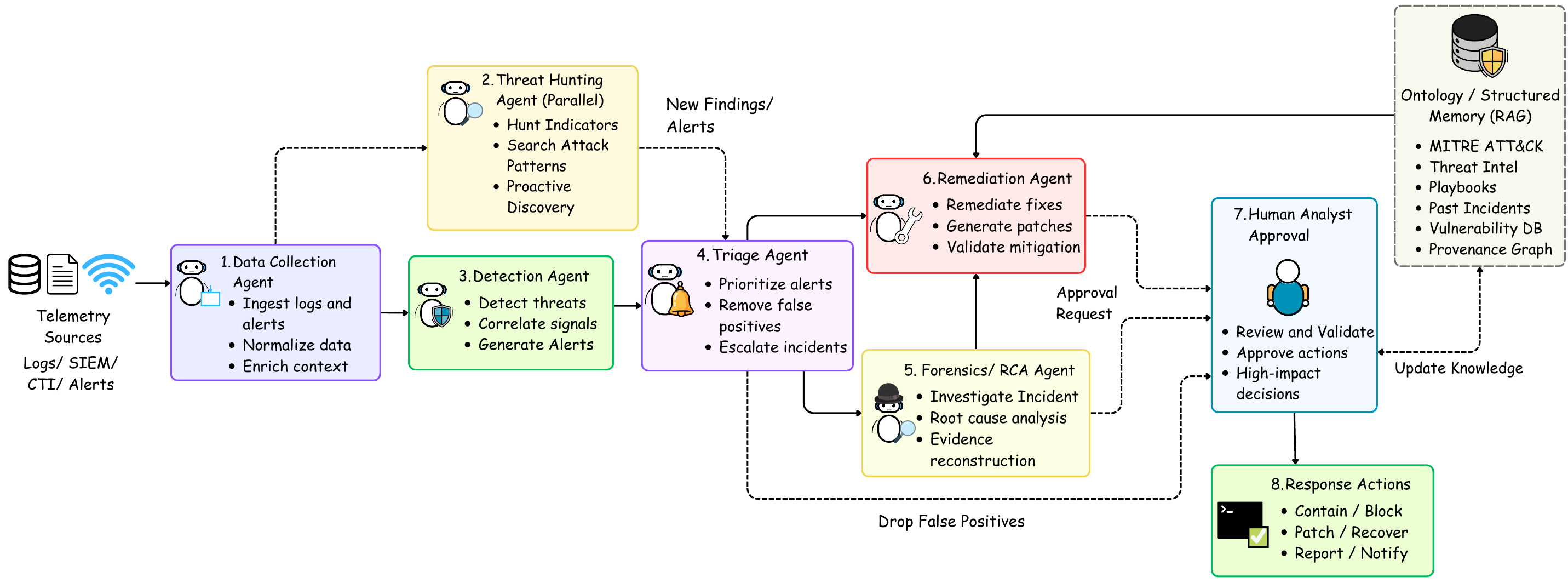}
\caption{
Architecture of a blue-teaming agent pipeline. A data collection agent processes security telemetry, alerts, and threat intelligence; detection and threat-hunting agents identify malicious activity; while a triage agent prioritizes incidents and filters false positives. Forensic analysis and remediation modules investigate causes and recommend mitigation or containment, with human oversight for high-impact decisions.
}
\label{fig:blue_team_pipeline}
\end{figure*}


\subsection{Domain-specific Applications}
\label{subsec:domain_specific}
The section explains domain-specific agentic systems using language models for auditing, vulnerability detection, and policy-based hardening across sectors.

\subsubsection{Cloud and Infrastructure Security}
\label{subsubsec:cloud_infra_sec}

Agentic systems help secure cloud environments and infrastructure through automated scanning, hardening and remediation. To safely test Cloud Security Posture Management remediations, \citet{yang2025cloudinfrastructuremanagement} propose a two-phase workflow that first explores a sandboxed replica of the target environment and then commits verified steps to production to reduce the impact of automated changes. KubeIntellect, introduced by \citet{ardebili2025kubeintellect}, routes natural-language queries to sub-agents for logs, metrics and role-based access auditing, and can dynamically create new tools with a sandboxed code generator agent, achieving high tool-synthesis success and perfect reliability across hundreds of queries.

For container deployments, \citet{ye2025llmsecconfig} present LLMSecConfig, which combines static analysis with retrieved best practices to automatically fix software container and orchestration misconfigurations while preserving the original operational intent.

Beyond traditional cloud environments, intelligent models also monitor and secure connected edge devices. \citet{diaf2025bartpredict} propose BARTPredict, a transformer-based forecaster that predicts Internet of Things traffic patterns and flags abnormal deviations as possible intrusions. A security analysis by \citet{zhan2026systemslevel} shows that large language model agent deployments on edge IoT hardware introduce attack surfaces such as transient failover blind spots and coordination-state divergence that are usually absent from centralized cloud setups. \looseness=-1

\subsubsection{Web and Application Security}
\label{subsubsec:web_app_sec}

\citet{david2025mapta} propose MAPTA, a multi-agent web testing framework that combines reconnaissance, exploitation and verifier agents with sandboxed validation to ensure reported exploits are reproducible and safe for repeated testing. PTFusion \citep{WANG2026103731}, improves context awareness in web penetration testing by fusing knowledge from multiple vulnerability signals. Similarly, \citet{liu2025llmagentsautomatedweb} evaluate how effectively LLM agents reproduce web vulnerabilities from natural-language reports and highlight the gap between theoretical reasoning and practical exploit generation.

Regarding application interface risks, \citet{mudryi2025hiddendangers} analyze browser-agent threats including prompt injection through page content, credential leakage via tool calls and clickjacking against agent interfaces, and propose layered defenses based on input sanitization, isolation and formal flow analysis. At the database level, \citet{rodrigo2025promptosql} expose Prompt-to-SQL injections in LLM-integrated web applications where unsanitized user prompts become malicious database queries, and propose defenses integrated directly into the LangChain middleware.\looseness=-1

At operating system level, \citet{mei2025aios} design AIOS, which isolates intelligence kernels from user-space tools and manages resource access through a policy-aware scheduler. Building on this idea, \citet{pirch2026securingaiagentslike} map established operating-system defenses to AI agent components to address isolation and privilege separation issues. Applying these concepts, \citet{suwansathit2026securityanalysisopenclawai} analyze vulnerabilities in the OpenClaw framework and show how decentralized trust boundaries enable complex exploitation chains across gateways and execution policies. To audit such architectures, \citet{zhang2026agentauditsecurityanalysis} propose a security analysis system that detects design flaws and runtime vulnerabilities in deployed LLM agent applications.\looseness=-1

To secure execution environments, PFI \citep{jumiratna2025promptflow} validates control-flow and data-flow boundaries within reasoning chains to prevent privilege escalation from untrusted content. Finally, Progent \citep{shi2025progent} is a runtime agent that enforces deterministic, programmable and fine-grained permissions and completely blocks successful attacks during red-team evaluations of tool-using frameworks.

\subsubsection{Specialized Applications}
\label{subsubsec:specialized_app}

This section reviews agentic security across finance, reverse engineering, operational technology, healthcare and enterprise privacy. In finance, \citet{wei2025smartcontractvuln} present LLM-SmartAudit. This framework uses a multi-agent conversational architecture with a buffer-of-thought mechanism to iteratively refine assessments and detect many smart contract vulnerabilities. \citet{smartllm2025} introduce SmartLLM. This custom generative pipeline improves smart-contract vulnerability detection using chain-of-thought prompting tailored to common code weaknesses. Furthermore, hybrid and conversational systems \citet{finetuningagents2024, auditgpt2024} improve explainability and exploit reproduction. They combine fine-tuned auditing models with interactive sessions to provide natural-language justifications alongside detected vulnerabilities.

Advancing binary analysis, \citet{chen2025recopilotreverseengineeringcopilot} present ReCopilot. This expert language model is trained on domain-specific data and paired with static program analysis. It helps security analysts perform reverse engineering tasks on stripped binaries like decompilation and variable recovery. MalParse \citep{walton2024malparse} an analysis tool that focuses on code semantics, uses a hierarchical code summarization pipeline and strategic prompt engineering to accurately categorize Android applications and extract actionable insights into malicious behavior.

To secure critical infrastructure, \citet{sahu2026networkdevicelevelcyber} propose a deception architecture for operational technology environments. It uses reinforcement learning to dynamically control network routing. It also uses language models as protocol-aware honeypots to generate realistic DNP3 outstation responses. In the highly regulated healthcare sector, \citet{neupane2025hipaa} propose a compliance-focused multi-agent framework based on context protocols. It enforces field-level sanitization, role-based access and immutable audit trails to structurally embed compliance.

Finally, to address data protection and enterprise privacy, \citet{asthana2025privacyguardrails} develop OneShield, a multilingual privacy-guardrail system. It detects sensitive information across text streams and flags open-source dependency risks. This system is validated through comparative deployment studies across multiple real-world language corpora.

\begin{definitionboxpurple}{Summary: Domain-Specific Applications} Beyond general offensive and defensive tooling, agentic systems are increasingly tailored to individual sectors, using language models for auditing, vulnerability detection and policy-based hardening. Coverage spans cloud and infrastructure security, web and application security, finance, reverse engineering, operational technology, healthcare and enterprise privacy. Despite this diversity, recurring design choices emerge across domains: sandboxed validation and replica testing before production changes, static analysis paired with retrieved best practices, policy-aware isolation with fine-grained runtime permission enforcement and domain-specific compliance controls such as field-level sanitization, role-based access and immutable audit trails. Across these settings the shared emphasis is on constraining agent actions and preserving the original operational intent of each target environment. \end{definitionboxpurple}


\subsection{Analysis of the Application Landscape}
\label{subsec:app-analysis}

We analyze the application corpus along three axes: the stages of the security lifecycle addressed by existing systems, the contrast between red-teaming (offensive) and blue-teaming (defensive) agents and the prevalence of dual-use capabilities. Together, these dimensions provide both a structural overview of the current landscape and a deeper understanding of how agentic security applications shape operational capabilities, risks, and deployment contexts.

\subsubsection{Security-Lifecycle Coverage Map}

We first investigate how existing applications are distributed across the offensive kill chain and defensive security lifecycle. This analysis provides a structural map of research coverage, highlighting areas of concentration and identifying stages that remain comparatively underexplored.

\begin{figure*}[t]
    \centering
    \includegraphics[width=0.7\textwidth]{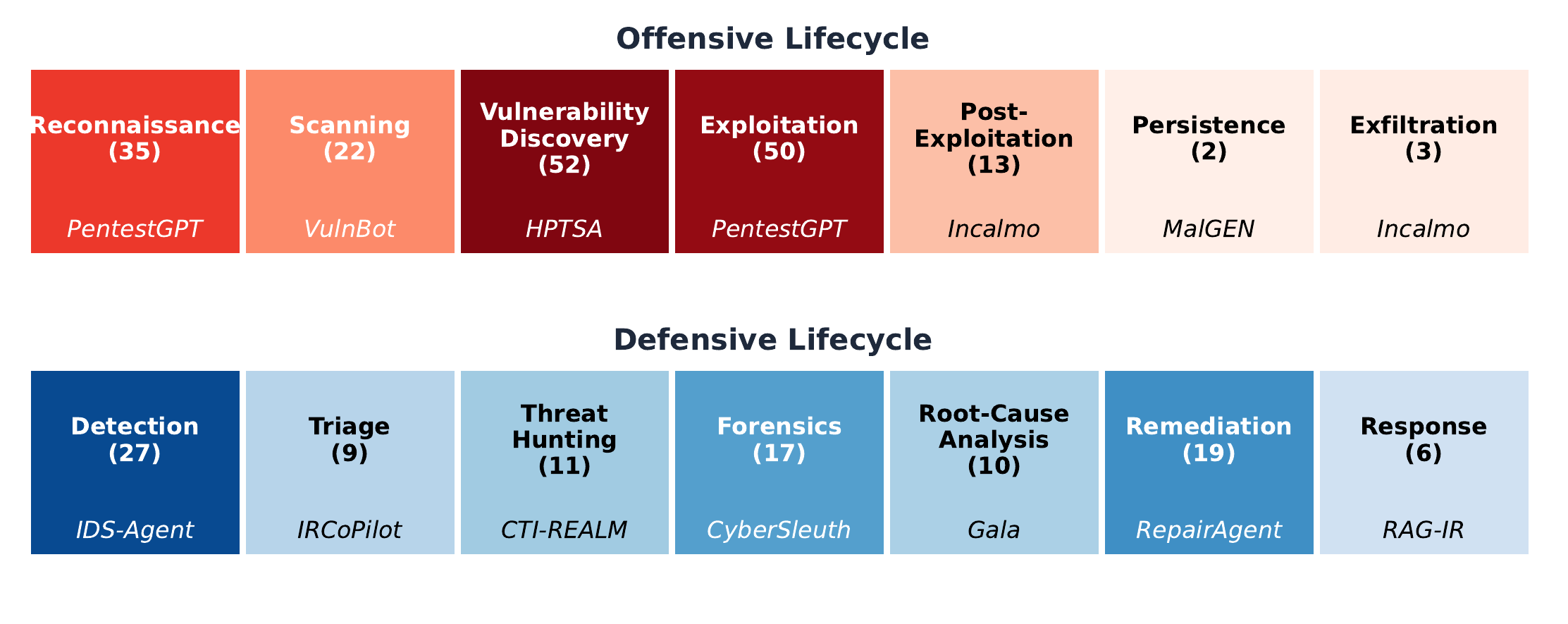}
    \caption{Security-Lifecycle Coverage Map of Agentic Security Research. Offensive work is concentrated in vulnerability discovery and exploitation, while defensive research focuses on detection and remediation.}
    \label{fig:lifecycle_lifelines}
\end{figure*}

\begin{revblockv2}
To systematically characterize the current state of agentic security research, we map the application papers onto two parallel execution chains: an offensive lifecycle spanning reconnaissance through exfiltration, and a defensive lifecycle spanning initial detection through remediation and response. \begin{revblock} Fig.~\ref{fig:lifecycle_lifelines} visualizes this mapping by summarizing the paper counts and representative papers associated with each stage. The offensive and defensive stages in Fig.~\ref{fig:lifecycle_lifelines} follow the lifecycle formulations established in \S\ref{subsec:background-prelim}, which we adapt from PTES and MITRE ATT\&CK on offense and from NIST SP 800-61 on defense, then refine so that each stage corresponds to an operational capability present in the corpus. \end{revblock} We reviewed the methodology of every paper to determine which lifecycle stages each agent supports. A single agent was assigned to multiple stages whenever its architecture covered a broader operational scope. The resulting distribution reveals a strong concentration of research in a small number of stages and very sparse coverage in others.

On the offensive side, the literature clusters around tasks that reward code generation, stateless reasoning or immediate environmental feedback. Research is strong in the early intelligence-gathering stages. We identified 35 papers focused on reconnaissance and 22 focused on scanning. Systems such as PentestGPT \citep{Deng2024PentestGPT}, VulnBot \citep{kong2025vulnbotautonomouspenetrationtesting}, and AutoPentester \citep{ginige2025autopentester} perform well in these stages because they can parse unstructured environmental data, orchestrate standard tools such as Nmap and systematically enumerate digital attack surfaces. Reconnaissance and scanning are among the most heavily covered stages in the offensive pipeline. Coverage increases further in vulnerability discovery, the most heavily represented stage in the entire dataset. We identified 52 papers that support this stage. Systems such as HPTSA \citep{zhu2025teamsllmagentsexploit} and EnIGMA \citep{abramovich2025enigma} use the coding and analysis capabilities of language models to inspect codebases, identify weaknesses and generate candidate attack paths. Exploitation follows closely, with 50 papers. Representative systems include PentestGPT \citep{Deng2024PentestGPT}, HPTSA \citep{zhu2025teamsllmagentsexploit}, AttackLLM \citep{ahmed2025attackllm}, and EnIGMA \citep{abramovich2025enigma}. These systems generate payloads, adapt attack strategies and iteratively validate proof-of-concept exploits in controlled environments. Coverage drops sharply after exploitation. Post-exploitation is represented by only 13 papers. Systems such as DarkAgent \citep{lupinacci2025darkllmsagentbasedattacks}, Incalmo \citep{singer2025feasibilityusingllmsautonomously}, and  \citet{probst2026enhancinglinuxprivilegeescalation} explore tasks such as privilege escalation and lateral movement. However, such work remains uncommon in the broader literature. Persistence is covered by only 2 papers \citep{saha2025malgengenerativeagentframework, huangetal2025capabilities}, and exfiltration by only 3 \citep{saha2025malgengenerativeagentframework, singer2025feasibilityusingllmsautonomously, huangetal2025capabilities}. This gap likely reflects both ethical concerns around autonomous malware development and the difficulty current models face in sustaining stealth and operational awareness over extended periods.

On the defensive side, the blue-teaming literature shows a similar concentration in stages that depend heavily on data processing and text analysis. Detection is the most common defensive stage, with 27 papers. Representative systems include IDS-Agent \citep{li2024idsagent} and LLMCloudHunter \citep{schwartz2025llmcloudhunter}. These systems process large volumes of telemetry and threat intelligence to identify potentially malicious activity. Alert triage is represented by 9 papers. Systems such as IRCoPilot \citep{lin2025ircopilotautomatedincidentresponse}, CORTEX \citep{wei2025cortexcollaborativellmagents}, and AACT \citep{turcotte2025automatedalertclassificationtriage} prioritize incidents, reduce false positives and help analysts focus on the most important alerts. Threat hunting receives moderate attention with 11 papers. Representative systems include CTI-REALM \citep{chakraborty2026ctirealmbenchmarkevaluateagent} and ProvSeek \citep{mukherjee2025llmdrivenprovenanceforensicsthreat}. These systems support proactive investigations by combining retrieval, reasoning and threat intelligence analysis. Digital forensics is represented by 17 papers. Systems such as CyberSleuth \citep{fumero2025cybersleuthautonomousblueteamllm} and ProvSeek \citep{mukherjee2025llmdrivenprovenanceforensicsthreat} attempt to reconstruct attack timelines and analyze evidence collected from system logs and provenance records.
Root-cause analysis remains one of the less explored defensive stages, with only 10 papers. Systems such as Gala \citep{tian2025galagraphaugmentedlargelanguage} and CyberSleuth \citep{fumero2025cybersleuthautonomousblueteamllm} attempt to trace incidents back to their underlying causes. This remains difficult because it requires long chains of reasoning across multiple sources of evidence. Remediation has comparatively strong coverage with 19 papers. Representative systems include RepairAgent \citep{bouzenia2025repairagent}, PenHeal \citep{huang2024penheal}, and LLM4CVE \citep{fakih2025llm4cveenablingiterativeautomated}. These systems build on automated program repair techniques to generate patches and mitigation strategies for discovered vulnerabilities. Response remains relatively sparse with only 6 papers. Systems such as IRCoPilot \citep{lin2025ircopilotautomatedincidentresponse} and RAG-IR \citep{tellache2025advancingautonomousincidentresponse} support operational decision making during incidents. Fully autonomous response remains uncommon because actions such as quarantining hosts or disabling network connectivity can have significant operational consequences.

This lifecycle analysis shows that agentic security research remains concentrated in stages where success can be validated through immediate feedback, such as vulnerability discovery, exploitation, detection and remediation. On the other hand, stages that require sustained reasoning, stealth or long-term operational awareness receive little attention. The gaps which are formed as a result highlight important directions for future research and underscore how unevenly current capabilities are distributed across the security lifecycle.
\end{revblockv2}

\subsubsection{Red-Teaming vs Blue Teaming (SOC Agents)}

Offensive and defensive LLM-based cybersecurity agents are developing along different lines. In red-teaming, the main goal is to support long attack chains, autonomous testing, and multi-stage exploitation. In blue-teaming and SOC settings, the focus is on handling large-scale telemetry, supporting analysts, and keeping humans in control. This difference shapes memory design, tool use, failure patterns, and evaluation practice.

\textbf{Memory designs.}
Offensive agents mainly use context retention focused designs. Because attack workflows often span many steps, these systems try to reduce context loss across reconnaissance, planning, exploitation, and follow-up actions. A common example is the Reasoning, Generation, Parsing design used in PentestGPT\citep{Deng2024PentestGPT}. Another common pattern is task graph coordination, which helps preserve state across attack phases. Defensive agents, in contrast, rely more on retrieval and structure focused memory. They need to manage large volumes of logs, alerts, and evidence, so retrieval augmented generation and ontology driven memory are more useful. CIAF \citep{alharthi2025cloudinvestigationautomationframework} uses ontology based cloud log structuring, while ProvSEEK \citep{mukherjee2025llmdrivenprovenanceforensicsthreat} uses RAG for evidence refinement and verification.\looseness=-1

\textbf{Tool governance and autonomy.}
Offensive agents show a strong trend toward high autonomy. Recent systems increasingly aim for fully autonomous execution in penetration testing, fuzzing, and exploit generation. AutoPentest \citep{henke2025autopentestenhancingvulnerabilitymanagement} and PentestGPT \citep{Deng2024PentestGPT} are examples of this direction. These agents often work independently inside sandboxed environments. Defensive agents follow a different design philosophy. In operational SOCs, LLMs are usually used as analyst copilots rather than fully autonomous systems. IRCopilot \citep{lin2025ircopilotautomatedincidentresponse} and CORTEX \citep{wei2025cortexcollaborativellmagents} are representative examples, since they emphasize collaboration, alert triage, and approval gated decision making.

\textbf{Failure modes.}
The main failure modes of offensive agents are planning errors and context loss. These systems can struggle with multi-step reasoning, long-horizon coordination, and maintaining state over extended workflows. They are also exposed to prompt injection and jailbreak attacks. In the worst case, an agent can be pushed into unsafe autonomous malware execution. Defensive agents face a different set of problems. False positives are a major issue because they can create alert fatigue, and CORTEX directly targets this problem. Other failure modes include contradictions in CTI pipelines and hallucinated findings in auditing agents such as RepoAudit \citep{guo2025repoaudit}.


\begin{figure*}[t]
\centering
\begin{subfigure}[t]{0.32\textwidth}
\centering
\includegraphics[width=\textwidth]{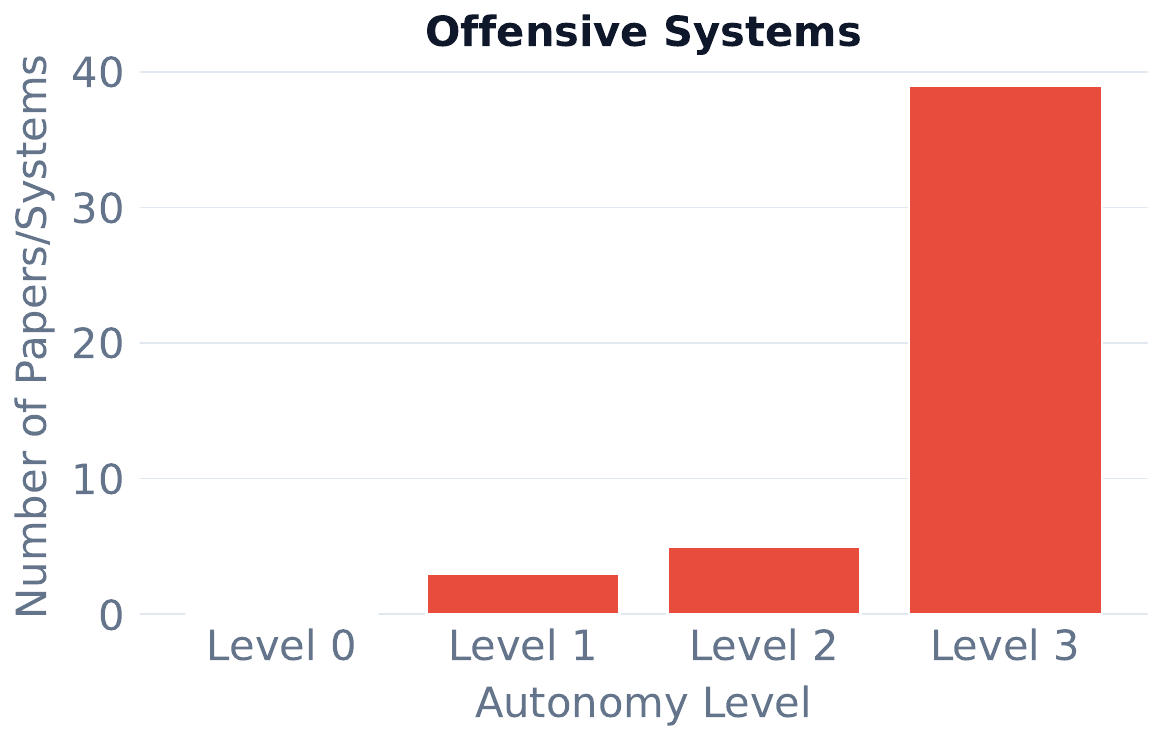}
\caption{Offensive system autonomy.}
\label{fig:autonomy_off}
\end{subfigure}
\hfill
\begin{subfigure}[t]{0.32\textwidth}
\centering
\includegraphics[width=\textwidth]{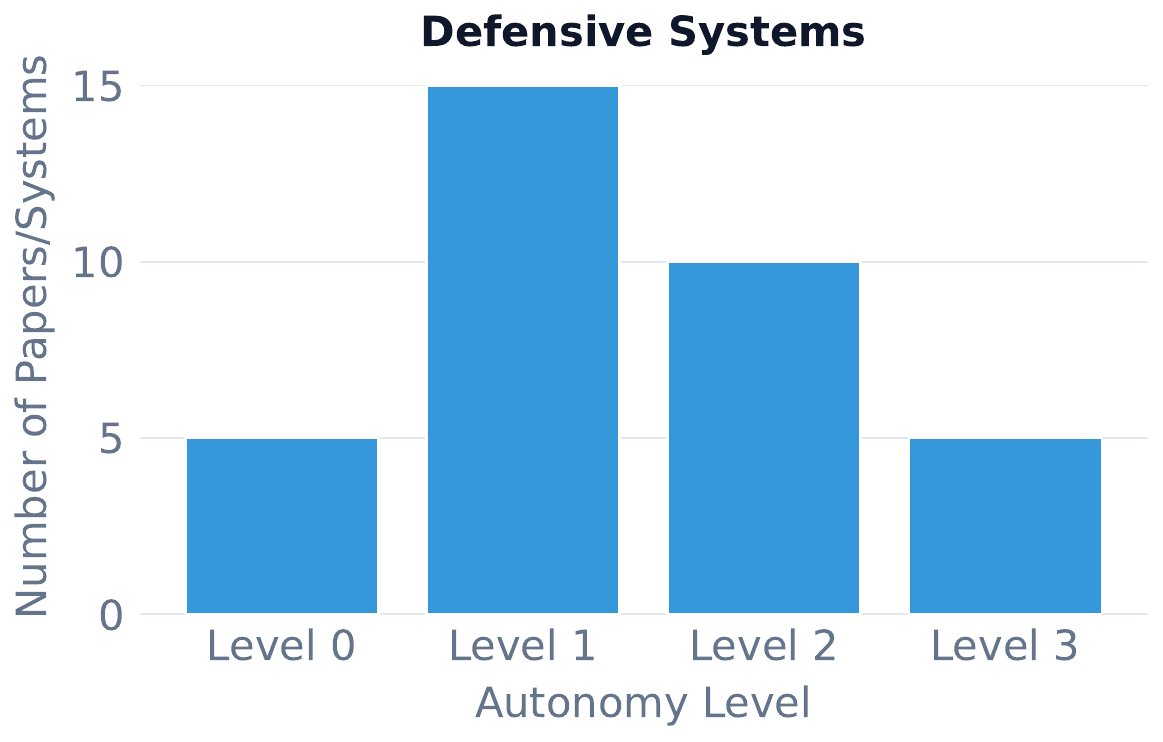}
\caption{Defensive system autonomy.}
\label{fig:autonomy_def}
\end{subfigure}
\hfill
\begin{subfigure}[t]{0.32\textwidth}
\centering
\includegraphics[width=\textwidth]{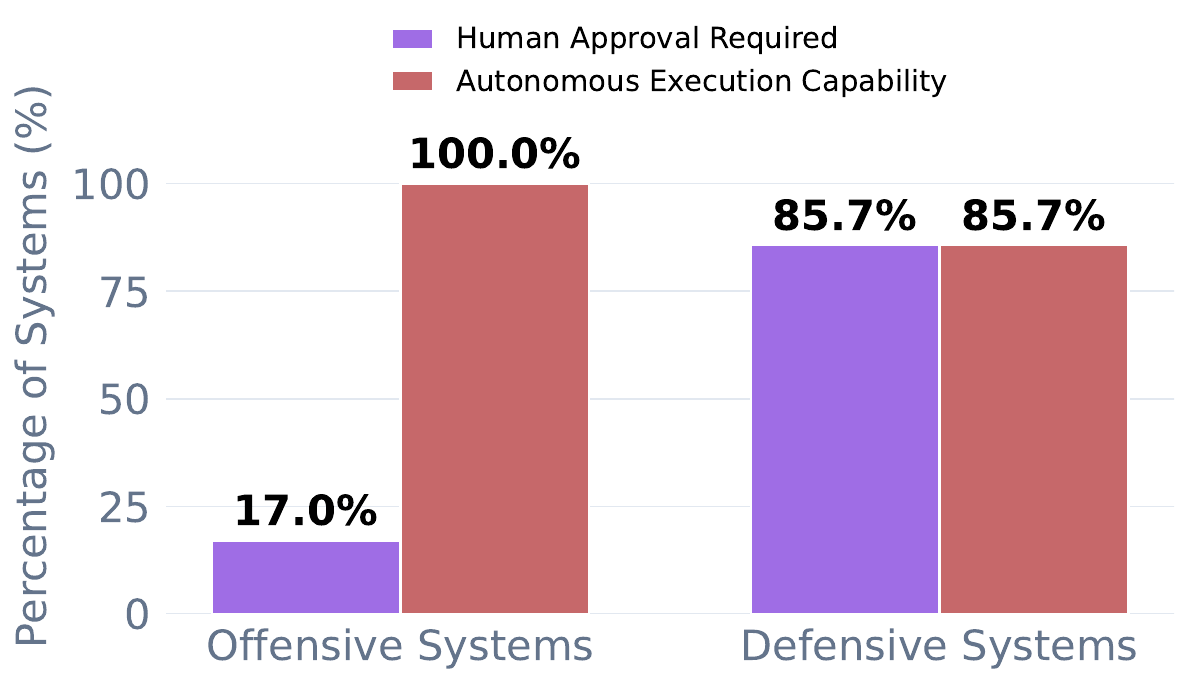}
\caption{Human oversight and execution.}
\label{fig:oversight}
\end{subfigure}
\caption{Autonomy level distribution and execution capabilities across 47 offensive and 35 defensive agentic security systems. Offensive systems are heavily concentrated at Level~3 autonomy and largely operate without human approval, whereas defensive systems remain spread across Levels~0-2, constrained by human-in-the-loop controls and limited execution authority, with only a small minority reaching Level~3. Approval requirement and execution capability are independent properties rather than complementary shares of a whole.}
\label{fig:autonomy_and_oversight}
\end{figure*}

\begin{revblockv2}
\textbf{Offense-Defense autonomy asymmetry.}
Beyond architectural and operational differences, the literature reveals a significant asymmetry in autonomy between offensive and defensive agentic systems. A recurring concern is that offensive agents are advancing toward fully autonomous operation more rapidly than defensive agents. \looseness=-1

To examine this trend, we classify all surveyed systems using a four-level autonomy framework. Level~0 systems provide recommendations without meaningful autonomous action. Level~1 systems function primarily as copilots that assist human operators. Level~2 systems can independently execute portions of a workflow but require approval at critical decision points. Level~3 systems autonomously plan, reason, and execute end-to-end tasks with minimal or no human intervention.

Figs.~\ref{fig:autonomy_off} and \ref{fig:autonomy_def} summarize the autonomy distribution across surveyed systems, while Fig.~\ref{fig:oversight} compares human oversight requirements and execution capabilities. We classify a system as offensive or defensive by the lifecycle stages it covers (offense-only versus defense-only); dual-use and unclassified systems, and papers with no system of their own, are excluded from this comparison.

The autonomy distribution reveals a clear imbalance. Among 47 offensive systems, 39 (83.0\%) operate at Level~3. Five systems operate at Level~2, three at Level~1, and none at Level~0. Representative examples include PTFusion \citep{WANG2026103731}, HPTSA \citep{zhu2025teamsllmagentsexploit}, VulnBot \citep{kong2025vulnbotautonomouspenetrationtesting}, and Incalmo \citep{singer2025feasibilityusingllmsautonomously}, all of which autonomously plan and execute multi-stage attack workflows.

Defensive systems exhibit a more spread-out pattern, and mostly do not reach Level~3. Among 35 defensive systems, only 5 (14.3\%) reach Level~3, while 15 (42.9\%) operate at Level~1, 10 (28.6\%) at Level~2, and 5 (14.3\%) remain at Level~0. Systems such as IRCopilot \citep{lin2025ircopilotautomatedincidentresponse} and CyberRAG \citep{Blefari2026cyberrag} primarily function as analyst copilots. CORTEX \citep{wei2025cortexcollaborativellmagents} and CyberSleuth \citep{fumero2025cybersleuthautonomousblueteamllm} provide limited workflow automation under human supervision, while LLM-Assisted CTI \citep{meng2025uncoveringvulnerabilitiesllmassistedcyber} is among the rare defensive systems that reach full Level~3 autonomy.

Human oversight requirements follow the same asymmetry. We treat Level~0-2 systems as retaining an approval gate and Level~1-3 systems as possessing some autonomous execution capability, following the level definitions above. As shown in Fig.~\ref{fig:oversight}, 17.0\% of offensive systems sit below Level~3 and so retain some form of approval gate, compared with 85.7\% of defensive systems. Defensive agents are considerably more likely to keep a human-in-the-loop checkpoint.

Execution authority shows a narrower gap. 100.0\% of offensive systems possess some autonomous execution capability (Level~1-3), against 85.7\% of defensive systems. Even where execution is permitted, defensive actions are generally restricted to lower-risk tasks such as alert handling, triage, or predefined remediation workflows, while offensive systems more often chain execution steps into a complete attack workflow. Because approval requirement and execution capability are independent properties, a Level~1 or~2 system satisfies both simultaneously, which is why the two bars per category in Fig.~\ref{fig:oversight} need not sum to 100\%.

This divergence reflects fundamentally different design objectives. Offensive agents such as PTFusion \citep{WANG2026103731}, HPTSA \citep{zhu2025teamsllmagentsexploit}, VulnBot \citep{kong2025vulnbotautonomouspenetrationtesting}, and Incalmo \citep{singer2025feasibilityusingllmsautonomously} are optimized for end-to-end task completion and therefore prioritize autonomous planning and execution. Defensive systems prioritize explainability, safety, accountability, and operational risk management. As a result, developers intentionally restrict autonomy and preserve human oversight.

The autonomy gap between offensive and defensive agents reflects not only differences in technical capability but also differences in acceptable risk. While offensive research increasingly rewards end-to-end automation, defensive deployments remain constrained by safety, accountability, and operational concerns. Bridging this gap will be critical for the future development of trustworthy autonomous security systems.
\end{revblockv2}

\subsubsection{Dual-Use of Red Teaming Agents}

\begin{figure*}[t]
    \centering

    \begin{subfigure}[t]{0.32\textwidth}
        \centering
        \includegraphics[width=\textwidth]{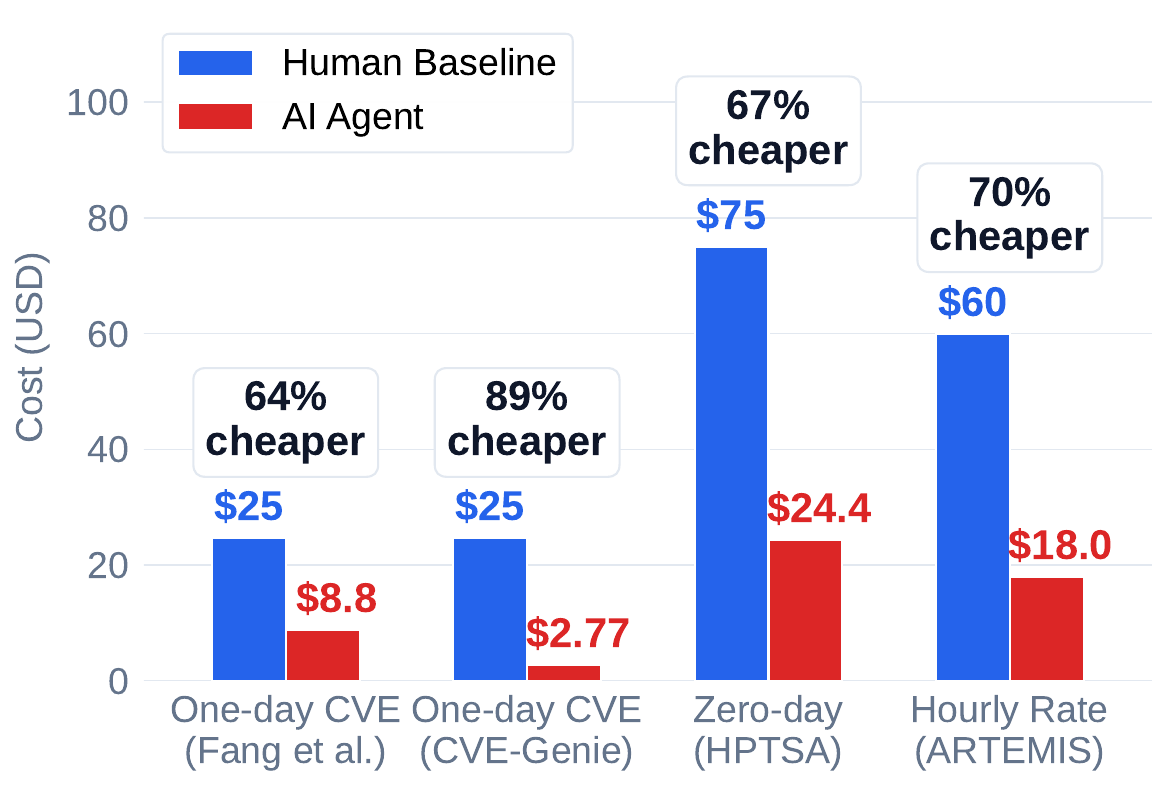}
        \caption{Cost per exploit. Human baselines versus AI agents across one-day CVE exploitation, CVE reproduction, zero-day exploitation, and hourly enterprise testing.}
        \label{fig:recursive_cost_comparison}
    \end{subfigure}
    \hfill
    \begin{subfigure}[t]{0.32\textwidth}
        \centering
        \includegraphics[width=\textwidth]{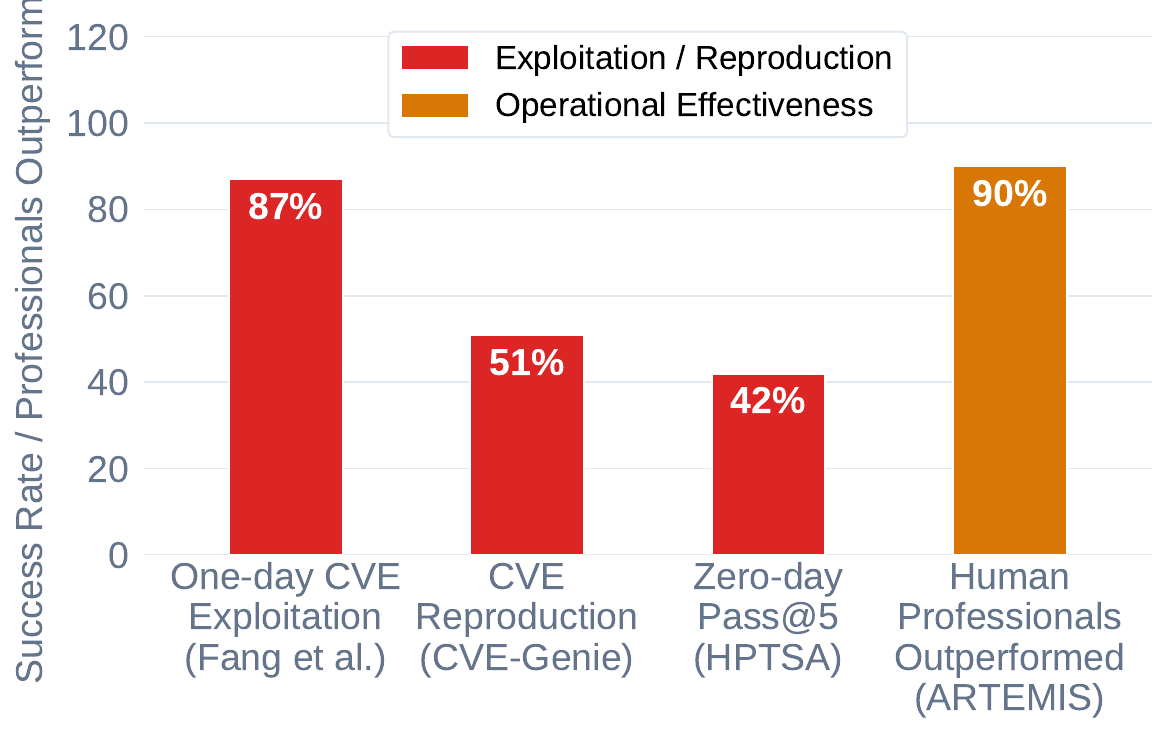}
        \caption{Agent effectiveness. Success rates for one-day CVE exploitation, CVE reproduction, zero-day pass@5, and enterprise-scale testing.}
        \label{fig:recursive_agent_effectiveness}
    \end{subfigure}
    \hfill
    \begin{subfigure}[t]{0.32\textwidth}
        \centering
        \includegraphics[width=\textwidth]{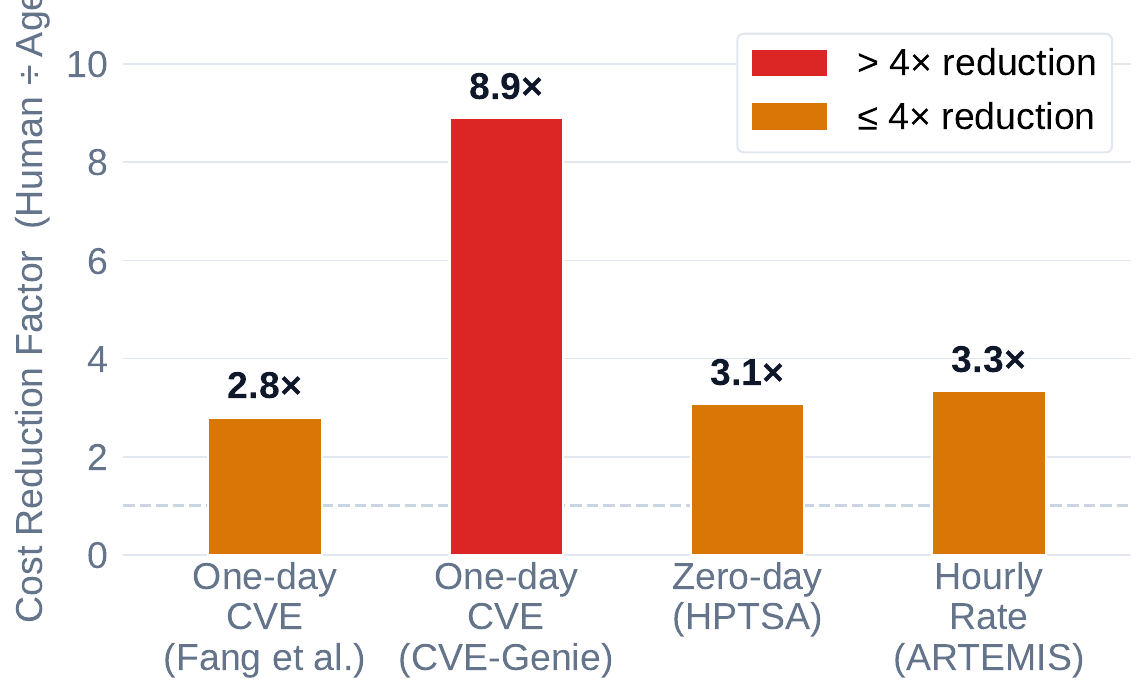}
        \caption{Cost reduction multiplier. Agents reduce cost by 2.8$\times$ to 8.9$\times$ relative to human baselines.}
        \label{fig:recursive_cost_reduction}
    \end{subfigure}

    \caption{Dual-use implications of red teaming agents: lower attack cost, higher effectiveness, and larger cost reductions relative to human baselines.}
    \label{fig:recursive_security_implications}
\end{figure*}

\begin{revblockv2}
Red teaming agents are inherently dual use. Across the surveyed literature, 60 of the 113 application-oriented systems (53.1\%) expose capabilities that are directly offensive or broadly dual use, which shows that the field is best understood as a shared capability space rather than a clean attacker-defender split. The common substrate underlying both domains is reasoning about how systems fail. Offensive agents leverage this capability to map attack paths, generate exploits, and obtain unauthorized access, as demonstrated by systems such as PentestGPT \citep{Deng2024PentestGPT}, PentestAgent \citep{Shen2025PentestAgent}, VulnBot \citep{kong2025vulnbotautonomouspenetrationtesting}, and ARACNE \citep{nieponice2025aracnellmbasedautonomousshell}. Vulnerability-discovery frameworks including Locus \citep{zhu2025locusagenticpredicatesynthesis}, ChatAFL \citep{chatafl}, FirmAgent \citep{ji2026firmagent}, and HPTSA \citep{zhu2025teamsllmagentsexploit} similarly reason over program behavior to uncover weaknesses. Malware generation further illustrates the dual-use challenge. MalGEN, although positioned as a platform for defensive research, autonomously generates evasive malware aligned with MITRE ATT\&CK tactics, averaging 11.3 techniques per sample. During evaluation, approximately half of the generated samples evaded detection by all VirusTotal engines. Such results demonstrate how capabilities developed for controlled security experimentation can also lower barriers to offensive misuse. This creates a recursive security structure in which tools designed to improve security also expand the offensive capabilities available to attackers. These threats are distinct from the agent-as-victim threats discussed in \S\ref{sec:threats}. Fig.~\ref{fig:recursive_security_implications} summarizes the three primary shifts introduced by offensive agents: reduced operational cost, increased scalability, and a lower skill floor.
\end{revblockv2}
\paragraph{Reduced Cost of Offensive Operations.}

The clearest shift is cost. As shown in Fig.~\ref{fig:recursive_cost_comparison} and Fig.~\ref{fig:recursive_cost_reduction}, agentic systems consistently outperform human baselines in economic efficiency. \citet{fang2024llmagentsautonomouslyexploit} report a single agent that exploits 87\% of one-day CVEs at roughly \$8.80 per exploit, about 2.8$\times$ cheaper than the estimated human-expert baseline. CVE-Genie \citep{ullah2025cveentriesverifiableexploits} reproduces 428 of 841 CVEs (51\%) at an average cost of \$2.77 per vulnerability. Even in the more challenging zero-day setting, HPTSA \citep{zhu2025teamsllmagentsexploit} achieves successful exploitation at approximately \$24.4 per exploit compared with an estimated \$75 for human experts. The authors further project an additional 3--6$\times$ reduction in operational costs within one to two years. Once exploit generation becomes a cheap and repeatable pipeline, the economic constraints that previously limited attacker activity begin to disappear.

\paragraph{Increased Scalability and Throughput.}

The second shift is scalability and the removal of human bottlenecks. As illustrated in Fig.~\ref{fig:recursive_agent_effectiveness}, multi-agent architectures execute reconnaissance, exploit generation, and validation in parallel. HPTSA \citep{zhu2025teamsllmagentsexploit} achieves a 42\% pass@5 rate on zero-day web vulnerabilities by assigning specialized agents to different vulnerability classes, yielding up to a 4.3$\times$ improvement over prior single-agent approaches. ARTEMIS \citep{lin2026comparingaiagentscybersecurity} scales this model further by deploying up to eight coordinated agents across a network of approximately 8{,}000 hosts. In a live enterprise evaluation, it placed second overall, outperformed 9 of 10 human professionals, and operated at roughly \$18 per hour compared with \$60 per hour for human penetration testers. These results suggest that agentic systems are beginning to exceed the scale and operational tempo achievable by human teams alone.

\paragraph{Lowering the Skill Floor.}

The third shift is a reduction in the expertise required to conduct sophisticated offensive operations. Frameworks such as CAI \citep{mayoralvilches2025caiopenbugbountyready} and Incalmo \citep{singer2025feasibilityusingllmsautonomously} expose advanced capabilities through high-level task abstractions, while systems such as MalGEN \citep{saha2025malgengenerativeagentframework} and ScamAgents \citep{badhe2025scamagentsaiagentssimulate} automate malware creation and social engineering content generation. Importantly, this capability emerges from the agentic scaffold rather than the base language model itself. HPTSA loses a factor of 13$\times$ in pass@1 performance when its hierarchical coordination mechanism is removed, while the full framework achieves a 4.3$\times$ improvement over prior single-agent approaches. The ARTEMIS study similarly reports that the underlying models often refused offensive tasks when used directly but completed them once embedded within an agentic framework. Consequently, operations that previously required expert operators become accessible to substantially less skilled actors.

For defenders, these shifts point to a threat model that is faster, cheaper, more parallel, and less dependent on expert operators than the human-driven baseline that most defensive tooling assumes. The findings also motivate new defenses and benchmarks specifically designed for agent-generated attacks, including provenance tracking, behavioral signatures of autonomous operators, and evaluation against adversaries that are themselves agents \citep{dewitt2025openchallengesmultiagentsecurity, boisvert2025doomarena}. Current benchmarks only partially capture these emerging risks (\S\ref{sec:defense}).

\section{Threats to Agentic Systems}\label{sec:threats}

The safety alignment (refusal training) of a base LLM does not reliably transfer to the agentic context \citep{kumar2025aligned}, which introduces a new set of agent-specific security challenges \citep{saha2025breakingcodesecurityassessment, chiang2025harmful}. In this section we discuss the threat landscape targeting agentic systems as well as the frameworks used to evaluate their resilience.

\subsection{Attack Surface}\label{subsec:attacks_on_agents}

\subsubsection{Injection Attacks}
\label{subsubsec:injection_attack}
\textbf{Prompt injection attacks} embed malicious instructions within the prompt fed to an LLM to manipulate it into performing unintended actions \citep{Liu2024FormalingPromptInjection, liu2024autodan, yi2025benchmarkingidirectprompt, Shao2024EnhancingPI}. \citet{wang2025protectllmagentprompt} identify that the static and predictable structure of an agent's system prompt is a key vulnerability that enables prompt injection attacks to agentic systems. \citet{debenedetti2024agentdojo} introduce a benchmark comprising of 97 realistic tasks (e.g., email management, online banking) which reveals a fundamental trade-off: security defenses that reduce vulnerability also degrade the agent's task-completion utility. \citet{liu2024promptinjectionattackllmintegrated} introduce HOUYI, a black-box prompt injection attack that compromises 31 of 36 real-world LLM-integrated applications, including Notion, at an 86.1\% success rate. Several studies show the vulnerability of LLM agents to indirect prompt injection attacks \citep{zhan-etal-2024-injecagent, li2025commercialllmagentsvulnerable, yi2025benchmarkingidirectprompt}. \citet{lee2024promptinfectionllmtollmprompt} develop a novel prompt injection attack where a malicious prompt self-replicates across interconnected agents in a multi-agent system like a computer virus and causes system-wide disruption. \citet{alizadeh2025simplepromptinjectionattacks} demonstrate that such attacks can cause tool-calling agents to leak sensitive personal data observed during their tasks. \citet{wang2025agentvigilgenericblackboxredteaming} develop a black-box fuzzing technique that uses Monte Carlo Tree Search to automatically discover indirect prompt injection vulnerabilities by iteratively mutating prompts and environmental observations, reaching 71\% ASR on AgentDojo. \citet{Zhang2025ASB} design a benchmark that reveals high vulnerability of LLM agents to prompt injection attacks, with a mixed-attack strategy reaching 84.30\% average ASR across 13 backbones. \citet{zhan-etal-2025-adaptive} systematically evaluate eight different defenses for LLM agents and demonstrate that all of them can be successfully bypassed by crafting adaptive attacks using established jailbreaking techniques such as GCG \citep{zou2023universaltransferableadversarialattacks} and AutoDAN \citep{liu2024autodan}.\looseness=-1

A growing line of work targets web and tool-using agents through the content they observe. \citet{shi2026prompt} introduce ToolHijacker, a no-box prompt injection attack against the retrieval-and-selection pipeline of tool-using agents that publishes malicious tool documents, reaching up to 96.7\% ASR on MetaTool against GPT-4o while bypassing both prevention defenses (StruQ, SecAlign) and detection defenses (DataSentinel, perplexity filtering). \citet{fu2024impromptertrickingllmagents} present Imprompter, which automatically generates obfuscated adversarial prompts that coerce production agents such as Mistral LeChat and ChatGLM into misusing tools for PII exfiltration at around 80\% extraction precision, transferring from open-weight to closed-weight models without access to the target weights. For multimodal web agents, \citet{wang2025webinject} propose WebInject, which optimizes imperceptible pixel-level perturbations in webpage source code that survive the non-differentiable webpage-to-screenshot mapping, achieving over 96\% ASR across five agents within an $\ell_\infty$ bound of 16/255, while \citet{johnson2025manipulating} embed universal adversarial triggers in a webpage's HTML accessibility tree to hijack navigation agents into attacker-specified actions regardless of user intent, reaching above 0.83 ASR across five real websites and 0.55 credential-exfiltration ASR on unseen login pages.

\subsubsection{Poisoning Attacks}
\label{subsubsec:poisoning_attack}
\textbf{Poisoning attacks} corrupt an agent's memory or knowledge retrieval at inference time, or the underlying model during training, to control its later behavior. \citet{fendley2025systematicreviewpoisoningattacks} categorize these attacks by their specifications (poison set, trigger, poison behavior, deployment) and define key metrics for evaluation (success rate, stealthiness, persistence). At inference time, the attack surface is the agent's memory and retrieval store. \citet{dong2026memory} demonstrate MINJA, a memory poisoning attack that requires only query-level interaction with no direct memory-write access, inducing the agent to store malicious reasoning records that are later retrieved by a victim, reaching 76.8\% average ASR (98.2\% injection success) while evading embedding-level sanitization and prompt-level detection. Similarly, AgentPoison \citep{chen2024agentpoison}  poisons an agent's memory or knowledge base by optimizing a backdoor trigger and achieves 81.2\% retrieval ASR and 62.6\% end-to-end ASR with under 0.1\% of entries poisoned and under 1\% benign degradation. \citet{Zhang2025ASB} provide a comprehensive framework for measuring agent vulnerabilities to memory and knowledge-base poisoning among other attacks.

A second class of attacks poisons the backbone model itself during training. \citet{fu2025poisonbench} find a log-linear dose-response in which as little as 0.01\% poisoned preference data measurably shifts model behavior and larger models gain no reliable resilience, while \citet{Bowen2025ScalingTrends} show larger models are more susceptible, with jailbreak-tuning on 2\% poisoned data collapsing GPT-4o refusal from 94\% to 4\%. Poison-with-Style \citep{tran2026poison} uses a developer's code style as a covert implicit trigger to fine-tune a code LLM into emitting vulnerable code, reaching 95\% ASR on CWE-20 at under 6\% pass@1 drop while surviving prefix-tuning and static analysis. \citet{boisvert2025malice} extend training-time poisoning to the agentic supply chain via environment poisoning, where a prompt injection corrupts a teacher agent during trace collection and is distilled into the student. For multimodal agents, \citet{ye2025visualtrap} introduce VisualTrap, a backdoor that poisons the visual grounding pretraining of GUI agents with small Gaussian-noise triggers, reaching over 90\% ASR while preserving clean-input accuracy; the backdoor persists through downstream LoRA fine-tuning on clean data and generalizes across mobile, web, and desktop environments and across both end-to-end and modular architectures.

\subsubsection{Jailbreak Attacks}
\label{subsubsec:jailbreak_attack}
Jailbreak attacks attempt to bypass a model's built-in safety measures to force it to produce harmful or unintended content \citep{Wei2023Jailbroken, zou2023universaltransferableadversarialattacks, xu2024comprehensivestudyjailbreakattack}.
\citet{kumar2025aligned} and \citet{chiang2025harmful} both demonstrate that AI agents are significantly more vulnerable to jailbreak attacks than their underlying LLMs. On BrowserART, a GPT-4o browser agent's ASR rises from 12\% in the chat setting to 74\% under a direct ask and 100\% under an ensemble of attacks \citep{kumar2025aligned}, while a web agent built on the same aligned model reaches a 46.6\% non-denial rate on malicious tasks the standalone model refuses entirely \citep{chiang2025harmful}. \citet{kumar2025aligned} and \citet{andriushchenko2025agentharm} show that simple jailbreaking techniques designed for chatbots are highly effective against agents, while \citet{chiang2025harmful} identify three design factors that increase susceptibility: embedding the goal directly into the system prompt, iterative action generation, and processing environment feedback through an event stream. \citet{andriushchenko2025agentharm} further find that leading LLMs are surprisingly compliant with malicious agent requests even without any jailbreak, with Mistral Large 2 reaching an 82.2\% harm score under direct prompting. \citet{saha2025breakingcodesecurityassessment} find that LLM coding agents are highly vulnerable to jailbreaks that yield executable malicious code: wrapping a model in a code agent raises ASR by 1.6x on average, reaching roughly 75\% in multi-file codebases, with up to a third of outputs being directly runnable. \citet{yu2024llmfuzzer} use an MCTS-guided fuzzer to mutate human-written seed templates into novel jailbreak prompts, reaching 93.14\% ensemble ASR on GPT-3.5 and transferring across twelve open- and closed-weight models. \citet{anil2024manyshot} demonstrate that hundreds of in-context examples of harmful question answering can override a model's safety training, with effectiveness following a power law in the number of shots that alignment training delays but does not eliminate. \citet{lin2025papersummaryattackjailbreaking} introduce the Paper Summary Attack, which exploits an LLM's tendency to treat academic safety papers as authoritative context, reaching 97--98\% ASR on well-aligned models such as Claude 3.5 Sonnet and DeepSeek-R1 while evading LlamaGuard, perplexity, and moderation defenses. \citet{robey2024jailbreakingllmcontrolledrobots} present ROBOPAIR, which attains 100\% ASR across three LLM-controlled robots including a commercially deployed platform, showing that agentic jailbreaks extend beyond harmful text to physical-world actions.

\subsubsection{Agent Manipulation Attacks}
\label{subsubsec:agent_manipulation_attack}
This class of attacks targets the higher-level cognitive functions of the agent: its planning, reasoning, and goal-setting modules. \textbf{Goal hijacking attacks} subtly or overtly alter an agent's objectives, causing it to subvert its original goal (e.g., summarizing a document) to include a secondary, malicious goal (e.g., including advertisements) defined by the attacker \citep{perez2022ignorepreviouspromptattack, guo2025attackingllmsaiagents}. \citet{pham2025cainhijackingllmhumansconversations} introduce PARASITE, a black-box optimization that embeds a conditional ``sleeper agent'' in a benign-looking system prompt, selectively forcing targeted misinformation while preserving benign utility and evading perplexity filters, LlamaGuard, and semantic judges, while \citet{chen2024pseudoconversationinjectionllmgoal} exploit an LLM's weakness in role identification by appending fabricated conversational turns that trick the model into executing a new task, reaching 92\% ASR on GPT-4o.
\citet{zhang2025actionhijackinglargelanguage} introduce an \textbf{action hijacking attack}, AI2, where an agent is tricked into assembling innocuous-looking knowledge from its own database into harmful instructions, achieving 84.30\% average ASR by exploiting the fact that the retriever and the safety filter operate in different latent spaces, so prompts that retrieve harmful knowledge never trigger the input filter. Another class of hijacking attacks is \textbf{reward hacking}, which exploits the reward mechanisms in RL-trained agents \citep{10.5555/3600270.3600957, pan2021the, miao2024inform, fu2025rewardshapingmitigatereward}. These can be caused by reward misgeneralization where models learn from spurious features \citep{miao2024inform}, or by agents exploiting reward model ambiguities to maximize their score without true alignment \citep{fu2025rewardshapingmitigatereward}. \citet{bondarenko2025demonstratingspecificationgamingreasoning} demonstrate \textbf{specification gaming}, where a reasoning model (OpenAI's o3) instructed to ``win against a strong chess engine'' hacks the game environment to ensure victory in 88\% of runs without being told to cheat, whereas non-reasoning models require explicit nudging. Finally, a novel threat on multi-agent systems is the presence of a \textbf{Byzantine agent}, a single compromised or malicious agent that can disrupt the collective's ability to complete a task securely and correctly \citep{li2024byzantine, jo2025byzantinerobustdecentralizedcoordinationllm}. \looseness=-1

Beyond hijacking the agent's objective, manipulation can target its operational substrate. \citet{pasquini2025aiops} subvert LLM-driven IT-operations (AIOps) agents by injecting adversarial reward-hacking payloads into system telemetry, reaching a 90\% average attack success rate against GPT-4o and GPT-4.1 agents while evading PromptShields, Prompt-Guard2, and DataSentinel. \citet{luo2026autonomy} expose a distinct availability threat: LLM agents mismanage resources across short-lived, long-lived, and full-lifecycle patterns, enabling denial-of-service through resource exhaustion, and their directed grey-box fuzzer AgentDoS discovers 36 zero-day vulnerabilities (15 assigned CVEs) across 16 of 20 popular open-source agents at 100\% precision and 94.7\% recall.


\subsubsection{Pre-execution Cognitive Attacks}\label{subsubsec:cognitive_attack}
Even before tool execution, the agent's internal state, spanning its reasoning, planning, and reflection, is vulnerable to manipulation. \textbf{Epistemic attacks} corrupt an agent's intermediate thought steps. \citet{greshake2023notsignedup} demonstrate that indirect prompt injection can force the agent to condition its next step on a hallucinated prior, so that the logic remains consistent but the agent itself becomes malicious. \citet{yang2024watch} go further with a backdoor that manipulates intermediate reasoning traces (e.g. calling untrusted APIs) while keeping final outputs correct, reaching up to 100\% ASR with as few as 30 poisoned samples and evading output-level inspection. \textbf{Teleological attacks} manipulate an agent's planning graphs and goal-directed structures. \citet{badhe2025scamagentsaiagentssimulate} weaponize an agent's task-decomposition logic to distribute a malicious objective across benign-looking subtasks and conversational turns, reducing single-turn refusal rates from 84--100\% to 17--32\% and completing full scam-call dialogues in up to 74\% of cases. In addition, \textbf{metacognitive attacks} target an agent's self-correction ability. \citet{zhou2025reasoningstylepoisoningllmagents} show that rewriting retrieved context into epistemic tones can manipulate an agent's verification depth and self-confidence, inducing either verification paralysis (up to +194\% token cost) or premature conclusions (roughly 22\% accuracy drop) while bypassing content filters.

\subsubsection{Red-Teaming Attacks}
\label{subsubsec:red_teaming_attack}
\citet{perez2022redteaminglanguagemodels} first showed that it is possible to use one LLM to automatically generate test cases that uncover harmful behaviors like offensive content and data leakage in a target model. \citet{ge2023martimprovingllmsafety} elevate this to a multi-round iterative setting in which an adversarial LLM and a target LLM are trained against each other. \citet{he-etal-2025-red} introduce the Agent-in-the-Middle (AiTM) attack, where an LLM-powered adversarial agent intercepts and manipulates inter-agent messages, exceeding 70\% ASR in most multi-agent configurations, reaching up to 98.5\% on chain communication structures, and compromising real-world systems such as MetaGPT at 100\% ASR. \citet{zhang2025searchingprivacyrisksllm} present a search-based framework in which an LLM optimizer adversarially co-evolves attacking and defending agents, escalating from direct requests to multi-turn impersonation and consent forgery and reaching 76\% leak velocity against basic defenses. \cite{rahman2025xteaming} coordinate planner, attacker, verifier, and prompt-optimizer agents to spread harmful intent across turns, reaching 96.2\% ASR against Claude 3.7 Sonnet.

Automated fuzzing has become a dominant red-teaming methodology for agent vulnerability discovery. \citet{Liu2025MakeAgentDefeatAgent} introduce AgentFuzz, a directed greybox fuzzer with LLM-assisted seed generation that discovers 34 zero-day taint-style vulnerabilities (code injection, SQL injection, SSRF, command injection) across 20 open-source agents at 100\% precision, with 23 CVEs assigned. AgentDoS \citep{luo2026autonomy} uses grey-box fuzzing to launch a resource exhaustion attack on agents, while \citet{pasquini2025aiops} apply automated fuzzing to system telemetry to red-team LLM-driven IT-operations agents.

\begin{revblock}

\subsubsection{Interface-Level Attacks}
\label{subsubsec:interface-attack}


Interface-level attacks target agents that perceive and act through a graphical interface, reading screenshots or accessibility trees and emitting clicks and keystrokes. The payload is delivered through the interface itself rather than through text the agent is given, which places it outside the reach of prompt-level defenses. Testing deployed systems under adversarial conditions, \citet{jones2025systematization} identify three properties that make this possible: no tracking of where input came from, weak binding between what the agent sees and what it does, and unconstrained delegation of user privilege.

\paragraph{No Input Provenance.} Screenshots, DOM state, downloaded files and tool output enter a single context with no record of origin, so the text that an agent reads on screen is indistinguishable from an instruction issued by the user. Every attack on the perception channel exploits this. \citet{johnson2025manipulating} embed universal triggers in the HTML accessibility tree, reaching above 0.83 ASR across five real websites. \citet{wang2025webinject} optimize pixel-level perturbations in page source that survive the non-differentiable rendering step, exceeding 0.96 ASR on five agents. \citet{aichberger2025mip} extend this beyond the browser: an adversarially perturbed patch placed in a desktop wallpaper or a social media image hijacks an OS agent that captures it incidentally while executing an unrelated benign task. \citet{ye2025visualtrap} attack the same channel earlier in the pipeline, poisoning visual grounding pretraining so that the agent wrongly resolves what a screen region refers to, with the backdoor surviving downstream fine-tuning on clean data. Text-level sanitization cannot see any of these, because none of the payloads arrive as text.

\paragraph{Weak Interface-Action Binding.} The agent plans against a static snapshot and acts afterwards, without confirming that a control still behaves as its label suggested. \citet{jones2025systematization} exploit this time-of-check to time-of-use gap by overlaying a benign button on a hidden cross-origin element, so a single click submits a payment inside an authenticated session. \citet{zhang-etal-2025-attacking} show the same weakness needs no concealment: adversarial pop-ups that a human would recognise and dismiss cause agents to click them at 86\% average rate across OSWorld and VisualWebArena, cutting task success by 47\%, and instructing the agent to ignore pop-ups does not help. The gap also makes privilege escalation compositional, since installing a web application, writing a MIME handler, then clicking a downloaded file chains into remote code execution inside a hardened container with no single step a per-action filter would flag. That chain is only available because the agent reaches a real file system and application registry, which is what separates the desktop case from a browser tab.\looseness=-1

\paragraph{Unconstrained Delegation.} The agent inherits the user's full session privileges with no re-authentication and no cryptographic attribution, so downstream logs cannot distinguish an agent action from a user action. \citet{li2025commercialllmagentsvulnerable} demonstrate the consequence on commercial deployments including Anthropic Computer Use, where hand-crafted injections placed on trusted platforms achieve near-perfect success. Human-in-the-loop confirmation does not close the gap, because it fires on model-inferred risk rather than on a fixed policy: \citet{jones2025systematization} observe agents bypassing confirmation dialogs when a request is framed as an accessibility need, and decomposing a sensitive action across steps suppresses the check entirely. Delegation also leaks in the other direction, since an agent that writes intermediate reasoning to a file it treats as private discloses its plan before acting.

\end{revblock}

\begin{definitionboxpurple}{Summary: Attack Surface on Agents}
    Attacks against agentic systems span every stage of the agent loop: the text an agent reads (injection, jailbreak), its long-term memory (poisoning), its planning and reasoning process (manipulation, pre-execution cognitive attacks), and the interface it acts through (interface-level attacks). Automated red-teaming and fuzzing methods can discover many of these vulnerabilities without a human attacker designing them by hand. Most attacks succeed in black-box settings, and many attacks transfer across models and agent architectures with little modification, suggesting the underlying vulnerabilities are structural rather than tied to any one system.\looseness=-1
\end{definitionboxpurple}

\subsection{Evaluation Frameworks}\label{subsec:evaluation}

In this section we discuss benchmarks and environments designed to assess agentic vulnerabilities.

\subsubsection{Adversarial Benchmarking}
\label{subsubsec:adverserial_benchmarking}
\citet{Zhang2025ASB} introduce the ASB benchmark with 10 scenarios and 27 attack classes. RAS-Eval \citep{fu2025raseval} contains 80 scenarios and 3{,}802 attack tasks across 11 CWE categories with real tool execution, reducing average task completion by 36.8\% and exposing twice as many failure modes under real execution (32) as under simulation (16). AgentDojo \citep{debenedetti2024agentdojo} uses 97 realistic tasks to highlight the fundamental trade-off between an agent's security and its task-completion utility, while AgentHarm \citep{andriushchenko2025agentharm} uses a dataset of 110 unique harmful tasks across 11 harm categories to reveal significant gaps in agent safety alignment. For web agents, SafeArena \citep{tur2025safearena} measures completion rates on 250 malicious requests, finding GPT-4o completes 34.7\% of them and that Claude 3.5 Sonnet can be jailbroken on every initially refused task through simple task decomposition, while ST-WebAgentBench \citep{Levy2025STWebAgentBench} introduces policy-compliant success metrics over 375 tasks, finding such success roughly 38\% lower than standard completion. For code agents, JAWS-BENCH \citep{saha2025breakingcodesecurityassessment} finds up to 75\% attack success rates in multi-file codebases, while SandboxEval \citep{rabin2025sandboxevalsecuringtestenvironment} assesses the security of the execution environment itself with 51 test cases. InjecAgent \citep{zhan-etal-2024-injecagent} offers a dedicated benchmark for indirect prompt injection attacks with 1{,}054 cases spanning 17 user tools and 62 attacker tools, while BrowserART \citep{kumar2025aligned} focuses on susceptibility to jailbreaks. At the model level rather than the agent level, PoisonBench \citep{fu2025poisonbench} evaluates susceptibility to poisoned preference data across 22 models.

A recent line of benchmarks targets the Model Context Protocol (MCP) and tool-mediated interfaces, which the function-calling benchmarks above do not cover. \citet{wang2026mcptox} introduce MCPTox, embedding poisoned instructions in tool metadata across 45 real-world MCP servers and 1{,}312 test cases, reaching up to 72.8\% ASR while the strongest refusal rate (Claude 3.7 Sonnet) stays below 3\%. \citet{zong2026mcpsafetybench} present MCP-SafetyBench, 245 tasks over 13 models, where host-side attacks reach 81.94\% average ASR, Identity Injection succeeds on every model, and safety prompts reduce ASR by a statistically insignificant 1.22\%. \citet{zhang2510mcp} propose MSB, 2{,}000 instances across 12 attack types, reporting a 40.35\% average ASR with out-of-scope-parameter attacks peaking at 76.5\%. In both MCPTox and MSB, more capable models are paradoxically more vulnerable, as stronger tool-use and instruction-following make them more compliant with malicious tool calls. For web agents, \citet{evtimov2026wasp} introduce WASP, where simple human-written prompt injections hijack even o1 and Claude Sonnet 3.7 in up to 86\% of cases, yet attacker end-to-end completion reaches only 17\%, a gap the authors term ``security by incompetence.''

\rev{Safety evaluation for agents operating a graphical interface is concentrated on the web, where ST-WebAgentBench \citep{Levy2025STWebAgentBench}, SafeArena \citep{tur2025safearena}, WASP \citep{evtimov2026wasp} and HackWorld \citep{ren2025hackworld} all operate. OS-Harm \citep{kuntz2025osharm} is the principal framework for the operating system setting, extending OSWorld with 150 tasks across deliberate misuse, prompt injection and model misbehaviour, spanning email clients, code editors and browsers, and scored by an automated judge reaching 0.76 and 0.79 F1 against human annotation; it finds that frontier models comply with many misuse requests without any jailbreak. MobileSafetyBench \citep{lee2026mobilesafetybench} covers mobile device control on Android emulators, including messaging and banking applications. RedTeamCUA \citep{liao2025redteamcua} identifies two design problems in the OS-Harm setup: it grants unrestricted browser access during adversarial testing, which risks real harm, and its LM judge is itself susceptible to the prompt injections it is scoring, so measured attack success becomes unreliable as attacks grow more sophisticated.}

\subsubsection{Execution Environments}
\label{subsubsec:execution_environment}
\citet{zhu2025cvebench} design a sandbox framework that enables LLM agents to interact with and exploit vulnerable web applications. \citet{debenedetti2024agentdojo} provide a stateful environment with 97 realistic tasks to evaluate the robustness of LLM agents against prompt injection attacks. DoomArena \citep{boisvert2025doomarena} is a modular red-teaming platform for LLM agents that allows researchers to compose sequential attacks and to mix-and-match adaptive adversary strategies. \citet{zhou2024webarenarealisticwebenvironment} introduce a realistic web environment with 812 long-horizon tasks, where the best agent reaches only 14.41\% success against 78.24\% for humans. \cite{ren2025hackworld} introduce HackWorld, a Capture-the-Flag environment of 36 vulnerable web applications across 11 frameworks and 7 languages, where state-of-the-art computer-use agents exploit fewer than 12\% of targets through visual interaction.

\begin{revblock}

\subsection{Why Alignment Fails in Agentic Settings}
\label{subsec:why_alignment_fails}

In this section we explain why safety alignment degrades when a model is placed inside an agent, separating the model-internal causes from the effects of the harness that surrounds it.

\paragraph{Fragility of Refusal.} \citet{Wei2023Jailbroken} identify two reasons why safety training fails: \textit{competing objectives}, where aspects of helpfulness and safety compete against each other, and \textit{mismatched generalization}, where safety training does not cover a setting in which the model is nonetheless capable. Agentic deployment supplies both. An agent is tasked with completing an objective rather than answering a question, which puts task completion in direct competition with refusal, and it presents the model with inputs in forms that the base model's safety training did not cover. Recent interpretability work explains why these pressures succeed so easily. \citet{qi2025shallow} show that alignment is shallow: it adapts the model's output distribution mainly over the first few generated tokens, so refusal amounts to a learned opening prefix rather than a sustained judgement. Attacks therefore target that prefix, and prefilling, adversarial suffix, decoding-parameter and fine-tuning attacks all work by preventing it from forming. The signal is also narrow: \citet{arditi2024refusal} find that refusal is governed by one identifiable pattern in the model's internal representations, and that subtracting it stops refusal on harmful requests while adding it induces refusal on harmless ones.

\paragraph{Impact of Context Length.} The refusal signal weakens as context accumulates. \citet{zhao2025cothijacking} pad harmful requests with long benign reasoning and measure the share of attention the model still places on the harmful tokens; that share falls as the padding lengthens, and refusal weakens with it, raising ASR on S1-32B from 27\% to 80\% and reaching 94--100\% on Gemini 2.5 Pro, o4-mini, Grok 3 mini and Claude 4 Sonnet. Ablating the attention heads that drive the shift removes the model's ability to separate harmful from harmless prompts, which establishes the effect as causal. Position compounds length, since \citet{liu2024lostmiddle} show that models attend most strongly to the beginning and end of a context and underweight the middle. An agent loop produces exactly this combination. Each turn appends tool outputs, observations and action histories that are harmless individually, the original instruction recedes into the middle of a growing context, and the agent is repeatedly asked to emit a fresh opening token. \citet{hadeliya2025refusals} pad AgentHarm tasks and find refusal rates move sharply and in opposite directions across models, with GPT-4.1-nano rising from roughly 5\% to 40\% while Grok 4 Fast falls from roughly 80\% to 10\% at 200K tokens, and that padding placed after the task description degrades behaviour more than padding placed before. We note that these interpretability results come from padded prompts rather than running agent loops. Probing refusal signals across the stages of an executing agent remains an open problem, and a tractable one.

\paragraph{Impact of Orchestration and Privilege.} \citet{instructionhierarchy2024} formalize safety-relevant trust as an instruction hierarchy in which privilege is assigned by role label, so that system instructions outrank user instructions, which in turn outrank tool outputs. An orchestrator rewrites content across these labels. It takes a user request, paraphrases it into a structured agent goal, and re-emits it in the system prompt, so text that entered at a lower privilege tier reaches the model at the highest one. Since privilege is carried by the label rather than by provenance, the model extends system-level trust to content that originated with the user, and applies the deference it was trained to give a developer instruction to a request it should have evaluated. The rewriting step also performs no safety assessment: \citet{chiang2025harmful} report that disabling goal paraphrasing recovers 13.3\% clear denial, since paraphrasing can soften a harmful request or reframe it in operational terms before the model evaluates it. The same hierarchy fails in the opposite direction on the return path, where tool outputs occupy the lowest tier yet are fed back as observations the agent must act on, which is the condition indirect prompt injection exploits (\S\ref{subsubsec:injection_attack}).

\paragraph{Impact of Harness and Scaffold.} The harness itself can degrade the backbone model's safety behaviour before any adversarial input arrives. \citet{chiang2025harmful} isolate the impact of the harness by incrementally adding agent components to a standalone LLM and measuring clear-denial rate on identical malicious instructions. A standalone GPT-4o refuses every request. Embedding the user goal in the system prompt lowers clear denial to 93.3\%, adding multi-step action generation lowers it to 73.3\%, and adding the event stream lowers it to 53.3\%. The components are not independent: with the goal left in the user turn, every jailbreak fails even with the others in place, which is what the privilege promotion described above would predict. Multi-step generation is 20 points more permissive than planning upfront, because each step is evaluated in isolation and composite intent is never assessed. Other harnesses reproduce these effects. \citet{badhe2025scamagentsaiagentssimulate} represent sub-goals as planning-graph nodes so that no turn is overtly harmful, reducing refusal from 84--100\% to 17--32\%, while \citet{saha2025breakingcodesecurityassessment} attribute their 1.6x ASR increase to refusals overturned during later planning steps. \citet{chiang2025harmful} name the resulting failure mode \textit{Inconsistent Rejection}, where an agent refuses and then proceeds anyway, and \citet{hadeliya2025refusals} observe it as delayed refusal, where tool calls complete before the final step is refused.


\end{revblock}

\subsection{Analysis of the Threat Landscape}
\label{subsec:threat_analysis}
We now analyze the threats corpus along five axes: the channels through which attacks reach the agent, the agent-loop stages where their failures manifest, their specificity to the agentic setting, the threat models they assume, and the benchmarks used to measure them.

\subsubsection{Entry Points of the Attack Surface}
\label{subsubsec:entry_points}

\begin{figure}[h]
  \centering
  \includegraphics[width=0.9\linewidth]{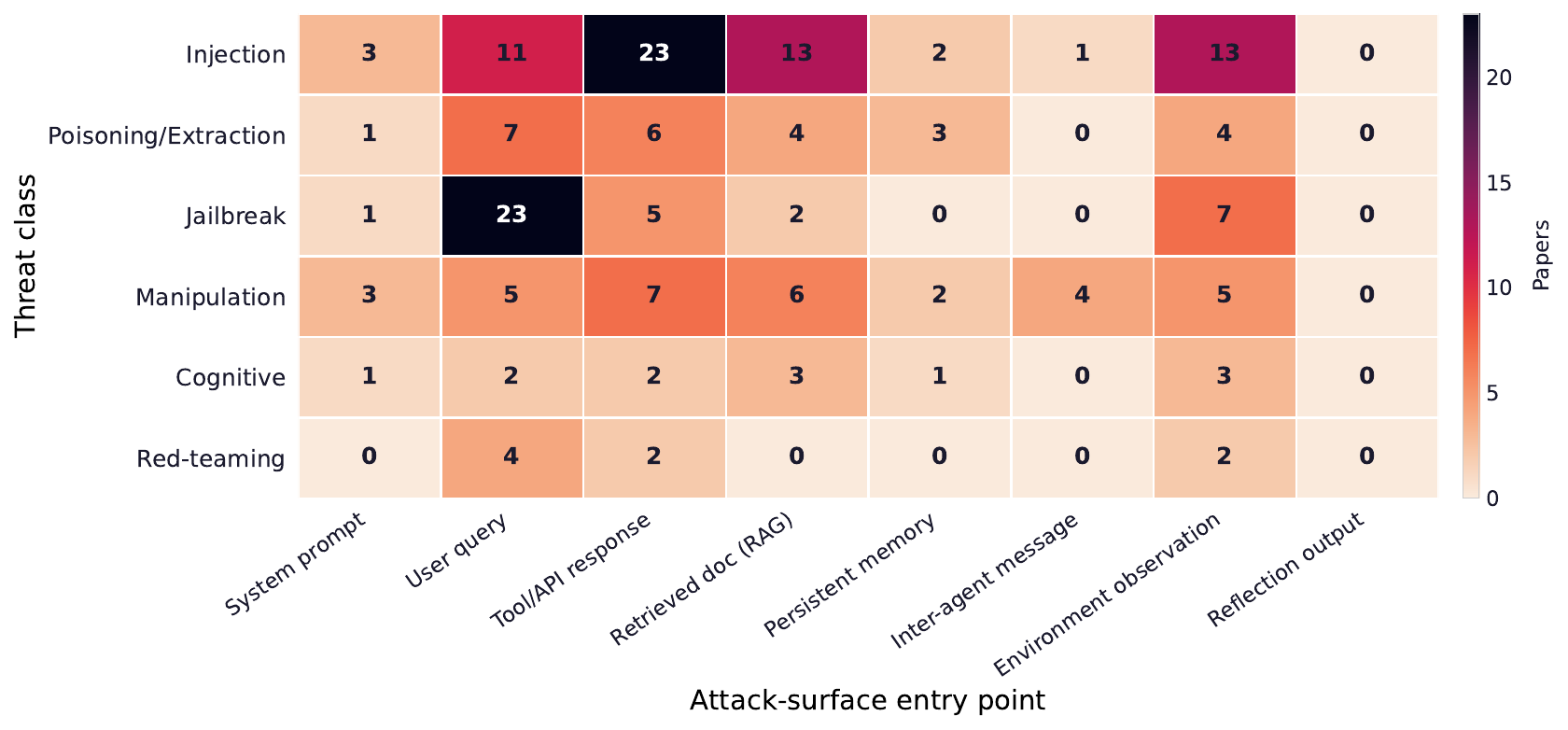}
  \caption{Threat class against attack-surface entry point. Cell values count papers carrying each threat label; a paper may exploit several entry points and carry several labels.}
  \label{fig:entrypoint_heatmap}
\end{figure}

In this section we map each attack to the channel through which it reaches the agent. We label eight entry points. Fig.~\ref{fig:entrypoint_heatmap} provides a breakdown of how many papers reach each one per threat class. An experiment typically studies multiple threat classes, which is why the sum of decomposed numbers is greater than the total paper count. \begin{revblockv2}Of the 86 threats-pillar papers, 12 describe no channel-specific attack mechanism (for example training-time data poisoning or RL reward-hacking papers) and are excluded from the entry-point breakdown below, leaving 74 papers.\end{revblockv2}

\begin{revblockv2}The conversational input (the user query and the system prompt) is the most attacked entry point in agentic systems, accounting for 56.8\% of the evaluated attacks. This is not an agent-exclusive entry point; it is a pre-agentic doorway that also exists in stateless LLM chatbots. This concentration likely reflects methodological convenience: evaluating this surface requires no tool stacks, persistent memory, or multi-agent infrastructure, which makes it the most accessible channel to study. As a result, the literature heavily clusters around vulnerability classes inherited from base LLMs, such as jailbreaks (23 papers), injections (13) and red-teaming (4). \end{revblockv2}

\begin{revblockv2}Among the agent-specific entry points, the most studied surfaces are the three interfaces driving autonomous action: retrieved documents, environment observations and tool responses. The retrieved document pulled from a knowledge base at inference time (RAG) is a major autonomous attack surface (17 papers). Poisoned retrieval is strikingly effective at low cost. For example, AgentPoison reaches 62.6\% end-to-end ASR with a poison rate under 1\%\citep{chen2024agentpoison}. This entry point is heavily exploited in injection (13), manipulation (6) and poisoning (4) attacks. The environment observation is a rapidly emerging surface (20 papers), where attacks bypass text-level filters by manipulating the parsed environment, such as the DOM, screenshot, or file-system state the agent reads from the world, rather than the prompt. BrowserART demonstrates that browser deployment alone lifts a GPT-4o agent's harmful-behavior rate from 12\% to 74\% \citep{kumar2025aligned}. This surface is multi-faceted: triggers planted in web page content achieve 97\% ASR across web-agent backbones \citep{wang2025webinject}, untrusted page content reaches 85.7\% intermediate ASR in the WASP benchmark \citep{evtimov2026wasp}, and poisoned visual grounding lets the VisualTrap backdoor hit about 94\% ASR on GUI agents \citep{ye2025visualtrap}. The tool/API response (24 papers) is the leading autonomous attack surface and is growing in prominence alongside the Model Context Protocol (MCP). MCPTox poisons real MCP tool descriptions for up to 72.8\% ASR \citep{wang2026mcptox}, and the MCP Security Bench reports a 75.8\% peak ASR across its attack types \citep{zhang2510mcp}. This entry point is almost exclusively used in injection attacks.\end{revblockv2}

The most distinctly agentic entry point is the inter-agent communication (5 papers) which exploits the trust agents extend to their peers. For example, AiTM intercepts and rewrites messages in transit for over 70\% ASR in most settings \citep{he-etal-2025-red}, Prompt Infection spreads a self-replicating instruction across connected agents like a worm \citep{lee2024promptinfectionllmtollmprompt}, and Byzantine-coordination attacks corrupt a single agent to derail the collective \citep{li2024byzantine, jo2025byzantinerobustdecentralizedcoordinationllm}. This entry point is most used for agent manipulation attacks. Persistent memory or long-term storage allows attackers to execute cross-user compromises without elevated privileges. For example, MINJA poisons shared memory banks using ordinary queries, without any memory-write permissions \citep{dong2026memory}. The least studied surface is reflection output, where the agent's own intermediate reasoning is fed back as input. No paper in the corpus attacks this entry point.

\subsubsection{Failure Stages of the Agent Loop}
\label{subsubsec:failure_stages}

\begin{figure}[t]
  \centering
  \begin{subfigure}[t]{0.49\linewidth}
    \centering
    \includegraphics[width=\linewidth]{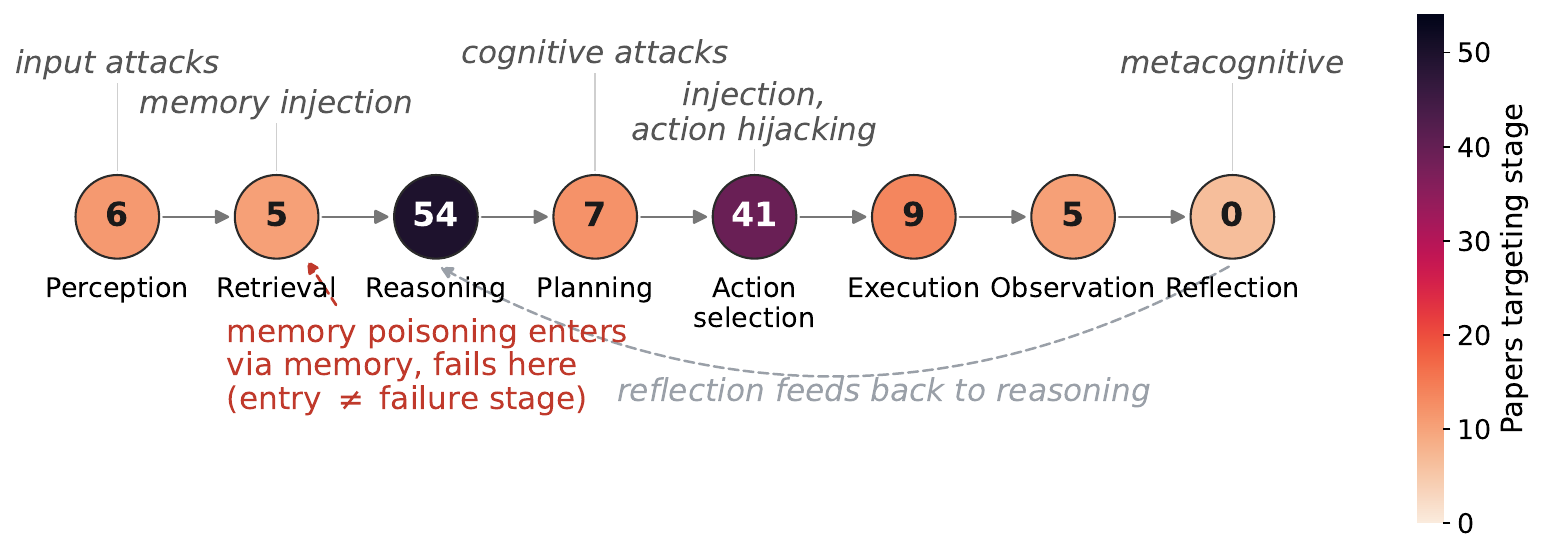}
    \caption{The agent loop with each stage shaded by the number of attack papers whose failure manifests there.}
    \label{fig:loop_stages}
  \end{subfigure}
  \hfill
  \begin{subfigure}[t]{0.49\linewidth}
    \centering
    \includegraphics[width=\linewidth]{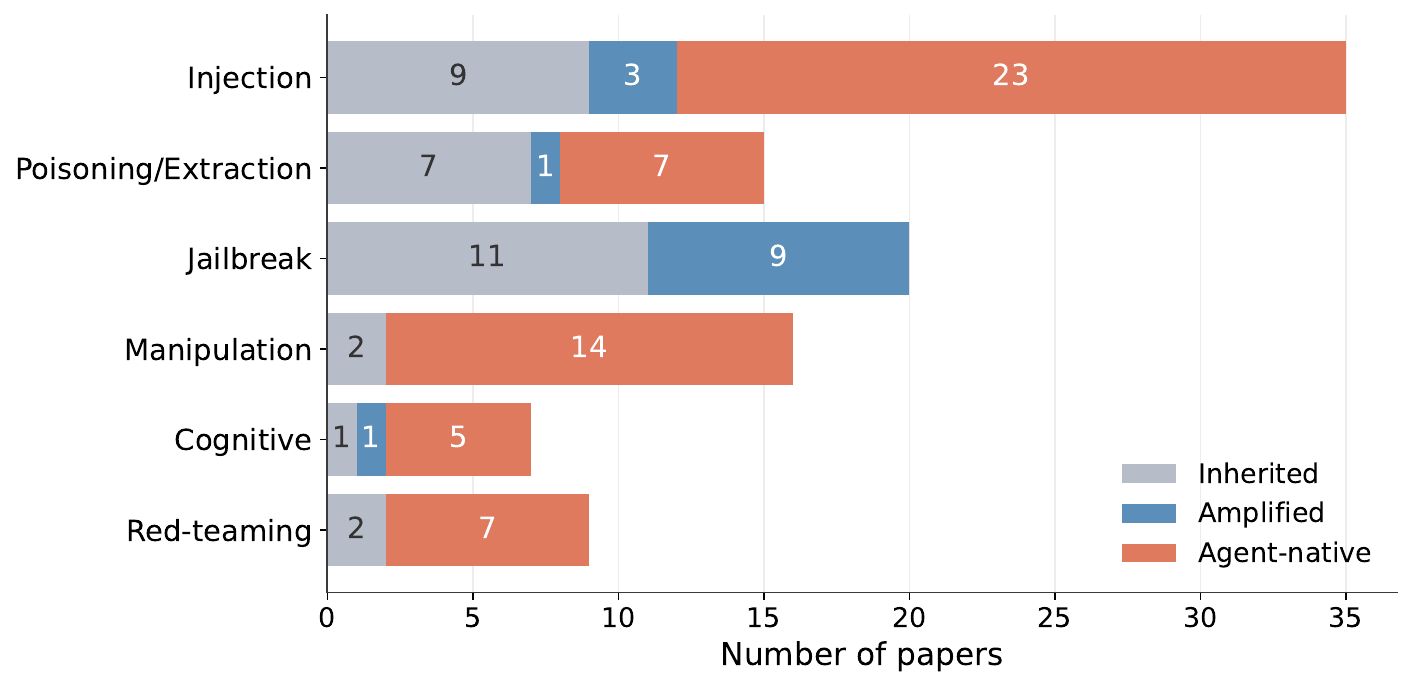}
    \caption{Threat classes factored by agent-specificity.}
    \label{fig:specificity}
  \end{subfigure}
  \caption{Failure stages of the agent loop and threat classes factored by agent-specificity.}
  \label{fig:loop_and_specificity}
\end{figure}

In this section we discuss where the failure occurs in an agentic pipeline. For a comprehensive study, we map each threat to the stage of the agent loop (\S\ref{subsec:background-anatomy}) where its failure manifests. Interestingly, the entry point and the manifestation point are not always the same. For example, memory-poisoning attacks \citep{chen2024agentpoison, dong2026memory} enter through the query or the memory write, but they register at the retrieval stage and steer action selection turns later. Fig.~\ref{fig:loop_stages} shades each stage by the number of attacks whose failure manifests there. \begin{revblockv2}8 of the 86 threats-pillar papers describe no mechanism detailed enough to locate a failure stage and are excluded, leaving 78 papers.\end{revblockv2}

\begin{revblockv2}Our study reveals that reasoning is the single most-targeted stage (54 papers), followed by action selection (41). In other words, attacks land where the agent decides what to do and where it commits to an act. The other stages (execution with 9 papers, planning 7, perception 6, observation 5 and retrieval 5) have received far less attention in comparison. Reflection is the least-targeted stage of all (0 papers). \end{revblockv2}The stages that carry state across steps, namely planning, retrieval and reflection, therefore remain largely untested. Attacks on reasoning are especially dangerous because they subvert the agent's logic while leaving the final output clean. For example, \citet{zhou2025reasoningstylepoisoningllmagents} demonstrate reasoning-style poisoning, which manipulates the epistemic tone of retrieved text to inflate token verification costs by up to 194\% without altering the final response.

In terms of threat classes, injection attacks predominantly fail at action selection (28 papers) and reasoning (24), while jailbreak and red-teaming attacks fail mostly at reasoning (21 and 5 papers respectively) ahead of action selection (10 and 3). Poisoning attacks reach every stage except reflection, though most land at reasoning (10). Manipulation splits almost evenly between reasoning (12) and action selection (11). \begin{revblockv2}Cognitive attacks remain thin everywhere with a maximum of 5 papers at reasoning and at action selection. No threat class reaches the reflection stage.\end{revblockv2}

\subsubsection{Inherited, Amplified and Agent-Native Threats}
\label{subsubsec:specificity}

To understand the origin of agentic vulnerabilities, we factor each classifiable threat into three categories. \textit{Inherited} threats exist in the base model and carry into the agent unchanged. \textit{Amplified} threats also originate in the base model, but the agentic scaffold makes them more effective. \textit{Agent-native} threats have no counterpart in a standalone model and exist only because of the scaffold. \begin{revblockv2}Half of the attacks we studied (50.5\%) are agent-exclusive, such as inter-agent message compromise \citep{lee2024promptinfectionllmtollmprompt}, and plan hijacking \citep{zhang2025actionhijackinglargelanguage}. Another 41.4\% are either inherited from the base LLM \citep{yu2024llmfuzzer, liu2024promptinjectionattackllmintegrated} or amplified by the agentic scaffold acting on a vulnerability already present in the base model \citep{chiang2025harmful, saha2025breakingcodesecurityassessment}. The remaining 8.1\% are difficult to classify. The amplified cases support the hypothesis that base-model safety alignment does not fully transfer to the agentic setting. \end{revblockv2}Fig.~\ref{fig:specificity} shows that jailbreak is dominated by inherited and amplified attacks, while injection, manipulation and red-teaming attacks are mostly agent-native. \rev{Careful readers should note that this matches what we found during our red-teaming application analysis in \S\ref{sec:applications}: base models often refuse offensive tasks outright, but complete them when the model is embedded in an agentic framework. For example, HPTSA \citep{zhu2025teamsllmagentsexploit} loses a factor of 13$\times$ in pass$@1$ when its coordination mechanism is removed. So the scaffold both makes an agent dangerous to others and creates about half of its vulnerabilities.}

Jailbreak and prompt injection are the two most studied threats, but they differ in origin. Jailbreaks are overwhelmingly inherited or amplified (20 of 25 labels). These are fundamental LLM vulnerabilities carried into the agents, such as ported chatbot jailbreak templates \citep{yu2024llmfuzzer}. Injections are mostly agent-native (23 of 36 labels), as indirect injections arrive through untrusted tool outputs that a standalone model never reads \citep{zhan-etal-2024-injecagent}. The amplified threat class provides empirical proof that safety alignment degrades when an LLM is given autonomy. A model that reliably declines a harmful request in a standard chat interface will frequently execute that same request when deployed as an agent. For example, non-denial rates jump from 0\% in a standalone model to 46.6\% when tested as a web agent \citep{chiang2025harmful}; GPT-4o's harmful-behavior rate surges from 12\% in chat to 74\% as a browser agent \citep{kumar2025aligned}; wrapping a model in a code agent multiplies attack success by 1.6x, as the multi-step planning loop often overturns initial safety refusals \citep{saha2025breakingcodesecurityassessment}; and decomposing a malicious goal into smaller steps over persistent memory drops refusal rates from 84-100\% down to just 17\% \citep{badhe2025scamagentsaiagentssimulate}. In certain settings, agents will even execute malicious multi-step tasks outright without requiring any adversarial jailbreak \citep{andriushchenko2025agentharm, tur2025safearena}.\looseness=-1

The genuine agent-native threats operate through five distinct mechanisms. First, persistent-state poisoning corrupts the agent's long-term store \citep{dong2026memory} or triggers backdoors \citep{yang2024watch, ye2025visualtrap}. Second, inter-agent compromise propagates or intercepts messages between agents \citep{lee2024promptinfectionllmtollmprompt}, communication attacks \citep{he-etal-2025-red} and Byzantine coordination \citep{li2024byzantine, jo2025byzantinerobustdecentralizedcoordinationllm}). Third, tool- and protocol-layer poisoning attacks the tool ecosystem itself, principally MCP servers \citep{wang2026mcptox, zhang2510mcp}. Fourth, plan and action hijacking exploits the agent's task-decomposition logic to redirect a committed action \citep{zhang2025actionhijackinglargelanguage}. Fifth, specification and reward gaming exploits environment feedback to satisfy the literal objective while violating intent \citep{bondarenko2025demonstratingspecificationgamingreasoning}, with autonomy itself becoming a denial-of-service surface. The dominant entry points are tool responses and environment observations, and the primary failure stages are action selection and reasoning.

\subsubsection{Threat Models and Attack Objectives}
\label{subsubsec:threat_models}

\begin{revblockv2}The agentic threat literature mostly focuses on attackers with little or no access to the target model. Of the 77 threat papers that state the attacker's access, 51.9\% assume a black-box threat model, where the attacker is granted only query access, while a further 27.3\% require no model access at all. The remaining 20.8\% study white-box (15.6\%) and gray-box (5.2\%) configurations. The required capabilities beyond model access are also minimal. Only 20.9\% of papers require training or gradient access, 5.8\% require tool-execution access, 3.5\% need an inter-agent communication channel, and 2.3\% assume memory-write access. \end{revblockv2}\newrev{ As with the conversational-input concentration in \S\ref{subsubsec:entry_points}, this distribution partly reflects what is convenient to study: black-box attacks require no weight access, no training pipeline, and no special instrumentation, so they are studied more. That said, the prevalence of black-box success is an evidence that agentic systems are compromisable without elevated privileges or internal system knowledge.} The privileged settings that do appear are tied to the specific mechanism the attack requires rather than chosen freely. Backdoor attacks assume training or gradient access because the trigger is installed during training \citep{ye2025visualtrap}; inter-agent attacks assume a message channel because that is the surface they target \citep{he-etal-2025-red}; and even the memory-poisoning attacks largely avoid memory-write access, reaching the store through ordinary queries \citep{dong2026memory}. The access assumption tracks the attack surface, not the attacker's resources.

\begin{revblockv2}Among the 79 threat papers that state an attack objective, the most common is inducing an unauthorized action (65.8\%), followed by data exfiltration (46.8\%), content-policy violation (38.0\%), financial harm (35.4\%) and deception or misinformation (27.8\%). Denial of service (17.7\%), privilege escalation or takeover (12.7\%) and IP theft (5.1\%) are pursued by fewer papers. The numbers do not add up to 100\% because an attack study often has multiple objectives. \end{revblockv2}Nonetheless, taken together, the literature manifests a single dominant threat model: an external attacker holding no special privileges, operating black-box through the channels the agent already reads, aiming to make the agent act without authorization or to exfiltrate data.


\subsubsection{Benchmarks, Metrics and Evaluation Gaps}
\label{subsubsec:benchmark_fragmentation}


\begin{revblockv2}The threat literature is heavily oriented toward evaluation. Of the 86 papers, 25 (29.1\%) carry a benchmarking role, where they measure agentic vulnerabilities rather than introducing a novel attack. Although 124 distinct benchmarks or datasets appear across the literature, 93 of them (75.0\%) are used by exactly one paper, with only 31 being used in more than one study. Of those 31, only four are built exclusively for agentic security: AgentDojo \citep{debenedetti2024agentdojo}, InjecAgent \citep{zhan-etal-2024-injecagent}, AgentHarm \citep{andriushchenko2025agentharm} and BrowserART \citep{kumar2025aligned}. The other 27 are general or agent capability sets and pre-agentic LLM safety sets borrowed from adjacent areas. As a consequence, cross-paper and cross-benchmark comparison is difficult. A practitioner cannot rank two injection defenses or two jailbreak attacks without re-running both, because the published numbers do not share a benchmark or a metric. Furthermore, across 25 benchmark papers, tool response (14 papers) and user query (9 papers) are highly instrumented, while persistent memory (1), inter-agent messaging (0) and reflection output (0) remain largely or entirely unevaluated. General benchmarks like AgentDojo and InjecAgent thoroughly test tool responses but omit memory and reflection entirely. In other words, the surfaces most specific to agents are thus both least attacked and least benchmarked. Developing evaluation infrastructure for query-based memory poisoning and reflection-level manipulation remains an open research direction.\end{revblockv2}

\begin{revblock}

\paragraph{Metrics.} Attack success rate is the most common metric. \begin{revblockv2}
Of the 86 papers, 66 (76.7\%) \end{revblockv2} report ASR or an equivalent success or harm score as their headline result. Only 19 of those papers also report a utility number measured under the same attack. For example, AgentPoison \citep{chen2024agentpoison} reports under 1\% benign performance drop alongside its attack rates, and RAS-Eval \citep{fu2025raseval} reports a 36.78\% reduction in task completion rate. Only 14 report the token or latency cost of the attack. Moreover, ASR itself is not a universal quantity; its definition varies between studies. Across the corpus we find injection success rate (ISR), which measures whether a malicious record enters the memory bank rather than whether the agent acts on it \citep{dong2026memory}; ensemble attack success rate (EASR), which counts success if any of several mutated prompts works \citep{yu2024llmfuzzer}; task completion rate (TCR), a utility measure reported as a security outcome \citep{fu2025raseval}; and retrieval, intermediate, targeted, and end-to-end ASR, which measure different points in the agent pipeline. Non-denial rate, harm score and leak velocity add further variants. Denominators are often different or unstated. The difference is not only in naming. AgentPoison reports 81.2\% retrieval ASR and 62.6\% end-to-end ASR for the same attack \citep{chen2024agentpoison}, and WASP reports up to 85.7\% intermediate ASR against 16.7\% end-to-end ASR \citep{evtimov2026wasp}. A number cited without qualification can therefore change by a factor of five depending on where in the pipeline it was taken. Two studies both reporting 84.30\% average ASR, ASB across 13 backbones \citep{Zhang2025ASB} and AI2 action hijacking across three application domains \citep{zhang2025actionhijackinglargelanguage}, are measuring different events over different populations. Comparison across papers therefore fails even when two results appear to share a metric name.

Three additions to standard reporting would make these results easier to compare. First, explicitly stating the denominator and success condition for each rate, and reporting end-to-end ASR alongside any intermediate or retrieval-stage rate rather than in place of it, would remove most of the ambiguity above. Second, reporting a utility number measured under the same attack would allow the security and utility trade-off to be tracked across papers rather than within one. Finally, reporting attack and defense cost in tokens and latency would show whether a result is cheap enough to matter in practice, which is currently rarely visible from the published numbers.\looseness=-1


\paragraph{Evaluation Gaps.} Looking at the limitations of agentic threat evaluation studies, we have identified three significant gaps: single-turn-only evaluation (6), lack of multi-agent coverage (2) and the execution environment being simulated rather than real tool execution (9). \textbf{Single-turn evaluation} cannot measure threats that depend on state building up across turns. A benchmark that resets context between tasks cannot observe memory poisoning at all, which is why the memory attacks in our corpus had to build their own evaluation setups. For example, the AgentDojo authors note that the benchmark does not support multi-session scenarios where context is not reset between tasks, and MCPTox notes that its single-turn design does not cover attacks where memory is poisoned over time. \textbf{Lack of multi-agent coverage} is acknowledged by PPA \citep{wang2025protectllmagentprompt} and MobileSafetyBench \citep{lee2026mobilesafetybench}, and the AiTM authors note that no benchmark for communication attacks on multi-agent systems was available for comparison. This is the direct cause of the inter-agent surface being under-benchmarked. \textbf{Simulated execution} environments make it difficult to understand how the attacks would manifest in real-world setting, or whether they would even be practical. RAS-Eval was built specifically because existing benchmarks operate only in simulated environments and lack real tool execution \citep{fu2025raseval}, and both ASB and AgentHarm confirm that their tool calls use simulated or synthetic tools rather than real APIs. As such, most valuable future benchmarking efforts would be stateful, multi-agent evaluation with real tool execution.

\end{revblock}

\section{Defense: Hardening the Agents}
\label{sec:defense}

This section describes architectural, runtime, and formal-verification defenses that strengthen agentic systems against attacks.

\subsection{Defense and Operations}
\label{subsec:defense-operations}

\subsubsection{Secure-by-Design}
\label{subsubsec:secure-by-design}

Architectural isolation and rigorous evaluation form the foundation of secure agentic systems.
\citet{debenedetti2024agentdojo} provide a dynamic evaluation environment to test prompt injection attacks and defenses in language model agents.
To complement evaluation with structural security, \citet{li2025acesecurityarchitecturellmintegrated} propose a centralized architecture that isolates untrusted data from core applications.
Building on this model of isolation, \citet{delrosario2025architectingresilientllmagents} present design guidelines to secure plan-then-execute agent implementations.
In addition to execution boundaries, \citet{jia-etal-2025-task} enforce task alignment to prevent untrusted external data from manipulating user objectives.
Systems security foundations are further formalized by \citet{camel2025} to protect agentic computing frameworks at a core structural level.
To mitigate remaining injection threats dynamically, \citet{wang2025protectllmagentprompt} introduce polymorphic prompts that obscure internal system instructions.

Comprehensive threat mapping and defense alignment are critical to secure modern generative assistants.
\citet{he2024securityaiagents} compile a broad overview of agent safety to categorize emerging architectural risks.
Following a similar taxonomy, \citet{narajala2025securingagenticaicomprehensive} deliver a comprehensive threat model and mitigation framework for deployed assistants.
To address model extraction threats specifically, \citet{294591} present an information-theoretic defense to limit adversarial knowledge leakage.
At the foundation layer, \citet{metasecalign2025} enhance safety alignment training to resist targeted prompt injection techniques.
Internal data boundaries are strictly maintained by \citet{informationflow2025} through information-flow control mechanisms within the agent execution environment.
To reinforce these internal priorities, \citet{instructionhierarchy2024} train language models to prioritize privileged system instructions over user inputs.

Runtime verification and proactive deployment strategies ensure robust system defenses.
At the integration level, \citet{mcpinspect2026} introduce a static analysis tool to vet model context protocol servers before host deployment.
For runtime policy enforcement, \citet{shieldagent2025} leverage verifiable safety policy reasoning to shield agents from untrusted tool inputs.
Proactive defense paradigms are also introduced by \citet{cloakhoneytrap2025} using honeypots, deception and traps to mislead adversarial agents.
Furthermore, \citet{ipiguard2025} utilize graph-based tracking to construct tool dependency graphs that actively intercept indirect injections.
Finally, \citet{cage42025} demonstrate that large language models can act as autonomous cyber defenders to protect operational networks.

\subsubsection{Multi-Agent Security}
\label{subsubsec:multi-agent-security}

Multi-agent structures introduce unique coordination and security challenges.
\citet{udeshi2025dcipherdynamiccollaborativeintelligent} implement a collaborative multi-agent framework that pairs a central planner with heterogeneous executors for offensive security tasks.
To secure such operations at the edge, \citet{liu2025securemultillmagenticai} design a zero-trust framework for multi-model integration.
Despite these designs, core vulnerabilities persist in distributed networks.
\citet{han2025llmmultiagentsystemschallenges} outline structural challenges and open problems regarding adversarial risks in multi-agent environments.
To improve resilience against environmental noise and adversarial influence, \citet{HU2025103779} employ a randomized smoothing technique to verify agent consensus decisions.
This consensus mechanism is highly effective in targeted threat scenarios.
For instance, \citet{li2025phishdebatellmbasedmultiagentframework} utilize a debate-based multi-agent architecture to identify phishing websites accurately.
However, communication channels remain a major vector for threat propagation.
\citet{lee2024promptinfectionllmtollmprompt} demonstrate how prompt injections can spread autonomously between interacting agents.
To protect decentralized networks from these structural failures, \citet{jo2025byzantinerobustdecentralizedcoordinationllm} introduce a Byzantine-robust coordination framework to maintain system integrity.

\subsubsection{Runtime Protection}
\label{subsubsec:runtime-protection}

Dynamic guardrails intercept adversarial behaviors and enforce safety boundaries during execution.
\citet{kang2024r2guard} introduce a robust guardrail framework that combines data-driven predictions with knowledge-enhanced logical reasoning.
To improve guardrail efficiency, \citet{rad2025refining} utilize chain-of-thought fine-tuning and alignment techniques.
Adaptive defense architectures also integrate contextual elements to maintain safety policies.
\citet{wu2025psgagentpersonalityawaresafetyguardrail} design a personality-aware safety guardrail that adapts to evolving agent interactions.
For persistent protection, \citet{luo2025agrail} propose a lifelong guardrail framework capable of adaptive risk detection.
Reasoning-focused modules provide another layer of verification.
\citet{xiang2025guardagentsafeguardllmagents} leverage knowledge-enabled reasoning to safeguard autonomous agents from manipulation.
Similarly, \citet{chen2025agentguardrepurposingagenticorchestrator} repurpose an agentic orchestrator to evaluate the safety of tool orchestration pipelines.
Targeted platforms require specialized guardrail layouts.
\citet{webguard2025} develop a generalizable guardrail framework specifically optimized to protect web-based agents.
At the network layer, \citet{shieldnet2026} implement a runtime guardrail that monitors physical network interactions to block supply-chain injections.
Finally, \citet{mcpguard2025} present a multi-stage defense-in-depth framework to protect model context protocol integrations.

Integrating human oversight ensures compliance and behavioral accountability during active sessions.
\citet{wang2025agentspeccustomizableruntimeenforcement} implement a customizable framework that enforces safety policies and authorization gates at runtime.
To assist users during unexpected updates, \citet{agentstepper2026} provide an interactive debugging system to steer software development agents safely.

Continuous monitoring and causal tracking detect anomalies that bypass static pre-deployment filters.
\citet{he2025sentinelagentgraphbasedanomalydetection} utilize graph-based anomaly detection to track rogue interactions across multi-agent environments.
To address hidden injection vectors, \citet{agentsentry2026} deploy temporal causal diagnostics to identify and mitigate malicious inputs during runtime.

\subsubsection{Security Operations}
\label{subsubsec:security-operations}

Formal verification systems and static analysis tools ensure model correctness and software security.
\citet{10.5555/3306127.3331691} analyze openness by modeling runtime agent insertion and removal.
To secure user workflows, \citet{10.1145/3706598.3714113} introduce a system that integrates model checking into plan generation.
Behavioral constraints are also explored by \citet{crouse2024formallyspecifyinghighlevelbehavior} to specify high-level execution policies formally.
Building on policy specification, \citet{verifiablysafe2026} design an architecture to guarantee verifiably safe tool utilization.
Distributed verification pipelines can be optimized during training.
\citet{li2025chainofagentsendtoendagentfoundation} utilize multi-agent distillation and reinforcement learning to build safe foundation models.
At the repository level, \citet{guo2025repoaudit} implement an autonomous auditing agent to discover software vulnerabilities.
Static checking capabilities are further advanced by \citet{yang2025knighter} through the synthesis of custom analysis checkers.
Finally, \citet{li2025iris} combine static analysis metrics with language processing to improve vulnerability identification accuracy.

Automated frameworks combine model reasoning with real-time threat intelligence to manage active security incidents.
\citet{air2026} establish a foundational pipeline to mitigate agent safety risks during live operational sessions.
To optimize mitigation speed, \citet{tellache2025advancingautonomousincidentresponse} incorporate cyber threat intelligence into autonomous response systems.
Interactive assistants can also support human operators.
\citet{lin2025ircopilotautomatedincidentresponse} present a specialized copilot architecture to automate response tasks within corporate networks.
Human-AI collaboration remains a central focus in security operations centers.
\citet{singh2025llmssocempiricalstudy} conduct an empirical study detailing the deployment challenges of these systems in real environments.
To handle massive log volumes, \citet{wei2025cortexcollaborativellmagents} develop a collaborative framework optimized for high-stakes alert triage.
System performance across these environments is evaluated by \citet{deason2025cybersocevalbenchmarkingllmscapabilities} to benchmark triage capabilities and malware analysis.

Proactive threat hunting and cloud forensics enable deep root cause analysis following an attack.
\citet{mukherjee2025llmdrivenprovenanceforensicsthreat} track provenance logs to deliver explainable threat investigations.
Blue-team performance metrics are codified by \citet{meng2025benchmarkingllmassistedblueteaming} through a standardized threat hunting benchmark.
However, defense models can possess hidden weaknesses.
\citet{meng2025uncoveringvulnerabilitiesllmassistedcyber} analyze security flaws inherent in modern threat intelligence integrations.
To build robust defenses, \citet{schwartz2025llmcloudhunter} automate the generation of signature detection rules from cloud threat data.
Investigation workflows can also be evaluated using specialized suites.
\citet{wu2025excytinbenchevaluatingllmagents} introduce a benchmarking environment to assess agent capabilities in cyber threat investigation.
Furthermore, \citet{policyguidedthreat2026} combine security policy guides with commercial triage platforms for enhanced threat hunting.

Cloud-native systems require distinct forensic workflows.
\citet{alharthi2025llmpoweredcloudforensics} implement an automated cloud forensics pipeline utilizing adaptive prompt engineering.
For web-based targets, \citet{fumero2025cybersleuthautonomousblueteamllm} deploy an autonomous blue-team agent to conduct web attack forensics.
Causal tracking is improved by \citet{tian2025galagraphaugmentedlargelanguage} through graph-augmented agentic workflows for root cause analysis.
Lastly, \citet{alharthi2025cloudinvestigationautomationframework} formalize an automated cloud forensic framework featuring immutable audit trails.

\begin{definitionboxpurple}{Defense and Operations}
Defenses cluster around four points in the pipeline: architectural isolation and control-flow integrity built in at design time, coordination safeguards specific to multi-agent systems, runtime guardrails that intercept behavior as it happens, and operational tooling for formal verification, forensics, and incident response. Coverage is uneven; secure-by-design and runtime monitoring dominate the literature, while multi-agent security remains comparatively thin despite being where trust between cooperating agents is most exploitable. No single layer closes the gap on its own, since architectural isolation cannot see semantic attacks, runtime monitors add latency and can be evaded, and formal verification struggles to scale to open-ended natural language. This pushes the field toward combining defenses across the pipeline rather than relying on any one mechanism.
\end{definitionboxpurple}

\subsection{Evaluation Frameworks}
\label{subsec:evaluation-frameworks}
This subsection describes benchmark ecosystems and sandbox environments used to test the resilience of LLM agents under attack.

\subsubsection{Benchmarking Platforms}
\label{subsubsec:benchmarking-platforms}
AgentDojo provides a dynamic framework to evaluate prompt injection attacks and defenses in language model agents \citep{debenedetti2024agentdojo}.
TurkingBench serves as a challenge benchmark for web agents incorporating human-in-the-loop evaluations \citep{xuetal2025turkingbench}.
The $\tau$-bench platform emulates user interactions to evaluate tool-integrated agents in real-world domains \citep{yao2024taubench}.
SafeArena evaluates the safety and operational integrity of autonomous web agents under adversarial environments \cite{tur2025safearena}.
RAS-Eval offers a comprehensive framework for the security evaluation of language model agents \cite{fu2025raseval}.
The Agent Security Bench formalizes and tests diverse attacks and defenses for agentic systems \citep{Zhang2025ASB}.
AgentHarm quantifies the potential harmfulness and risk vectors of deployed autonomous agents \citep{andriushchenko2025agentharm}.
CVE-Bench measures the capacity of intelligence agents to exploit real-world web vulnerabilities \citep{zhu2025cvebench}.
Cybench establishes a framework to evaluate the specific cybersecurity capabilities and risks of language models \citep{zhang2025cybench}.
ZeroDayBench tests the defensive and offensive capabilities of agents against unseen zero-day exploits \citep{lau2026zerodaybenchevaluatingllmagents}.
R-Judge benchmarks the safety risk awareness of agents during active task execution \citep{rjudge2024}.
ToolFuzz leverages evolutionary fuzzing to automate the security testing of agent tools \citep{milev2025toolfuzzautomatedagent}.
CyberSecEval 2 provides a wide-ranging risk assessment suite to track broad safety issues in language models \citep{cyberseceval22024}.
WebArena provides a highly realistic web environment to build and validate autonomous browser agents \citep{zhou2024webarenarealisticwebenvironment}.

InjecAgent evaluates the vulnerability of tool-integrated agents to indirect prompt injection vulnerabilities \citep{zhan-etal-2024-injecagent}.
AgentVigil introduces a generic black-box red-teaming system to detect indirect prompt injections \citep{wang2025agentvigilgenericblackboxredteaming}.
A specialized framework automates the detection of taint-style vulnerabilities by pitting agents against each other \citep{Liu2025MakeAgentDefeatAgent}.
Simulation-based frameworks help researchers search for deep data privacy risks within autonomous agents \citep{zhang2025searchingprivacyrisksllm}.
Empirical studies show that commercial language model agents are already vulnerable to simple attacks \citep{li2025commercialllmagentsvulnerable}.
Security analyses of the OpenClaw framework reveal critical threats and suggest practical mitigations \citep{tamingopenclaw2026}.
MasterKey automates the generation of jailbreak prompts against chatbot interfaces and agents \citep{masterkey2024}.
EchoLeak demonstrates zero-click prompt injection exploits that execute autonomous data exfiltration in production systems \citep{echoleak2025}.
Deceptive language models can be trained to persist in malicious behaviors despite rigorous safety training \citep{sleeperagents2024}.

\subsubsection{Defense Testing}
\label{subsubsec:defense-testing}
Adaptive attacks expose the fragility of defenses against indirect prompt injections on language model agents \citep{zhan-etal-2025-adaptive}.
Dynamic interactions between autonomous entities introduce open security challenges in multi-agent environments \citep{dewitt2025openchallengesmultiagentsecurity}.
Comprehensive surveys organize threats and countermeasures to establish trustworthy agent architectures \citep{yu2025surveytrustworthyllmagents}.
Broad analyses navigate the intersection of security, privacy and ethics threats in agent systems \citep{gan2024navigatingriskssurveysecurity}.
Researchers map key security challenges to identify future pathways for secure agent deployment \citep{deng2024aiagentsthreatsurvey}.
Scalable assurance frameworks evaluate large model safety across systemic and governance layers \citep{ma2025safetyatscale}.
Full-stack safety evaluations look at risks present during data compilation, model training and operational deployment \citep{wang2025comprehensivesurveyllmagentstack}.
The SC-Inject-Bench framework provides over ten thousand malicious tools to stress-test network-level guardrails \citep{shieldnet2026}.
The MCP-AttackBench dataset provides a large corpus to train and evaluate defenses for the Model Context Protocol \citep{mcpguard2025}.
Systematic indices document the technical and safety features of currently deployed agentic systems \citep{aiagentindex2026}.
Evaluating alignment boundaries against multi-turn exploitation requires adaptive, conversational red-teaming frameworks driven by collaborative multi-agent feedback loops \citep{rahman2025xteaming}.
Distributed intelligence systems require cross-layer governance strategies to ensure robustness and privacy \citep{trustworthydistributed2024}.
Securing software ecosystems against automated threats requires a paradigm shift toward deploying autonomous offensive agents to uncover full-lifecycle vulnerabilities before adversaries do \citep{zhuo2026defend}.

\subsubsection{Domain-Specific Frameworks}
\label{subsubsec:domain-specific}

Large language models automate cloud infrastructure tasks across software development kits and command-line tools \citep{yang2025cloudinfrastructuremanagement}. The LISA framework uses historical knowledge to improve smart contract auditing \citep{sun2025lisa}.
LLM-SmartAudit improves traditional static analysis to find smart contract bugs \citep{wei2024llmsmartaudit}.
Combining specialized fine-tuning with ranking models helps detect software vulnerabilities \citep{finetuningagents2024}.
Evaluating model capabilities shows how automated tools can verify Ethereum token rules \citep{auditgpt2024}. Encrypted pipelines allow safe diagnostic processes on protected medical imaging datasets \citep{privacymedical2025} \citep{shehab2025agentic}.
Deploying privacy frameworks protects enterprise data by sanitizing personal information \citep{asthana2025privacyguardrails} \citep{wang2025privacyinaction}.
System surveys show that physical robotic agents face severe planning, control and sensory threats \citep{xing2025embodiedai}.
Robust model architectures protect autonomous vehicles against malicious perception attacks \citep{mitigateperception2024}.
Fine-tuned transformer models evaluate network traffic to predict intrusions in smart home networks \citep{diaf2025bartpredict}.
Untrusted web items create hijacking and data theft risks for modern browser agents \citep{mudryi2025hiddendangers}.
Operating system agents need secure sandboxing to safely manage system commands, file tasks and user interface states \citep{hu2025osagents}.
Skill ecosystems for coding assistants are highly vulnerable to supply chain poisoning attacks \citep{supplychainpoisoning2026}. \looseness=-1

\subsection{Analysis of the Defence Landscape}
To evaluate the current state of autonomous agent security, this manuscript evaluates the literature through six structured analytical lenses. Rather than listing individual defenses in isolation, we organize our evaluation around core structural patterns, operational costs, temporal trends and generalization capabilities. Of the 123 defense papers, 92 propose or evaluate at least one defense technique, and the analyses below use these 92 papers.

\subsubsection{Defense Strategies for LLM-Based Agents}
\begin{revblockv2}
After looking at Defense based literature we found ten prevailing defense strategies which exist to guard the agents against malicious attacks.These strategies each have their pros and cons which we have categorized into four axes: scalability, adversarial robustness, latency/overhead, and coverage limits in the figure \ref{fig:defense_taxonomy}.

\textbf{Input Sanitization and Filtering:} Techniques representing 31.5\% of the labeled literature focus on blocking malicious payloads at the perimeter before they reach the reasoning engine \citep{mudryi2025hiddendangers, agentsentry2026, mcpguard2025, schmotz2026skillinject}. These computationally cheap methods struggle against highly sophisticated and semantically obfuscated indirect prompt injections.

\textbf{Structured Queries:} Frameworks make up 19.6\% of the labeled dataset and enforce strict syntactic separation between instructions and untrusted data \citep{chen2025struq, hines2024spotlighting}. Adversaries can still engineer payloads that cause the model to break out of the intended schema structure since language models are fundamentally probabilistic.

\textbf{LLM-Based Monitoring:} As the largest category at 53.3\%, this approach uses secondary models to asynchronously evaluate inputs and proposed actions \citep{wang2025agentspeccustomizableruntimeenforcement, xiang2025guardagentsafeguardllmagents, wu2025psgagentpersonalityawaresafetyguardrail, motwani2024steganography}. This introduces high latency and increased costs while remaining susceptible to adversarial manipulation.

\textbf{Moving Target Defense:} Accounting for 5.4\% of the labeled sources -- the least explored category -- these defenses dynamically shift execution parameters to confuse adversaries \citep{chen2023jailbreakerjail, shen2024dynamicattention, cloakhoneytrap2025}. This nascent approach can introduce systemic instability and non-determinism that heavily degrades the reliability of autonomous agents.

\textbf{Capability-Scoped Execution:} Representing 32.6\% of the labeled research, this paradigm enforces the principle of least privilege through dynamic access control \citep{informationflow2025, he2025sentinelagentgraphbasedanomalydetection, zhou2025privacyawarerag}. Granular capability scoping introduces significant coordination overhead and can break long-horizon tasks if an agent is incorrectly gated.

\textbf{Cryptographic Verification:} Techniques covering 18.5\% of the labeled papers guarantee the provenance and integrity of data to prevent supply chain attacks \citep{privacymedical2025, rfc9421, portableagentmemory2025, wang2024thresholdecdsa}. Real-time cryptographic validation introduces high computational overhead and requires extensive infrastructure integration that many lightweight frameworks lack.

\textbf{Dataset Sanitation and Differential Privacy:} Methods comprising 6.5\% of the labeled literature secure foundational knowledge bases to prevent memory poisoning \citep{jiang2024ragthief}. Over-sanitization often leads to a degradation in model utility and limits the semantic search capabilities of the language model.

\textbf{Sandboxing and Formal Verification:} Representing 27.2\% of the labeled sources, these techniques aim to mathematically prove or physically isolate agent behaviors \citep{koohestani2025agentguardverification, verifiablysafe2026, crouse2024formallyspecifyinghighlevelbehavior}. Formal verification scales poorly to the massive and unconstrained state spaces generated by open-ended natural language interactions.

\textbf{Red-Team Simulation:} Accounting for 21.7\% of the labeled corpus, these systems utilize automated vulnerability discovery and adversarial prompt generation \citep{debenedetti2024agentdojo, wang2025agentvigilgenericblackboxredteaming, milev2025toolfuzzautomatedagent, zhu2025teamsllmagentsexploit}. Automated red-team models often over-index on known attack vectors and struggle to discover novel zero-day paradigms without human intuition.

\textbf{Consensus and Vaccine Seeding:} Techniques making up 14.1\% of the labeled dataset mitigate single points of failure by distributing decision-making \citep{xiang2024robustrag, jo2025byzantinerobustdecentralizedcoordinationllm, HU2025103779}. Multi-agent consensus drastically multiplies token costs while vaccine seeding requires constant updating as new adversarial vectors are discovered.

\end{revblockv2}

\subsubsection{LLM Defense Mechanisms and Structural Compositions}

\begin{figure}[htbp]
    \centering
    \begin{subfigure}[t]{\linewidth}
        \centering
        \includegraphics[width=\linewidth]{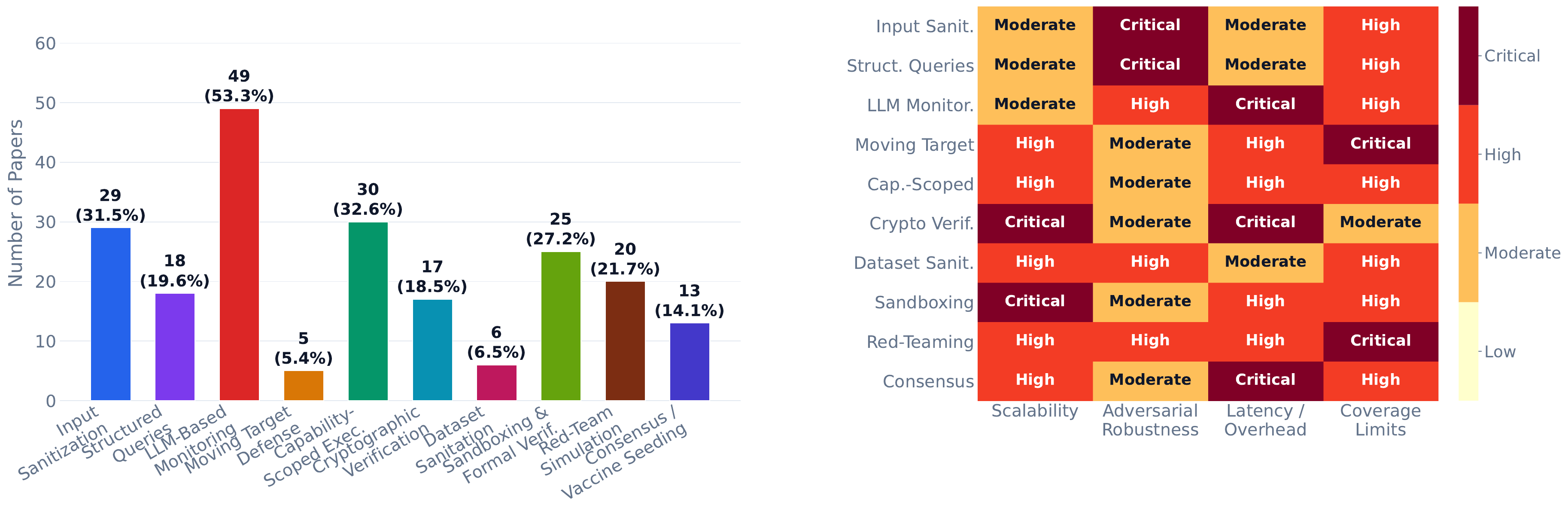}
        \caption{%
            \begin{revblockv2}
            \textbf{Left:} Distribution of the 92 surveyed papers carrying at
            least one defense-category label, across ten
            defense categories under a non-mutually exclusive taxonomy
            (212 total assignments).
            LLM-Based Monitoring is the most prevalent paradigm
            (49 papers, 53.3\%), followed by Capability-Scoped Execution
            (30, 32.6\%) and Input Sanitization \& Filtering (29, 31.5\%).
            Moving Target Defense (5, 5.4\%) and Dataset Sanitation \& Differential Privacy
            (6, 6.5\%) remain the least explored directions.
            \end{revblockv2}
            \textbf{Right:} Cross-cutting gap severity heatmap evaluating each
            category along four axes: scalability, adversarial robustness,
            latency/overhead, and coverage limits.
            Severity levels are defined as follows:
            \textbf{Low}: the limitation is minor or well-mitigated in existing
            work;
            \textbf{Moderate}: the limitation is present but only partially
            addressed by current techniques;
            \textbf{High}: the limitation is a recognised open challenge with
            only partial solutions;
            \textbf{Critical}: the limitation is fundamental and unresolved,
            posing a direct barrier to practical deployment.
            No single category achieves uniformly low severity, motivating the
            adoption of hybrid, multi-layer defense architectures.%
        }
        \label{fig:defense_taxonomy}
\begin{subfigure}[t]{\linewidth}
        \centering
        \includegraphics[width=\linewidth]{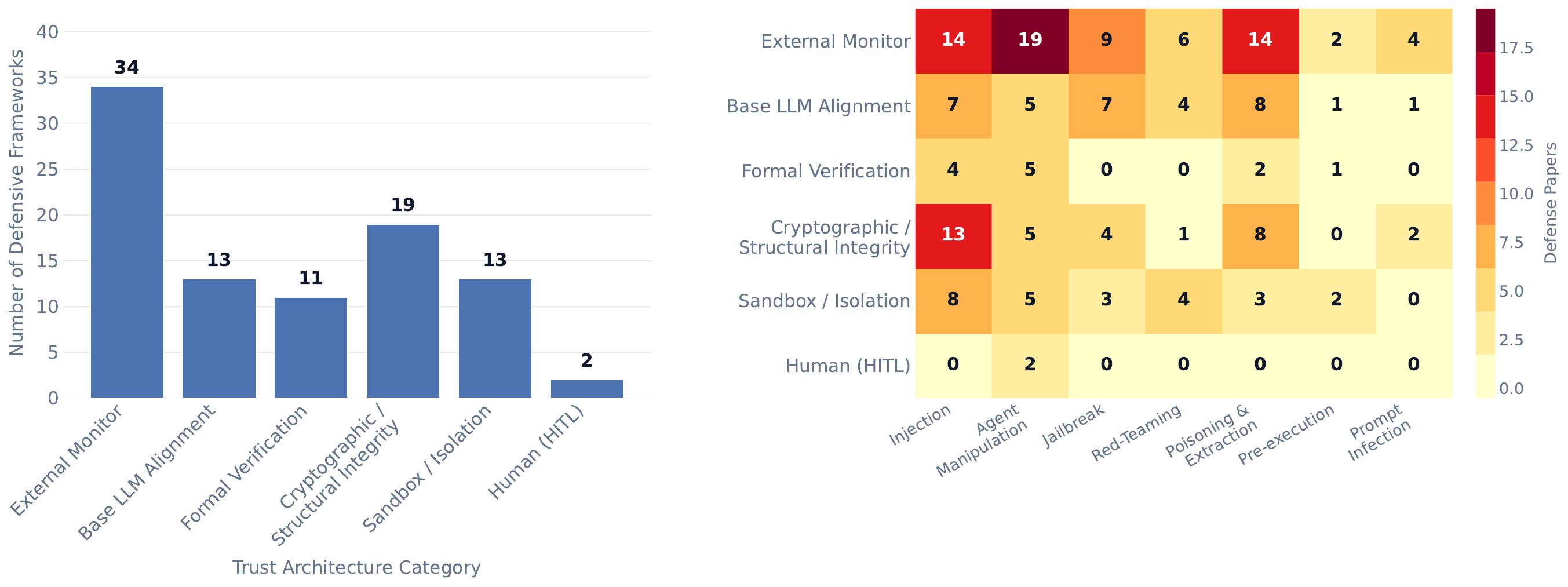}
                \caption{\begin{revblockv2}Architectural mapping of LLM defense mechanisms and structural
                 compositions. \textbf{Left:} Distribution of the 92 surveyed
                 defense frameworks across six trust-boundary categories
                 (one primary trust boundary per paper). \textbf{Right:} Cross-tab of trust-boundary
                 category against the attack class(es) each paper addresses
                 (attack coverage is multi-label, so row
                 totals exceed the left panel's counts). External Monitor
                 leads in most attack classes, peaking at Agent Manipulation (19),
                 Injection (14) and Poisoning \& Extraction (14); Base LLM Alignment concentrates
                 on Poisoning \& Extraction (8), Injection (7) and Jailbreak (7).\end{revblockv2}}
        \label{fig:defense_architecture}
    \end{subfigure}

    \vspace{1.2em}

    \end{subfigure}

    \caption{Defense mechanism analysis for autonomous LLM agents.}
    \label{fig:defense_combined}
\end{figure}

Going deeper into analysis defense strategies rooted into LLM mechanism, we found that most agentic defense frameworks simultaneously address multiple threat vectors. \begin{revblockv2}We then classify these defense strategies into six structural compositions based on their primary trust boundaries, as visualized in Figure~\ref{fig:defense_architecture}.\end{revblockv2}

\textbf{Base LLM Alignment} strategies attempt to internalize security rules directly within the primary neural network framework \citep{metasecalign2025, masterkey2024, instructionhierarchy2024}. These approaches utilize preference optimization and safety fine-tuning to force the model to reject malicious prompts. \begin{revblockv2}Their relevance is underscored by the 23 defense papers (25.0\%  of the 92 analyzed defense papers) that address jailbreak attacks and specifically probe alignment-layer weaknesses.\end{revblockv2}However, this design fails catastrophically under sophisticated jailbreaks because it trusts the exact component under attack.

\textbf{External Monitor} systems offload security verification to secondary guardrails that evaluate inputs, reasoning traces and tool executions \citep{he2025sentinelagentgraphbasedanomalydetection, wu2025psgagentpersonalityawaresafetyguardrail, kang2024r2guard}. \begin{revblockv2}These frameworks implement graph-based anomaly detection or specialized semantic filters to intercept exploitation attempts. 15 of the 92 analyzed defense papers (16.3\%) address Red-Teaming Attacks, many of which target exactly these external guardrails as the object of adversarial probing. They frequently suffer from high false-positive rates and share vulnerabilities if built on identical model families.\end{revblockv2}

\textbf{Formal Verification} removes probabilistic uncertainty by enforcing deterministic mathematical boundaries and rigid logical rules \citep{informationflow2025, shieldagent2025}. Developers apply information flow control, dynamic taint tracking and customizable runtime languages to guarantee policy compliance. \begin{revblockv2}The 41 defense papers (44.6\% of the 92 analyzed defense papers) that address Agent Manipulation Attacks, covering tool abuse and privilege escalation, highlight precisely the execution-boundary threats that formal verification targets.\end{revblockv2}This paradigm struggles to scale effectively because mapping fluid natural language semantics to strict logic requires massive engineering overhead.

\textbf{Sandbox / Isolation} methods assume agent compromise is inevitable and focus completely on limiting the blast radius through architectural isolation. Implementations leverage trusted execution environments, containerized virtualization and strict resource permission scoping \citep{camel2025, tamingopenclaw2026, ZOU2026104305}. \begin{revblockv2}The 6 defense papers (6.5\% of the 92 analyzed defense papers) that address Pre-execution Attacks, covering supply chain vulnerabilities and poisoned plugins, represent the class of threats most likely to undermine isolation boundaries before they are fully instantiated.\end{revblockv2}These boundaries fail when the underlying environment API surface itself becomes a target for privilege escalation or data exfiltration.

\begin{revblockv2}
\textbf{Cryptographic / Structural Integrity} defenses preserve data provenance and trace execution causality using unalterable logs and strict formatting rules \citep{lee2024promptinfectionllmtollmprompt, ipiguard2025}. Security tools deploy provenance-based intrusion detection and tool dependency graphs to maintain explicit prompt-data separation. 35 of the 92 analyzed defense papers (38.0\%) address Poisoning \& Extraction Attacks, the supply-chain vector that most directly challenges provenance assumptions. Adaptive attackers can still bypass these structures through format-mimicking tokens, making reactive logging insufficient for prevention.

\textbf{Human in the Loop (HITL)} serves as the ultimate failsafe by inserting manual verification checkpoints before high-stakes or irreversible actions \citep{policyguidedthreat2026, yang2025cloudinfrastructuremanagement}. Architectures employ threshold-based pauses and multi-agent supervisor frameworks to escalate risky tool executions to human operators. 7 of the 92 analyzed defense papers (7.6\%) address PromptInfection, which models self-replicating lateral propagation across multi-agent systems and represents scenarios where human checkpoints are the last viable containment layer, given that perimeter defenses are entirely blind to steganographic inter-agent dialogue. This approach introduces severe scalability bottlenecks and risks rubber-stamping behaviors caused by widespread alert fatigue. \looseness=-1
\end{revblockv2}

\subsubsection{Cost and Security Trade-offs in Agentic Defense Systems}

In this section we analyze how recent studies report the trade-off between security gains and operational costs in agentic AI systems. Fig. \ref{subfig:cost_security_tradeoffs} presents the distribution of papers across the four reporting categories. The largest group treats cost only qualitatively or theoretically, while a smaller set provides comprehensive reporting that includes security, latency, utility and cost metrics.

\begin{revblockv2}The \textbf{Comprehensive Cost Reporting} category, comprising 18 papers (19.6\%), represents the most complete evaluation style.\end{revblockv2}These studies jointly measure attack mitigation, utility preservation, token consumption and latency overhead. Systems such as IPIGUARD demonstrate that layered defenses can improve security while reducing unnecessary computational cost \citep{ipiguard2025}. In practice, IPIGUARD reduces the Attack Success Rate (ASR) below 1\% at a cost of 14,605 tokens per task and 13.88 seconds of latency, preserving 67.7\% benign utility. A common finding is that lightweight filters, such as the PPA small model operating at approximately 0.06 ms with zero token overhead, should screen requests before expensive LLM-based reasoning is triggered.

\begin{revblockv2}The \textbf{Security and Utility Focus} category covers 23 papers (25.0\%) and prioritizes \end{revblockv2}maintaining agent performance during normal operation. Studies such as AgentDojo investigate how defenses affect task completion and false refusal behavior while paying less attention to infrastructure latency \citep{debenedetti2024agentdojo}. These works show that stronger alignment mechanisms can reduce attack success rates, but they may also decrease performance on complex multi-step tasks when the defense becomes overly restrictive. The core contradiction here is well-established: techniques like Sandwich Prevention effectively suppress ASR but frequently cause catastrophic forgetting on long-horizon agentic workflows.

\begin{revblockv2}The \textbf{Security and Compute Focus} category covers 15 papers (16.3\%) and treats \end{revblockv2}security as a scalability challenge. Research such as GuardAgent, MCP-GUARD and R$^2$-Guard concentrates on detection quality, token usage and execution cost \citep{xiang2025guardagentsafeguardllmagents, mcpguard2025, kang2024r2guard}. The costs here are substantial: GuardAgent achieves 81.4\% accuracy consuming 6,116 tokens per task over 6.30 seconds, and even MCP-GUARD adds roughly 455.9 ms of overhead for a +23.4\% F1 improvement. These systems use triage pipelines, specialized agents and data reduction techniques to lower processing costs before invoking large models, though adversaries can still exploit the cheaper triage layers.

\begin{revblockv2}The \textbf{Qualitative and Theoretical} category accounts for 36 papers (39.1\%), the largest group, and focuses \end{revblockv2} on conceptual security frameworks and architectural guidance. Works in this group discuss zero-trust principles, compartmentalization and continuous monitoring for multi-agent environments \citep{dewitt2025openchallengesmultiagentsecurity}. However, many proposals remain theoretical and lack large-scale empirical validation to demonstrate their operational feasibility. \looseness=-1

To better understand the relationship between security gains and operational overhead, representative metrics from the literature are summarized in Table~\ref{tab:tradeoff_metrics}. \newrev{It should be noted that the purpose of the table is to illustrate the range of overheads reported in the literature. These systems solve different problems on different workloads and are not mutually comparable, so any ranking of security gain or latency and financial cost from the table would be uninformative.} \begin{revblockv2}This is the same metric-comparability problem discussed for attack-success-rate reporting in \S\ref{subsubsec:benchmark_fragmentation}: a number taken out of its task, workload and hardware context is not meaningful next to another number that merely shares a column header.\end{revblockv2}

\begin{table}[ht]
\centering
\caption{\begin{revblockv2}Illustrative range of security-cost magnitudes reported across defense mechanisms. Each row is measured on its own paper's task, workload and hardware; the rows are not mutually comparable and this table is not a ranking.\end{revblockv2}}
\label{tab:tradeoff_metrics}
\small
\begin{tabular}{|l|l|l|l|l|}
\hline
\textbf{Defense Mechanism} & \textbf{Security Gain} & \textbf{Latency Cost} & \textbf{Financial Cost} & \textbf{Utility} \\
\hline
PPA (Small Model) & High (Bypass Resistant) & $\sim$0.06 ms & No Overhead & Minimal \\
MCP-GUARD & +23.4\% F1 Score & $\sim$455.9 ms & Low & High Utility \\
IPIGUARD & ASR $<$ 1\% & $\sim$13.88 s & 14,605 Tokens per Task & 67.7\% \\
PhishDebate & 94.14\% Accuracy & $\sim$37.50 s & High & N/A \\
GuardAgent & 81.4\% Accuracy & $\sim$6.30 s & 6,116 Tokens per Task & Moderate \\
PROVSEEK & 0.92 F1 Score & $\sim$30 min & 440K Tokens & N/A \\
\hline
\end{tabular}
\end{table}

\begin{revblockv2}The results show substantial variation across approaches, each evaluated on its own task and workload. Latency ranges from sub-millisecond filtering mechanisms to forensic systems that require several minutes of processing time, and financial cost ranges from no measurable overhead to hundreds of thousands of tokens per task. This spread illustrates the range of overheads defenses report, not a ranking of which system is best. MCP-GUARD's low latency and IPIGUARD, GuardAgent and PROVSEEK's higher latency and token consumption reflect each system's own design point on its own workload rather than points on a shared trade-off curve \citep{mcpguard2025,ipiguard2025,xiang2025guardagentsafeguardllmagents,mukherjee2025llmdrivenprovenanceforensicsthreat}.\end{revblockv2}

\subsubsection{Defense Placement Across the Agent Lifecycle}\label{subsubsec:defense-placements}

\begin{figure}[htbp]
    \centering
    \begin{subfigure}{0.68\linewidth}
        \centering
        \includegraphics[width=\linewidth]{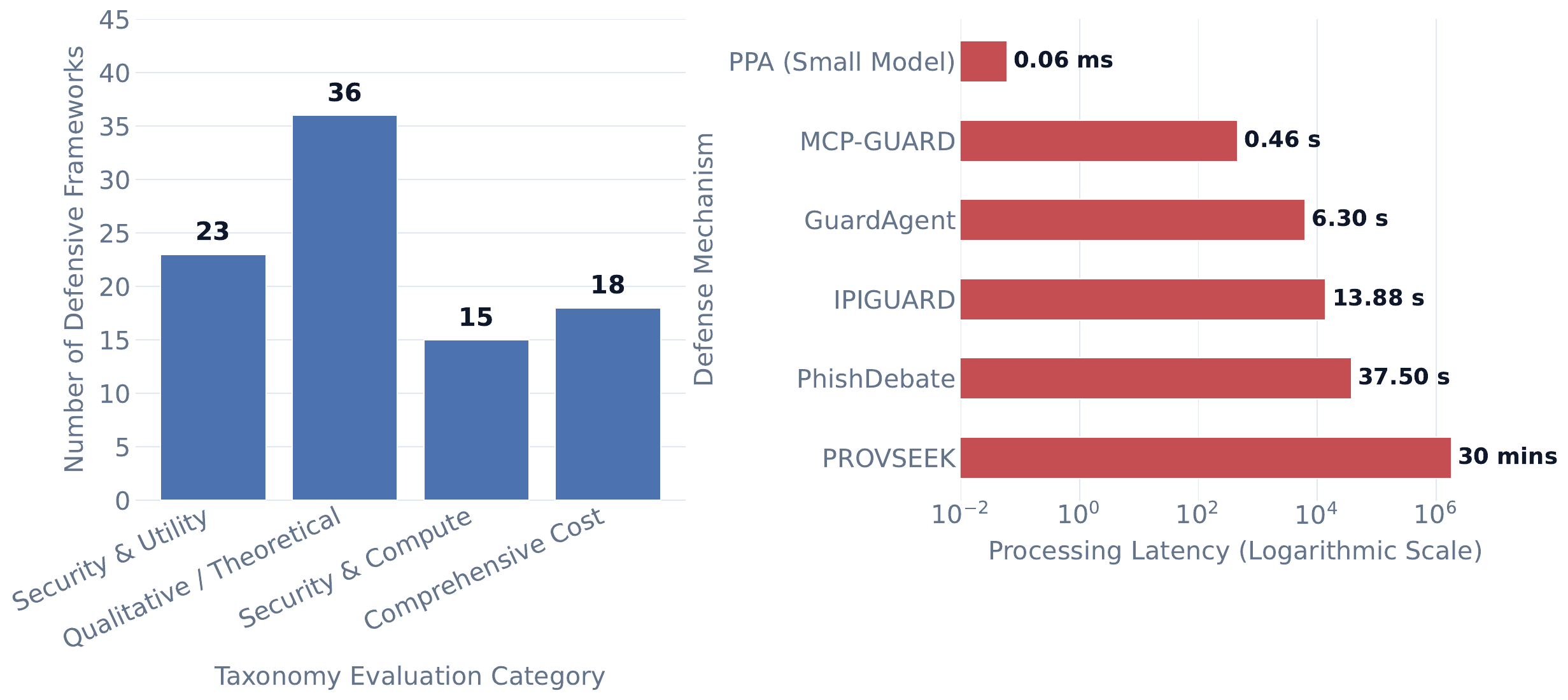}
        \caption{Methodological Prioritization and Computational Cost Barriers in LLM Defenses.}
        \label{subfig:cost_security_tradeoffs}
    \end{subfigure}
    \hfill
    \begin{subfigure}{0.26\linewidth}
        \centering
        \includegraphics[width=\linewidth]{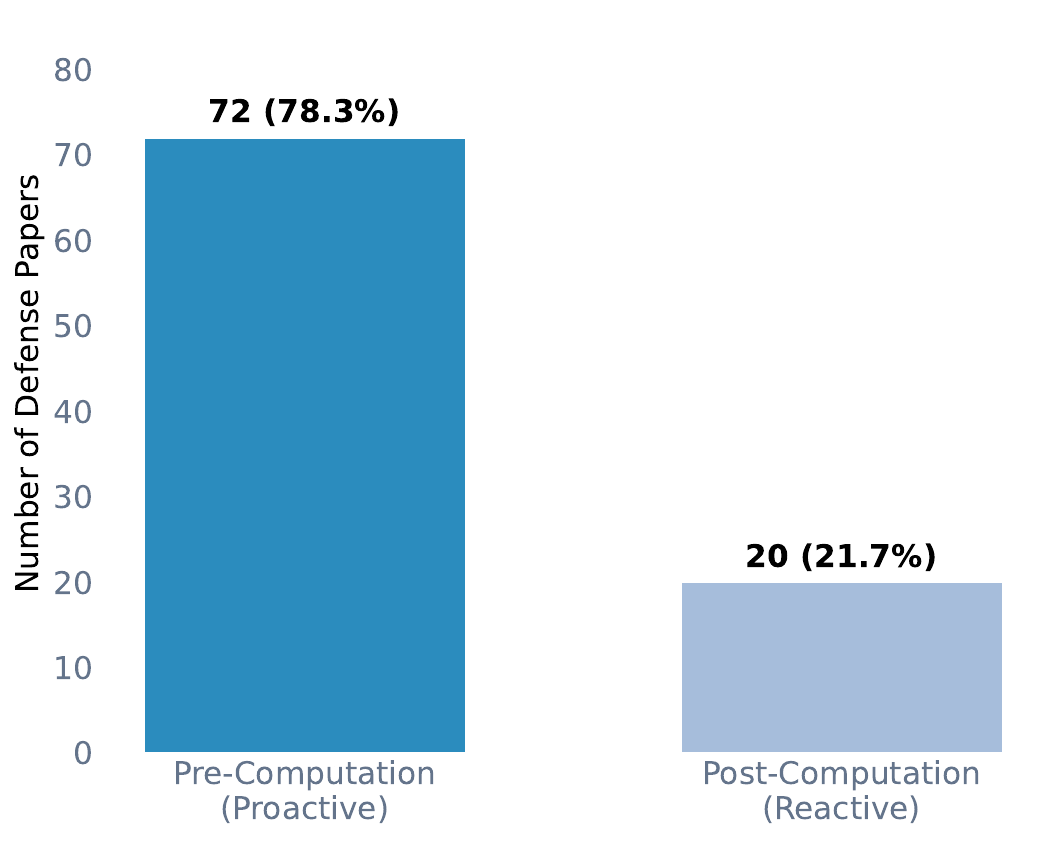}
        \caption{Temporal Analysis of LLM Defense Mechanisms (Proactive vs. Reactive).}
        \label{subfig:proactive_reactive}
    \end{subfigure}
    \caption{Operational Profiling: Computational Cost-Security Trade-offs and Temporal Intervention Points in LLM Defenses.}
    \label{fig:cost_and_temporal_analysis}
\end{figure}

\begin{revblockv2}This section maps where defenses are deployed within the autonomous agent lifecycle across 92 surveyed papers, assigning each paper the point at which its primary defense intervenes. Of these, 72 papers (78.3\%) place their primary defense before computation and 20 papers (21.7\%) place it after computation. \end{revblockv2}Pre-computation defenses therefore outnumber post-computation defenses by more than three to one. Figure~\ref{subfig:proactive_reactive} summarizes this distribution.

\textbf{Pre-Computation Defenses.}
Pre-computation defenses prevent malicious content from influencing the agent before reasoning begins. Common techniques include prompt hardening, input sanitization, dynamic prompt rewriting and adversarially robust alignment. Studies such as \citep{lee2024promptinfectionllmtollmprompt} emphasize separating trusted instructions from untrusted external content before it enters the context window. \begin{revblockv2}Yet these 72 papers (78.3\%) collectively reveal a shared limitation: \end{revblockv2}static filtering remains vulnerable to adaptive adversaries that use semantic obfuscation or embed malicious instructions inside external resources the agent must process, making complete prevention difficult without reducing utility.

\textbf{Post-Computation Defenses.}
Post-computation defenses operate after information enters the reasoning pipeline, assuming some attacks will bypass preventive controls. Systems such as AgentSentry and ShieldAgent implement tool-call auditing, execution monitoring, semantic rollback and policy verification \citep{agentsentry2026, shieldagent2025}. AgentSentry demonstrates this through causal diagnostics that identify malicious intent shifts and remove contaminated reasoning traces before execution \citep{agentsentry2026}. \begin{revblockv2}The 20 papers (21.7\%) in this group reflect \end{revblockv2}an ``assume breach'' mindset, where post-computation monitoring provides a critical secondary security layer.

\textbf{Tradeoffs.}
\begin{revblockv2}Because each paper is assigned a single primary placement, this distribution does not measure how many defenses combine both layers. \end{revblockv2} Both strategies exhibit a robustness-utility tradeoff: strong preventive controls may block legitimate information while aggressive monitoring may interrupt valid workflows. Post-computation mechanisms also struggle against subtle memory poisoning attacks where corrupted information gradually alters future reasoning without triggering an explicit policy violation. \begin{revblockv2}The strong tilt toward pre-computation controls suggests \end{revblockv2}that post-computation monitoring remains comparatively underexplored as a primary defense against complex and persistent threats.\looseness=-1

\section{Cross-Cutting Analysis and Trends}\label{sec:analysis}

This section provides a cross-cutting analysis of the security agent literature. We evaluate the collected papers across four primary dimensions such as defense coverage across attack scenarios, structural and foundational analysis, language model selection and security evaluations, and data modalities.

\begin{revblock}

\begin{figure}[ht]
    \centering
    \includegraphics[width=\textwidth]{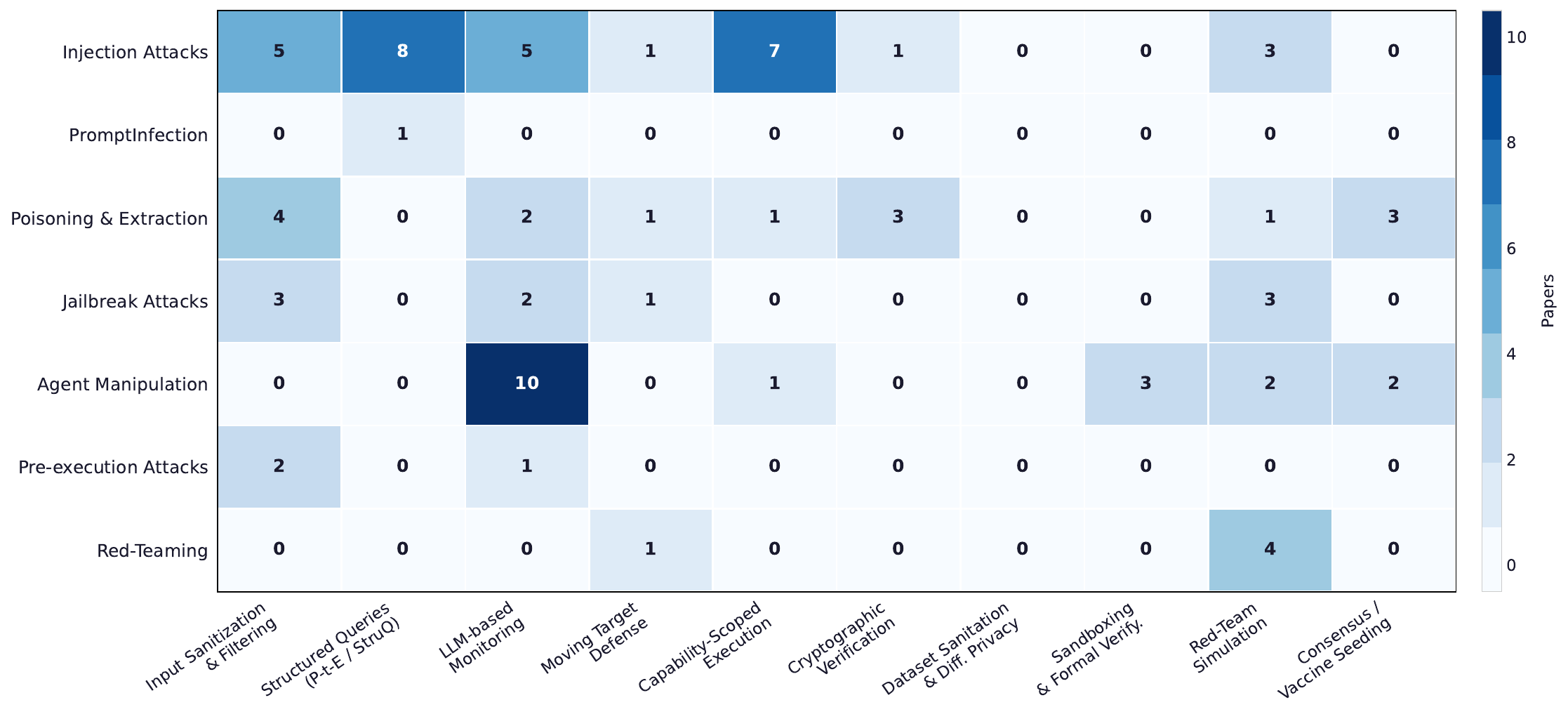}
    \caption{Heatmap of each defense paper's primary attack category against its primary
    defense mechanism, drawn from the 81 defense papers that have a primary attack category.
    Each paper is assigned a single primary attack category and a single primary
    defense mechanism rather than every category it touches on, so the cells sum
    to exactly 81.}
    \label{fig:attack_def_heatmap}
\end{figure}

\subsection{Defense Coverage Across Attack Categories}

The objective of this analysis is to understand which attack classes receive the most defensive attention and to identify areas where current defenses remain limited. The results also highlight important overlaps between attack categories. Figure~\ref{fig:attack_def_heatmap} shows the distribution across seven attack categories. \newrev{It should be noted that the cell values report defensive attention, not measured effectiveness. Each of the 81 defense papers with a primary attack category is assigned exactly one primary attack category and one primary defense mechanism, the pairing the paper is centrally concerned with, rather than every attack or defense category it touches on.} The section tries to extensively analyze the attack corpus which goes beyond the 10 defense categories we mentioned in the figure.

\textbf{Injection Attacks} are the most common primary attack category, accounting for 30 papers. Studies such
as SecAlign and AgentSentry reduce indirect prompt injection risks through instruction
isolation and temporal diagnostics \cite{metasecalign2025,agentsentry2026}, though
stronger defenses consistently reduce task performance. \newrev{Structured queries are the most common primary defense against injection attacks (8 papers), ahead of capability-scoped execution (7 papers).}

\textbf{Agent Manipulation Attacks} account for 18 papers, the second most common primary attack category,
focusing on tool misuse, environment manipulation and privilege escalation. Execution-stage
controls are essential here because attacks exploit the interaction between planning and
tool usage \cite{tamingopenclaw2026}, yet delayed execution attacks remain difficult
to trace. \newrev{LLM-based monitoring is the most frequently applied primary defense against agent manipulation (10 papers).}

\textbf{Jailbreak Attacks} account for 9 papers. Research on MASTERKEY and R$^2$-Guard
demonstrates both the sophistication of modern jailbreak attacks and the need for
stronger monitoring \cite{masterkey2024,kang2024r2guard}, with adaptive multi-turn
attacks still evading many existing protections. \newrev{Red-team simulation and input sanitization are tied as the most frequently applied primary defenses against jailbreak attacks (3 papers each).}

\textbf{Red-Teaming Attacks} account for 5 papers. Frameworks such as ToolFuzz
proactively generate attack scenarios to
uncover weaknesses before deployment \cite{milev2025toolfuzzautomatedagent}. \newrev{Red-team simulation is the most frequently applied primary defense against red-teaming attacks (4 papers), ahead of moving target defense (1 paper).}

\textbf{Poisoning and Extraction Attacks} account for 15 papers. AgentPoison illustrates
how malicious content inserted into agent memories can influence future behavior
\cite{chen2024agentpoison}, with retrieval repositories and external skill ecosystems
identified as major trust assumptions. \newrev{Input sanitization is the most frequently applied primary defense against poisoning and extraction attacks (4 papers).}

\textbf{Pre-execution Attacks} account for 3 papers, the second least common primary attack category, focusing on poisoned plugins and
compromised supply chains. Malicious artifacts introduced during initialization can
influence behavior long before runtime defenses activate \cite{chen2024agentpoison},
with verifiable provenance for third-party tools remaining an open challenge. \newrev{Input sanitization is the most frequently applied primary defense against pre-execution attacks (2 papers).}

\textbf{PromptInfection} accounts for 1 paper, the least common primary attack category. The original
Prompt Infection work shows how adversarial prompts spread across multi-agent
environments and compromise downstream workflows \cite{lee2024promptinfectionllmtollmprompt},
with covert communication between compromised agents still difficult to detect. \newrev{Its only primary defense is a structured query defense that tags the source of each inter-agent message (1 paper).}

\end{revblock}

\subsection{Structural and Functional Analysis of Agentic AI Systems}

\begin{figure}[ht]
    \centering
    
    \begin{subfigure}{0.38\linewidth}
        \centering
        \includegraphics[width=\linewidth]{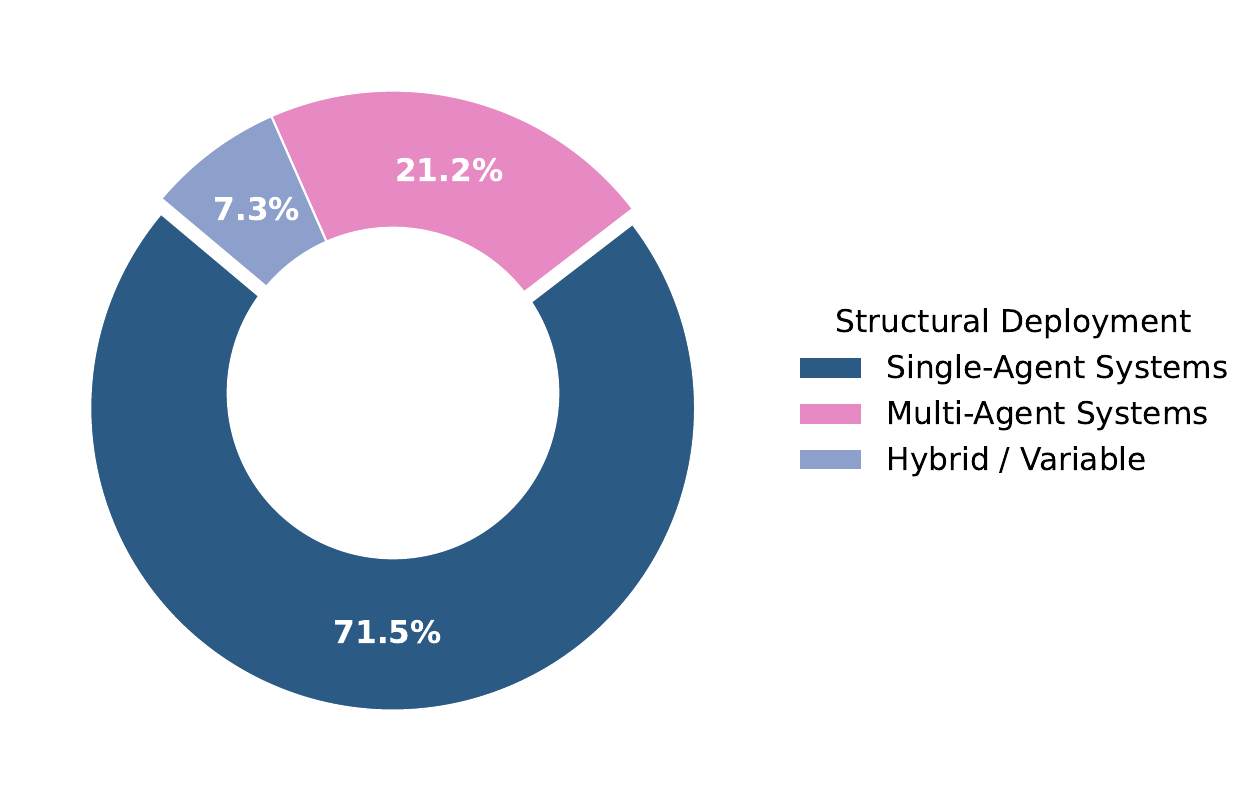}
        \caption{Structural Complexity and Cardinality Distribution of Evaluated Agent Environments.}
        \label{subfig:agent_distribution}
    \end{subfigure}
    \hfill
    \begin{subfigure}{0.58\linewidth}
        \centering
        \includegraphics[width=\linewidth]{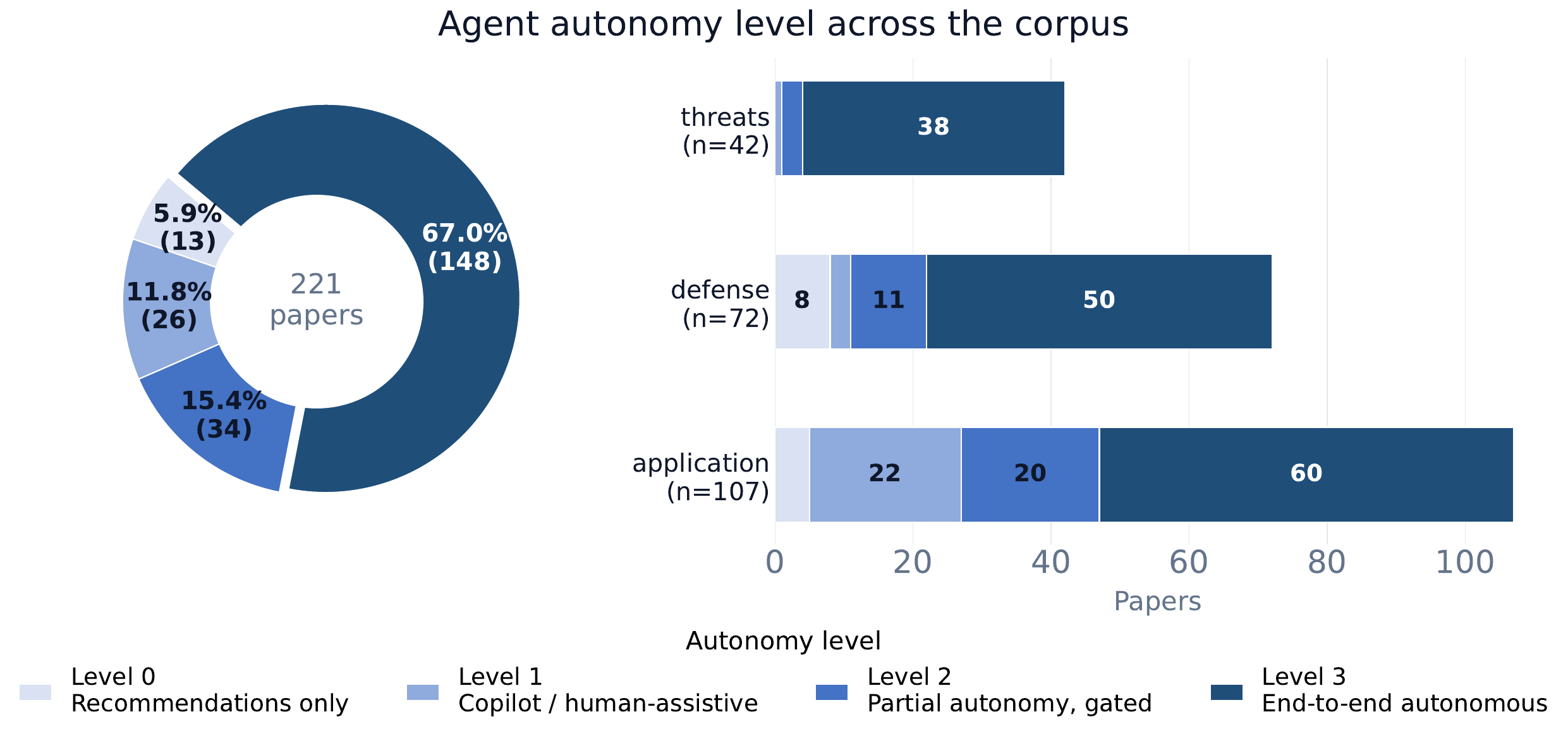}
        \caption{Evaluation of Human Oversight Boundaries and Operational Autonomy Modes in Agentic Systems.}
        \label{subfig:autonomy_chart}
    \end{subfigure}
    
    \caption{Structural Mapping of Agent Environments: Empirical Analysis of Multi-Agent Cardinality and Execution Autonomy Profiles.}
    \label{fig:agent_structural_autonomy_analysis}
\end{figure}

We tried to dive deep into the structural analysis of the Agentic AI corpus and classified all the papers across four dimensions: agent cardinality, autonomy level, architectural paradigm and cognitive role distribution, revealing how the field organizes intelligence, delegates decision-making and distributes security responsibility. \looseness=-1

\textbf{Agent Cardinality.}

\begin{revblockv2}
Figure~\ref{subfig:agent_distribution} shows that single-agent systems dominate at 196 papers (71.5\%), followed by multi-agent systems at 58 papers (21.2\%) and hybrid frameworks at 20 papers (7.3\%).
\end{revblockv2} \citet{Levy2025STWebAgentBench} benchmark single-agent architectures on safety and trustworthiness in web tasks, showing that reasoning loops like ReAct \citep{yao2023reactsynergizingreasoningacting} and Reflexion \citep{shinn2023reflexionlanguageagentsverbal} face severe constraints in long-horizon execution due to isolated memory contexts. These systems leverage strict execution sandboxing to prevent permission overstepping but lack cross-validation mechanisms, leaving them vulnerable to prompt injections and resource exhaustion \citep{shieldagent2025}. \citet{pan2025why} investigate why multi-agent systems fail, showing that decentralized deployments introduce token overhead and latency bottlenecks, where rogue nodes bypass trust boundaries to trigger cascading failures. Hybrid frameworks target universal benchmarking and platform infrastructure \citep{aiagentindex2026, ma2025safetyatscale, mei2025aios} but struggle to engineer guardrails that scale without degrading cooperative utility.

\begin{revblock}
\textbf{Agent Autonomy Levels.}
We classify every paper in the corpus by the authority its central system is designed to hold: Level~0 (recommendations only), Level~1 (copilot assisting a human), Level~2 (partial autonomy with approval at critical points), or Level~3 (end-to-end autonomous). Autonomy level is not a meaningful property of a paper with no system of its own, so 53 papers, mostly benchmarks evaluating third-party agents, surveys, and theoretical or position papers with no deployed system of their own, are excluded from this analysis. Of the remaining 221 papers, Figure~\ref{subfig:autonomy_chart} shows Level~3 dominates at 148 papers (67.0\%), followed by Level~2 at 34 papers (15.4\%), Level~1 at 26 papers (11.8\%), and Level~0 at 13 papers (5.9\%). \citet{yao2023reactsynergizingreasoningacting} demonstrate that Level~3 agents rely on continuous execution cycles, yet \citet{fang2024llmagentsautonomouslyexploit} show these systems compound hallucinations and violate least-privilege principles, making them highly susceptible to automated exploits and data poisoning. \citet{mei2025aios} show that Level~2 systems enforce check-and-approve mechanisms and access control to decouple reasoning from catastrophic execution errors, though the added human interventions introduce overhead and latency \citep{ZOU2026104305}. \citet{tur2025safearena} and \citet{andriushchenko2025agentharm} evaluate systems that treat autonomy as a configurable spectrum using sandbox benchmarks and attribute-based access control, while \citet{deng2024aiagentsthreatsurvey} survey the key security challenges these frameworks must address. Level~0 and~1 systems remain the most secure and drift-free but cannot scale against automated cyberattacks.

\end{revblock}

\textbf{Architectural Paradigms.}

\begin{figure}[htbp]
    \centering
    
    \begin{subfigure}{0.35\linewidth}
        \centering
        \includegraphics[width=\linewidth]{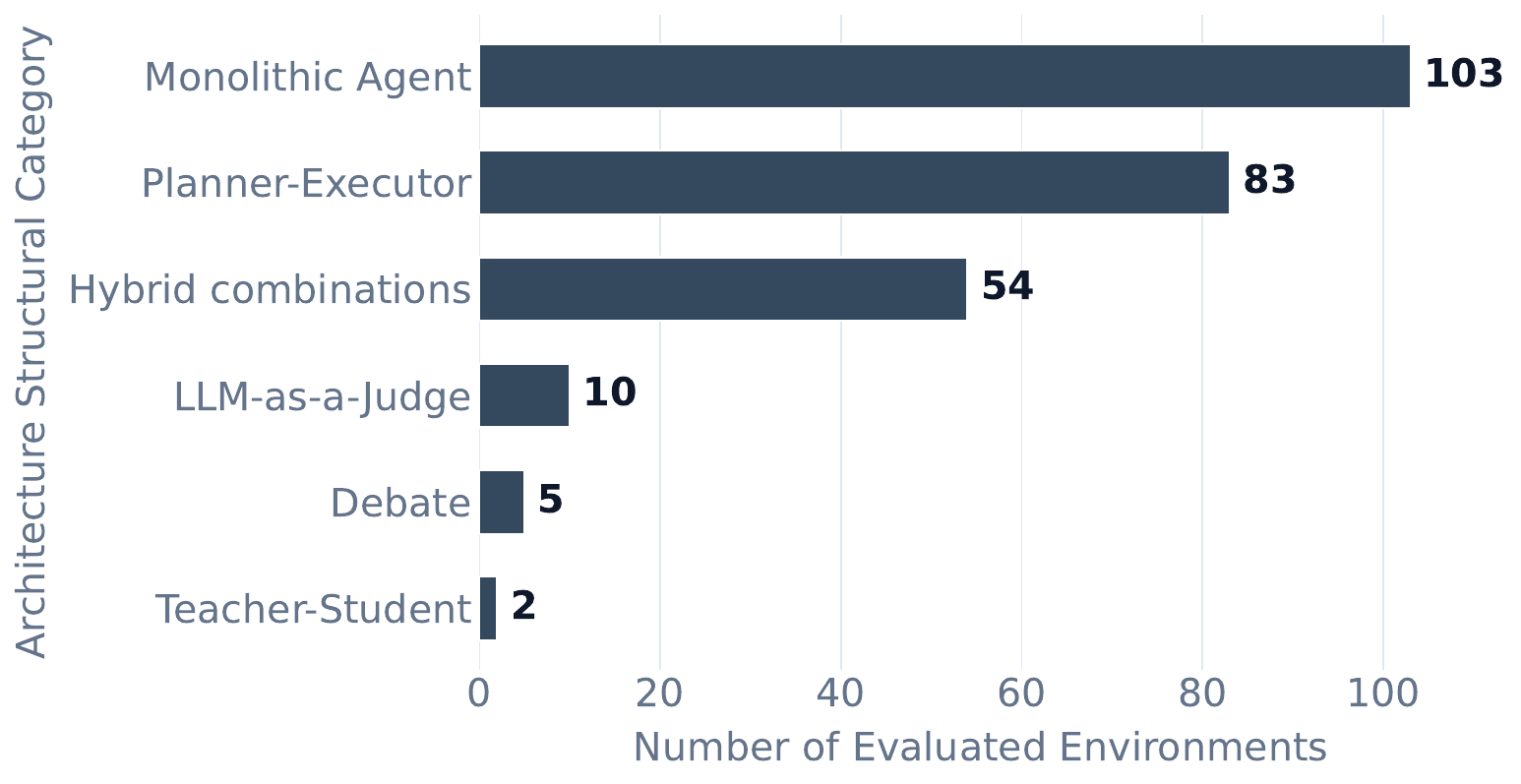}
        \caption{Empirical Census of Agent Architectural Paradigms within Evaluation Frameworks.}
        \label{subfig:arch_dist}
    \end{subfigure}
    \hfill
    \begin{subfigure}{0.59\linewidth}
        \centering
        \includegraphics[width=\linewidth]{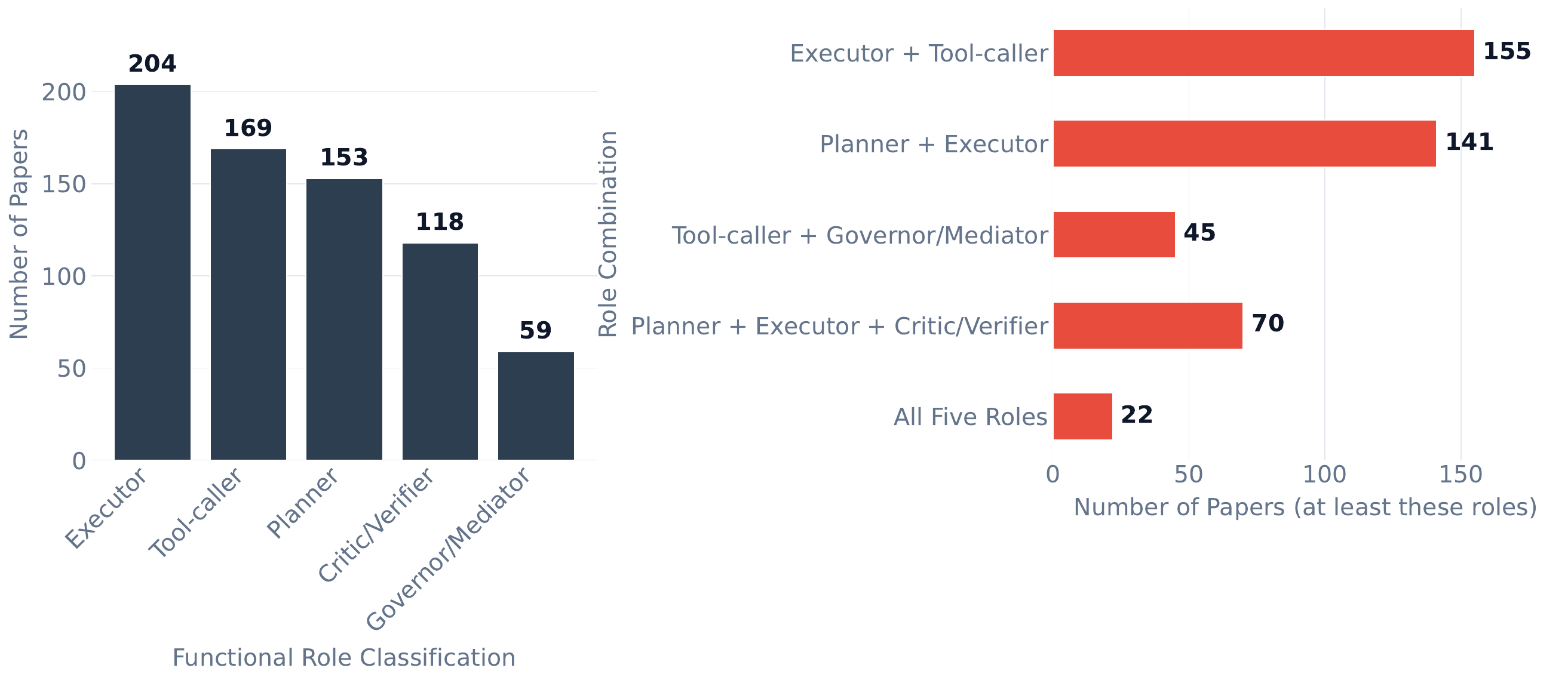}
        \caption{Role Differentiation and Functional Intersections within Multi-Agent Architectures.}
        \label{subfig:role_intersection}
    \end{subfigure}
    
    \caption{Architectural and Behavioral Profiling: Core Management Paradigms and Role Differentiation in Multi-Agent Systems.}
    \label{fig:agent_architecture_roles_analysis}
\end{figure}

\begin{revblockv2}
\textbf{Architectural Paradigms.}
Of the 274 papers in the corpus, 17 have no identifiable architectural paradigm and are excluded. The remaining 257 span six paradigms: monolithic (103 papers, 40.1\%), planner-executor (83 papers, 32.3\%), hybrid systems (54 papers, 21.0\%), LLM-as-a-Judge (10 papers, 3.9\%), debate-centric (5 papers, 1.9\%) and teacher-student (2 papers, 0.8\%) as illustrated in Figure ~\ref{subfig:arch_dist}.
\end{revblockv2} \citet{delrosario2025architectingresilientllmagents} and \citet{huang2024penheal} show that planner-executor architectures map goals into deterministic state machines for control-flow integrity but suffer re-planning latency under zero-day conditions. Monolithic agents rely on unified ReAct loops \citep{yao2023reactsynergizingreasoningacting} and prompt engineering \citep{schick2023toolformer} yet remain vulnerable to context exhaustion and prompt injection. \citet{li2025phishdebatellmbasedmultiagentframework} demonstrate that debate-centric systems improve factual accuracy through quorum-based critique protocols, though high token costs and sycophancy risks persist. \citet{rjudge2024} show that LLM-as-a-Judge pipelines automate safety evaluation across benchmarks yet exhibit self-preference biases. Hybrid systems handle enterprise-scale orchestration but untracked error propagation allows a single compromised sub-agent to trigger collective failure spirals.

\textbf{Cognitive Role Distribution.}

\begin{revblockv2}
\textbf{Cognitive Role Distribution.}
Of the 274 papers in the corpus, 44 have no cognitive role coded and are excluded. Figure~\ref{subfig:role_intersection} records 703 total role assignments across the remaining 230 papers, averaging 3.06 roles per paper, confirming that modern pipelines are inherently multi-functional. Executors appear in 204 papers (88.7\%) and run inside sandboxed environments to contain malicious fallout, yet remain vulnerable to data poisoning and indirect prompt injections \citep{kong2025vulnbotautonomouspenetrationtesting}. \citet{chen2025agentguardrepurposingagenticorchestrator} demonstrate that tool-callers, present in 169 papers (73.5\%), secure API boundaries through JSON schema validation and zero-trust registries, though malicious payloads frequently hide behind valid schemas. \citet{delrosario2025architectingresilientllmagents} show that planners, appearing in 153 papers (66.5\%), implement hierarchical task management to decouple intent from execution, though they remain vulnerable under adversarial re-planning conditions \citep{informationflow2025}. Critics appear in 118 papers (51.3\%) and deploy automated judge pipelines to filter false positives, but \citet{finetuningagents2024} and \citet{he2026coredteamorchestratedsecuritydiscovery} identify recursive blind spots and collusion risks in multi-agent verification loops. \citet{shieldagent2025} and \citet{shi2025progent} show that governors, present in 59 papers (25.7\%), enforce least-privilege boundaries through taint tracking and information-flow control, though strict deterministic enforcement frequently degrades agent utility.
\end{revblockv2}

\begin{figure}[htbp]
    \centering
    
    \begin{subfigure}{0.85\linewidth}
        \centering
        \includegraphics[width=\linewidth]{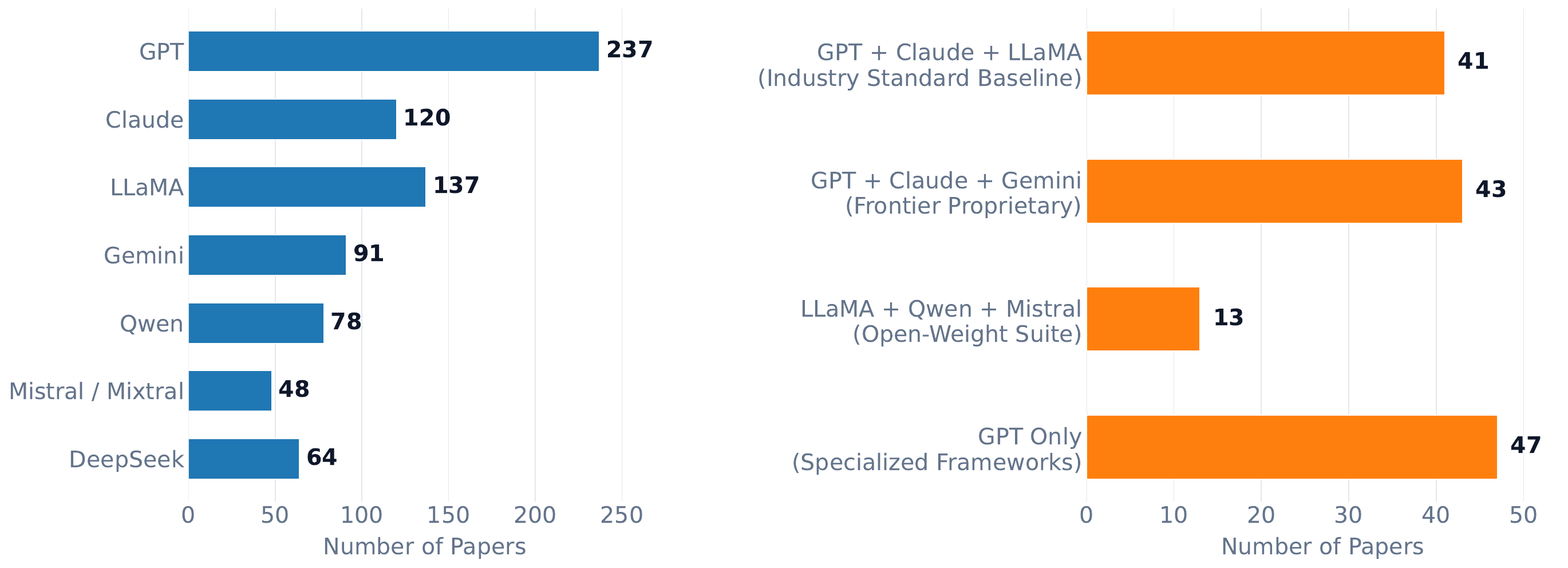}
        \caption{Marketplace Saturation and Evaluation Synergy of Base Foundation Models.}
        \label{subfig:llm_overlap}
    \end{subfigure}
    
    \vspace{1em}
    
    \begin{subfigure}{0.85\linewidth}
        \centering
        \includegraphics[width=\linewidth]{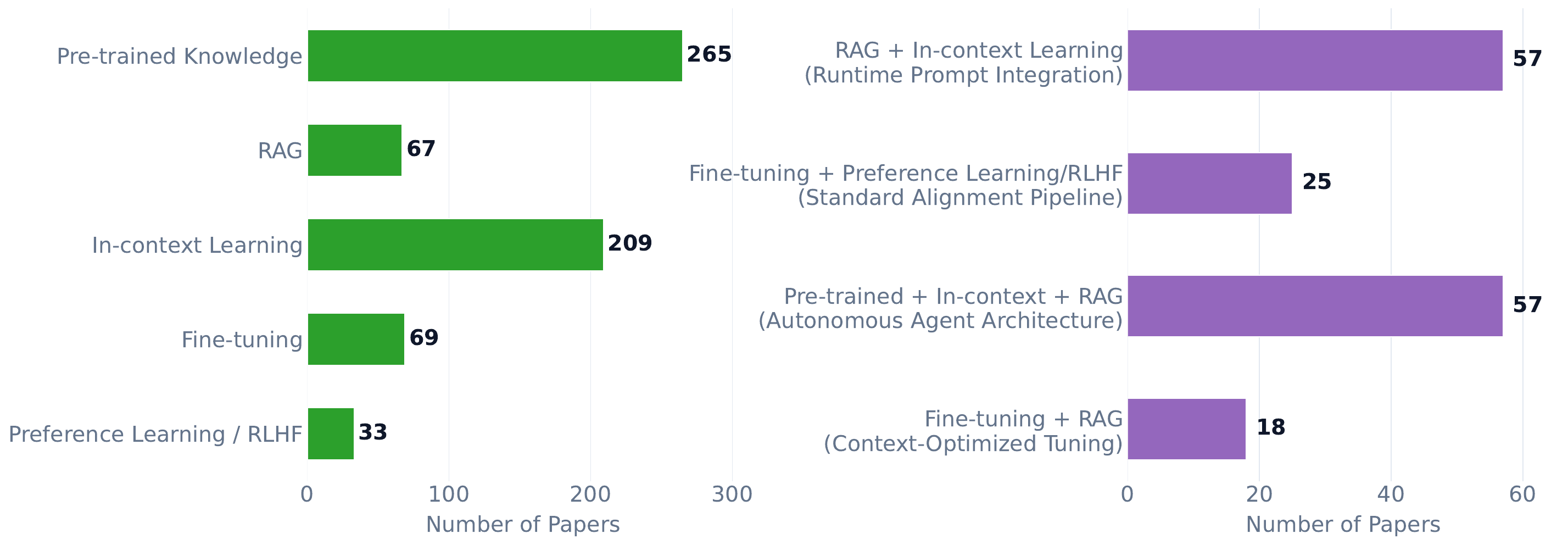}
        \caption{Methodological Taxonomy and Intersectional Pipelines of Knowledge Configurations.}
        \label{subfig:knowledge_sources_plot}
    \end{subfigure}
    
    \caption{Empirical Profiling of Foundation Model Deployment Saturation and Underlying Knowledge Source Infrastructures.}
    \label{fig:agent_llm_knowledge_landscape}
\end{figure}

\begin{revblock}
\subsection{Foundation Model Usage Patterns and Security Evaluations}

This subsection details the empirical distribution of large language model families and knowledge acquisition paradigms across 274 security research papers. The 775 model-family assignments span three roles per family (evaluation subject, system component and discussed only), with a family appearing in at most one role per paper. The 643 knowledge-source assignments are non-mutually exclusive, reflecting that most studies combine several learning strategies within a single pipeline. The findings are illustrated in Figure~\ref{fig:agent_llm_knowledge_landscape}.

\textbf{GPT Models.}
GPT variants appear in 237 papers (86.5\%), making them the near-universal capability baseline (190 as an evaluation subject, 22 as a backbone/judge component, 25 discussed only). \citet{debenedetti2024agentdojo} construct a dynamic prompt injection benchmark evaluated across GPT, Claude, Gemini, and LLaMA, and \citet{Deng2024PentestGPT} harness GPT for penetration testing frameworks. \citet{fu2024impromptertrickingllmagents} use GPT as the backbone of the attacked agent while benchmarking transferability against Mistral/Mixtral and LLaMA targets, with 47 papers evaluating GPT in isolation.

\textbf{Claude Models.}
Claude appears in 120 papers (43.8\%): 83 as an evaluation subject, 4 as a backbone/judge component, and 33 discussed only. 43 papers benchmark Claude alongside GPT and Gemini as a frontier proprietary evaluation suite.

\textbf{LLaMA Models.}
LLaMA appears in 137 papers (50.0\%) and anchors open-weight deployment risk analysis: 102 as a direct evaluation subject, 4 as a component, and 31 discussed at survey level. \citet{gan2024navigatingriskssurveysecurity} survey security and privacy threats unique to open-weight deployments, discussing LLaMA and GPT without directly benchmarking either, while \citet{yang2024watch} evaluate LLaMA directly to investigate backdoor threats introduced via fine-tuning and weight manipulation.

\textbf{Gemini Models.}
Gemini appears in 91 papers (33.2\%) and anchors multimodal security research: 66 as an evaluation subject, 2 as a component, and 23 discussed only. \citet{wang2025agentvigilgenericblackboxredteaming} evaluate Gemini alongside GPT, Claude, LLaMA, and Qwen in a black-box red-teaming pipeline for indirect prompt injection, with multimodal exposure surfacing image-borne injection channels that bypass text-only guardrails.

\textbf{Qwen Models.}
Qwen appears in 78 papers (28.5\%) and is heavily utilized in code execution and exploit generation: 68 as an evaluation subject, 1 as a component, and 9 discussed only. \citet{zhuo2026cyberzero} train cybersecurity agents using Qwen's open-weight coding capabilities alongside Claude, DeepSeek, and Gemini, combining pre-trained knowledge with task-specific fine-tuning.

\textbf{Mistral and Mixtral Models.}
Mistral and Mixtral appear in 48 papers (17.5\%) as an efficiency-oriented alternative: 34 as an evaluation subject, 3 as a component, and 11 discussed only, most often paired with other open-weight families rather than evaluated alone. \citet{meng2025uncoveringvulnerabilitiesllmassistedcyber} evaluate Mistral/Mixtral alongside GPT, Claude, Gemini, LLaMA, and Qwen in a threat-intelligence extraction pipeline spanning both the applications and defenses pillars, illustrating the family's role as one entry in broad, defense-relevant model comparisons rather than a standalone case study.

\textbf{DeepSeek Models.}
DeepSeek appears in 64 papers (23.4\%): 52 as an evaluation subject, 3 as a component, and 9 discussed only, concentrated in exploit-generation and incident-response pipelines that exercise extended reasoning. \citet{chen2026vulnsage} evaluate DeepSeek alongside GPT and Qwen in a multi-agent exploit-generation framework built around constraint-guided comprehension and iterative reflection, using the planning/reflection loop itself as the mechanism for improving exploit precision. This is the same extended-reasoning surface that a security-focused reading of this literature flags as an emerging target for planning-phase manipulation.

\textbf{Pre-trained Knowledge.}
Pre-trained knowledge appears in 265 papers (96.7\%) and provides the foundational reasoning capability nearly every agent in the corpus builds on. \citet{li2025chainofagentsendtoendagentfoundation} combine pre-trained knowledge with fine-tuning, in-context learning, and RAG to extend base model capabilities via multi-agent distillation and agentic reinforcement learning, illustrating how pre-training is treated as a substrate layered with other knowledge sources rather than a standalone paradigm.

\textbf{In-Context Learning.}
In-context learning appears in 209 papers (76.3\%) and elicits multi-step planning without weight modification. It is the most frequently paired knowledge source after pre-trained knowledge, co-occurring with RAG in 57 papers, as retrieved data requires prompt-based integration to be actionable. Context window limitations and output fragility under prompt variation remain its primary weaknesses in long-horizon agentic workflows.

\textbf{Retrieval-Augmented Generation.}
RAG appears in 67 papers (24.5\%) and links static parameters with external databases to minimize hallucinations. \citet{Blefari2026cyberrag} build a modular agentic RAG framework for attack classification and \citet{mukherjee2025llmdrivenprovenanceforensicsthreat} deploy retrievers over provenance graphs for forensic investigation. These systems nevertheless remain vulnerable to vector database poisoning, with 18 papers combining RAG with fine-tuning to train models that more reliably utilize retrieved context.

\textbf{Fine-tuning.}
Fine-tuning appears in 69 papers (25.2\%) and permanently modifies weights to enforce consistent structure and narrow expertise via parameter-efficient techniques like LoRA. 25 papers pair fine-tuning with preference learning as a standard alignment pipeline, though this paradigm requires expensive curated datasets and risks catastrophic forgetting.

\textbf{Preference Learning and RLHF.}
Preference learning appears in 33 papers (12.0\%) and aligns agent behaviors with human values of safety, helpfulness, and harmlessness. \citet{fu2025rewardshapingmitigatereward} demonstrate that reward shaping mitigates reward hacking, where agents exploit misspecified proxy functions through verbose or sycophantic generation. 57 papers combine pre-trained knowledge, in-context learning, and RAG as the standard autonomous agent architecture, with preference learning addressing the residual alignment gap in the smaller subset of papers where it appears at all.

\end{revblock}

\subsection{Data Modalities and Cross-Modal Security Workloads}

\begin{figure}[htbp]
    \centering
    \includegraphics[width=0.85\textwidth]{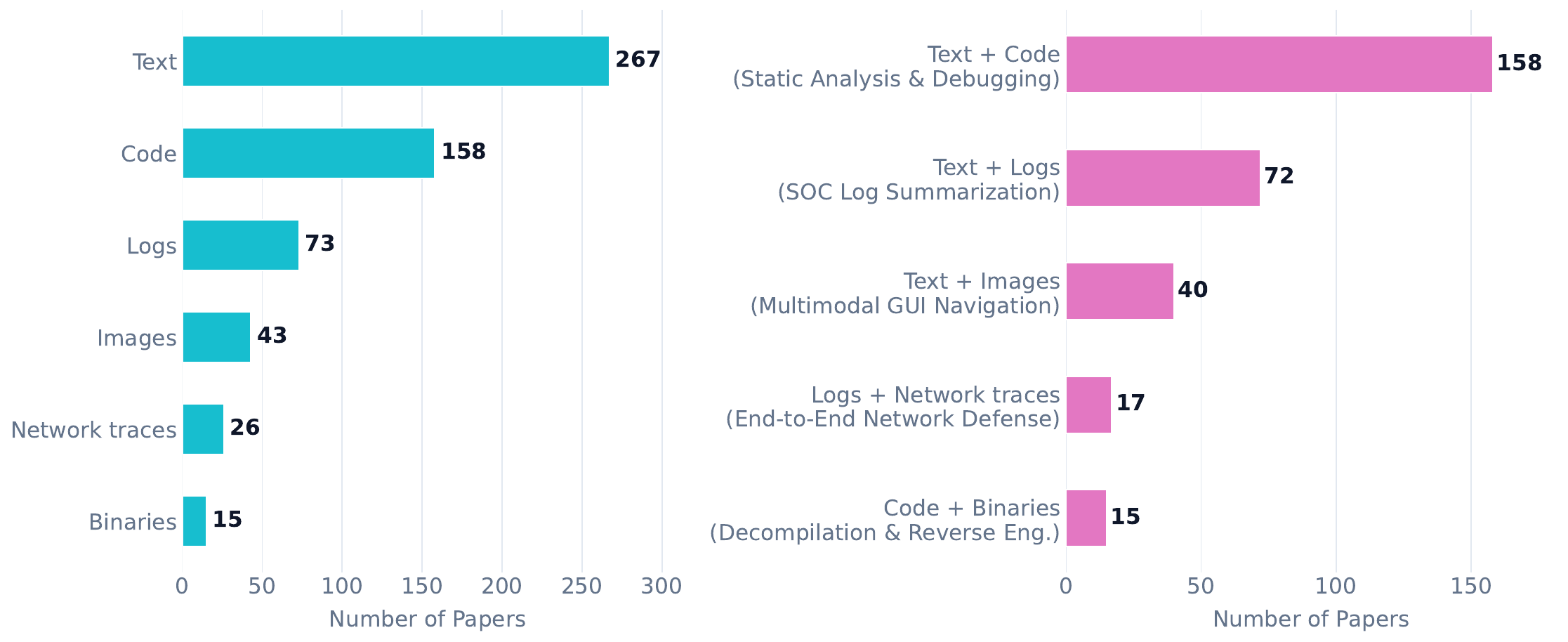}
    \caption{Empirical Census and Structural Intersections of Data Modalities in LLM Security Research. (Left) Distribution of individual input modalities analyzed across defensive literature, illustrating heavy industry path-dependency toward high-level semantic formats (Text and Code) over low-level forensic structures (Network traces and Binaries). (Right) Prevalence of multi-modal cross-evaluation suites used to check framework flexibility across data boundaries. The prominence of Text + Code pairings highlights that contemporary agent verification focuses primarily on static source auditing and natural language context processing, revealing a significant coverage gap in active end-to-end network or binary runtime security enforcement.}
    \label{fig:data_modalities}
\end{figure}

\begin{revblockv2}
This subsection evaluates the distribution of six data modalities processed by LLM agents across 272 of the 274 cybersecurity research papers in the corpus using LLMs, totaling 582 modality assignments and averaging 2.14 per paper. The taxonomy reveals how heterogeneous input streams shape threat detection, exploit generation and automated defense pipelines which are further illustrated in Figure~\ref{fig:data_modalities}.

\textbf{Text.}
Text is the dominant modality at 267 papers (98.2\%) and serves as the primary input for high-level reasoning and cyber threat intelligence parsing \citep{liu2024promptinjectionattackllmintegrated}. Security frameworks rely on in-context learning and sequential prompting to replicate human analyst workflows, though language agents face severe memory limitations and context forgetting during long-horizon security operations.

\textbf{Code.}
Code appears in 158 papers (58.1\%) and supports automated vulnerability detection, program repair, smart contract auditing and exploit generation. Systems use execution-aware pre-training alongside structural representations like abstract syntax trees, though a persistent gap between syntactic correctness and true semantic safety limits reliable output verification. Every paper that uses code also uses text, reflecting that code is always processed alongside natural-language context rather than in isolation.

\textbf{Logs.}
Log analysis appears in 73 papers (26.8\%) and mitigates SOC alert fatigue by parsing structured system events and cloud telemetry. \citet{hans2025securitylogsattckinsights} leverage LLMs to map anomalous machine behaviors to ATT\&CK tactics through provenance graphs and temporal models, though massive enterprise log volumes require aggressive pre-filtering to stay within token constraints.

\textbf{Images.}
The image modality appears in 43 papers (15.8\%) and enables vision-language models to interact with graphical user interfaces and visual web environments for coordinate grounding and autonomous navigation. Visual components nevertheless remain highly susceptible to visual prompt injections and steganographic triggers that bypass text-only guardrails.

\textbf{Network Traces.}
Network trace analysis appears in 26 papers (9.6\%) and drives real-time intrusion detection through packet captures, netflow statistics and request logs. Detection systems serialize raw packet streams into text representations or couple foundation models with reinforcement learning, though a critical latency mismatch between line-rate network speeds and model inference times limits deployment in live environments.

\textbf{Binaries.}
Binary analysis is the most specialized modality at 15 papers (5.5\%) and addresses reverse engineering, firmware fuzzing and malware decompilation. \citet{chen2025recopilotreverseengineeringcopilot} propose a reverse engineering copilot that decompiles stripped executables into structural pseudo-code to recover function profiles, though the complete absence of semantic metadata in compiled binaries severely limits identification accuracy. As with code, every binaries paper also uses code, consistent with binary analysis being a specialized extension of code-processing pipelines rather than a standalone input stream.
\end{revblockv2}

\section{Conclusion}\label{sec:conclusion}


In this survey we explore the agentic security landscape through three pillars: Applications, Threats, and Defenses. Across these pillars, our analysis shows that offensive applications concentrate in feedback-rich stages while persistence and exfiltration remain barely explored, that offensive agents are markedly more autonomous than defensive ones, and that red-teaming agents are largely dual-use. On the threat side, about half of attacks are agent-native rather than inherited or amplified from the base model, most succeed under black-box access, and most target the reasoning and action-selection stages, while the surfaces most particular to agents, namely memory, reflection, and inter-agent communication, remain the least attacked and least benchmarked. On the defense side, no single strategy is robust across all dimensions, most defenses still intervene before computation rather than after it, and prompt infection and pre-execution attacks remain the least covered attack classes. \begin{revblockv2}Overall, the community remains predominantly single-agent and monolithic, with multi-agent and planner-executor designs a substantial but secondary pattern, \end{revblockv2}while remaining dependent on GPT as a backbone and on a fragmented, largely single-use benchmark ecosystem.

\paragraph{Future Work Recommendation.} First, the least covered stages of the security lifecycle, particularly persistence and exfiltration on offense and their detection counterparts on defense, deserve dedicated study. Second, the autonomy gap suggests that defensive agents need to advance toward higher autonomy to increase their scalability. Third, adversarial investigations should focus on the agent-native surfaces of memory, reflection, and inter-agent communication need attacks, benchmarks, and defenses designed specifically for them rather than inherited from stateless LLM. Fourth, the field needs unified and balanced benchmarks that make cross-paper comparison possible and that instrument these under-evaluated surfaces. Finally, defense research should move beyond single-layer fixes toward lifecycle-wide, defense-in-depth architectures with provable safety guarantees, while broadening coverage to under-served modalities such as network traces and binaries and to the underdefended attack classes identified above.

\paragraph{Limitations.} This survey has a few limitations. It mainly focuses on software-based threats and does not explore physical-world or embodied agent attacks (like those involving robots or sensors) in detail. Our coverage is also limited to academic papers, so it may miss non-archival industrial research. \rev{Our coverage is also limited to publicly available research, which introduces a publication bias that is difficult to avoid in this domain. A great deal of security research, including both offensive capabilities evaluation and defensive tooling, are developed inside organizations and not disclosed publicly. The agentic security landscape we describe therefore reflects what the research community publishes rather than what is deployed in practice, and the gaps we identify, particularly the sparse coverage of persistence, exfiltration, and lifecycle-wide defenses, may be narrower in industry than the literature suggests.} Finally, most studies emphasize accuracy and safety rather than practical aspects like cost, speed, or energy use, and our own taxonomy involves some subjective choices. \looseness=-1

\paragraph{Broader Impact Statement.} We do not present any novel attacks or executable exploits; we only analyze peer-reviewed literature. \rev{However, the coverage of adversarial agentic capabilities could introduce a number of risks. First, several of the threats we discuss are severely dangerous because they succeed quietly. In particular, reward hacking and specification gaming produce agents that satisfy their stated objective while violating its intent, and the agent continues to report success throughout, so an operator monitoring outputs sees a working system while the threat remains silent. The same holds for agent manipulation attacks that redirect planning or action selection while leaving the final output clean. In other words, deploying agentic systems itself introduces irreversible risks, in addition to the explicit threats professed by the attackers. Second, the red-teaming agents we study include frameworks that autonomously generate evasive malware aligned to MITRE ATT\&CK. In particular, roughly half of MalGEN's generated samples evade detection by all VirusTotal engines. Furthermore, many threat papers describe directly reusable offensive capabilities, including working attack prompts, public repositories, and end-to-end social engineering pipelines. These are operating systems rather than descriptions of attacks, and they lower the expertise required to run them.}

\rev{While we acknowledge the risks, we believe a transparent academic study of vulnerabilities and countermeasures are essential for ensuring the safety and security of agentic systems. We believe standardized and isolated sandbox environments are a precondition for future offensive agent research, covering containment of generated payloads and artifacts, controlled evaluation targets, and responsible disclosure of discovered vulnerabilities.}

\bibliography{tmlr}
\bibliographystyle{tmlr}

\appendix

\section{Paper Collection Methodology}\label{appendix:paper_collection}

To ensure a comprehensive and reproducible review of the agentic security landscape, we employed a multi-stage paper collection methodology combining automated searches, manual curation, and snowballing techniques.\looseness=-1

\subsection*{Automated Database Search}
We conducted an automated search across major academic repositories, including ACL Anthology, IEEE Xplore, ACM Digital Library, and arXiv, covering publications from January 2023 to May 2026. Using a Boolean query, we combined two groups keywords related to agentic systems and security concepts.

\paragraph{Search Query Structure}
\texttt{(Group 1 Keywords) AND (Group 2 Keywords)}
\begin{itemize}
    \item \textbf{Group 1 (Agent-related):} \texttt{("LLM agent" OR "AI agent" OR "agentic AI" OR "autonomous agent" OR "multi-agent system")}
    \item \textbf{Group 2 (Security-related):} \texttt{("security" OR "threat" OR "vulnerability" OR "attack" OR "defense" OR "red team" OR "blue team" OR "penetration testing" OR "fuzzing" OR "jailbreak" OR "prompt injection" OR "poisoning" OR "hardening" OR "adversarial")}
\end{itemize}

\subsection*{Manual Curation}
To identify relevant work that our keyword search may had missed, we manually scanned the proceedings of top-tier security (e.g., USENIX, ACM CCS, Oakland) and AI (e.g., ACL, EMNLP, NeurIPS, ICLR, ICML) conferences from the same period.

\subsection*{Inclusion and Exclusion Criteria}

\paragraph{Inclusion Criteria}
The paper's primary subject must be \textbf{LLM-based agents}. It must have a substantial focus on a \textbf{technical security aspect}, aligning with one of our three pillars. The work must be a peer-reviewed publication or a preprint judged relevant to the taxonomy. \rev{Given the pace of the field, we include recent preprints that have not yet accrued many citations, but introduce an important system, benchmark or attack that is very relevant to agentic security.}

\paragraph{Exclusion Criteria}
We excluded papers on general LLM safety that do not address agentic systems, studies on non-LLM agents, works centered on high-level ethics or policy without technical details, and non-technical articles.

\subsection*{Snowballing}
Finally, we performed backward and forward snowballing on the curated set of included papers. We reviewed the reference lists of these key papers to identify foundational or related works we might have missed.

\begin{newrevblock}
\section{Coding Procedure}
\label{appendix:coding_procedure}

\paragraph{Annotation.}
The corpus was first coded against specific axes by three authors, with each being responsible for one pillar. Every pillar was then re-checked by an author who had not coded it. In cases where the two labels differed, the third author reviewed both and decided. The cross-cutting axes reported in \S\ref{sec:analysis}, namely agent cardinality,
autonomy level, architectural paradigm, cognitive roles, model families, knowledge
sources and data modalities, are properties of a paper rather than of a pillar and
are recorded once per paper.

\paragraph{Pillar overlaps.}
A paper can be coded under more than one pillar (e.g., a paper introducing both an attack and a defense for it), so the pillar-wise counts do not sum to the total corpus size. Applications account for 113 papers, Threats for 86, and Defenses for 123. Figure~\ref{fig:corpus_composition_venn} shows the pairwise and three-way overlaps and the resulting deduplicated total. Every percentage reported in a single-pillar subsection (\S\ref{sec:applications}, \S\ref{sec:threats}, \S\ref{sec:defense}) uses that pillar's raw count as its denominator unless a smaller analysis set is stated; every percentage reported in the cross-cutting analysis (\S\ref{sec:analysis}) uses the deduplicated total of 274 papers unless a smaller analysis set is stated.

\begin{figure}[h]
\centering
\includegraphics[width=0.6\columnwidth]{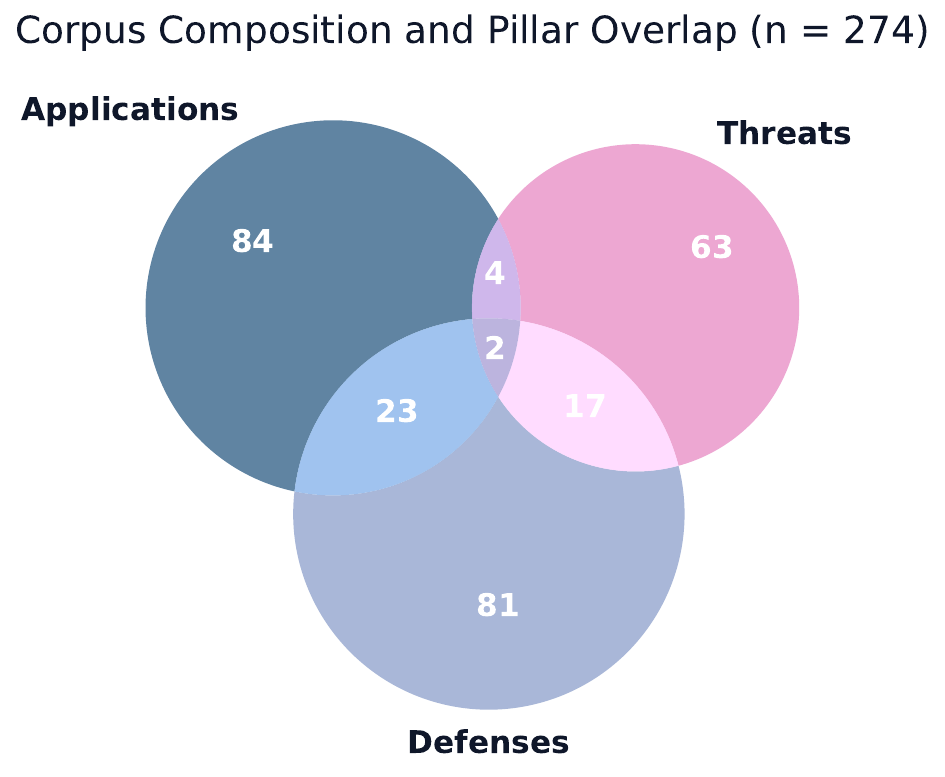}
\caption{Corpus composition across the three pillars. Each region gives the number of papers coded under that combination of pillars; the deduplicated total is 274 papers.}
\label{fig:corpus_composition_venn}
\end{figure}

\paragraph{Decision rules for the least self-evident axes.}
Two axes require boundaries that the literature does not fix, and we state the rules
applied. For \textit{agent-specificity}, a threat is \emph{inherited} if it is
demonstrated against a standalone model and carries into the agent with no change in
mechanism or effectiveness, \emph{amplified} if the mechanism exists in the base
model but the paper reports a higher success rate, a lower refusal rate, or a broader
reachable surface once the scaffold is present, and \emph{agent-native} if the attack
has no realisable form without a scaffold component, meaning memory, a tool
interface, an inter-agent channel, or a planning loop. Where a paper reports both a
standalone and an agentic condition, that comparison decides the label; where it
reports only the agentic condition, we assign \emph{amplified} only if the paper
attributes the effect to a specific scaffold component, and otherwise leave the paper
unclassified. For \textit{failure stage}, we code the agent-loop stage at which
behaviour first deviates from correct execution, which is not always the stage at
which the payload enters. Memory-poisoning attacks enter through the user query or a
memory write but are coded at retrieval, since that is where the corrupted record
re-enters the loop and begins to steer action selection. Entry point and failure
stage are recorded as separate columns for every threat paper.

\paragraph{Handling of undeterminable assignments.}
Assignments that cannot be determined from the paper are labeled as unknown (\texttt{unk}) rather than beign forced into a category, and the unclassified share is reported alongside each affected figure: architectural paradigm (17 papers), entry point (12), agent-loop
stage (8), and agent-specificity (8). Autonomy level is not a
meaningful property of a paper that presents no system of its own, so the 53
benchmark, survey and position papers in this group are excluded from that
analysis rather than assigned a value. The 44 papers with no cognitive role coded are
excluded from the role analysis, and the 31 defense papers that propose or evaluate no
defense technique are excluded from the defense analyses. The reduced denominator is stated where
the figures appear.
Attacker capability requirements are recorded only where a paper states a requirement beyond model access, and a blank denotes no stated requirement.
Two axes involve judgment that reasonable coders could exercise differently, the
separation of amplified from agent-native threats and the assignment of a primary
defense category to papers combining several mechanisms, and the released corpus
records the basis for each of these assignments.

\end{newrevblock}

\begin{newrevblock}
\subsection*{Corpus Release and Data Availability}
\label{appendix:corpus_release}

We release the coded corpus so that others can check and reuse our results. It is a single table with one row for each of the 274 papers and 43 fields that record how we coded each paper. Table~\ref{tab:corpus_schema} sorts these fields into five groups. Metadata and structural fields are filled for every paper. Application, threat and defense fields are filled only for papers in that pillar. When a paper has several values for one field, such as several data modalities, they are listed in that field separated by semicolons. Every count, percentage and figure in this survey is computed from this table.

\begin{table}[h]
\centering
\caption{Corpus schema. Pillar-specific groups are populated only for papers coded under that pillar; the other two groups are populated for every row.}
\label{tab:corpus_schema}
\begin{tabularx}{\linewidth}{lcc>{\raggedright\arraybackslash}X}
\toprule
\textbf{Field group} & \textbf{Papers} & \textbf{\# Fields} & \textbf{Example fields} \\
\midrule
Corpus metadata & 274 & 8 & paper identifier, pillars a paper is coded under \\
\addlinespace
Structural & 274 & 9 & autonomy level, agent cardinality, architecture paradigm, cognitive roles, backbone models, knowledge sources, data modalities \\
\addlinespace
Application & 113 & 3 & category, lifecycle stage(s), autonomy level \\
\addlinespace
Threat & 86 & 15 & threat class, entry point, agent-loop stage targeted, attacker model, attacker goal, evaluation metric \\
\addlinespace
Defense & 123 & 8 & trust boundary, cost/security tradeoff, defensive technique, defense positioning, defense coverage \\
\midrule
\textbf{Total} & & \textbf{43} & \\
\bottomrule
\end{tabularx}
\end{table}
\end{newrevblock}




\end{document}